\documentclass{article}

\usepackage[final]{neurips_2024}

\usepackage[utf8]{inputenc} %
\usepackage[T1]{fontenc}    %
\usepackage{hyperref}       %
\usepackage{url}            %
\usepackage{booktabs}       %
\usepackage{amsfonts}       %
\usepackage{nicefrac}       %
\usepackage{microtype}      %
\usepackage[dvipsnames,table]{xcolor} %
\usepackage{amssymb}        %
\usepackage{pifont}
\definecolor{mygreen}{RGB}{0,128,0}

\usepackage{svg}
\usepackage{array}
\usepackage{amsmath}
\usepackage{doi}
\usepackage{makecell}
\usepackage{multirow}
\usepackage{graphicx}
\usepackage{subcaption}
\usepackage{geometry}
\usepackage{comment}
\usepackage{floatrow}
\usepackage{wrapfig}
\usepackage{placeins}  %
\usepackage{tabularx}
\usepackage{mathtools}
\newcolumntype{Y}{>{\centering\arraybackslash}X}

\usepackage{booktabs}
\usepackage{floatrow}
\usepackage{makecell}
\usepackage{titlesec}
\usepackage[capitalize,noabbrev]{cleveref}
\usepackage{comment} 
\usepackage{siunitx}     %
\usepackage[titletoc,toc,page]{appendix}
\usepackage{makecell}
\usepackage{titlesec}
\usepackage[capitalize,noabbrev]{cleveref}

\newcommand{\hider}[1]{}

\newcommand{\hreffoot}[2]{\href{#1}{#2}\footnote{\url{#1}}}

\newcommand{\xmark}{\ding{55}}
\newcommand{\myxmark}{{\color{gray!40}\xmark{}}}
\newcommand{\mycmark}{{\color{black}\checkmark{}}}

\usepackage{xspace}

\usepackage{soul}           %

\makeatletter
\DeclareRobustCommand\onedot{\futurelet\@let@token\@onedot}
\def\@onedot{\ifx\@let@token.\else.\null\fi\xspace}

\def\vs{\emph{vs}\onedot}
\makeatother

\usepackage{enumitem}

\newcommand{\mypara}[1]{\vspace{4pt}\noindent\textbf{#1}}

\newcommand{\tcf}[1]{\textbf{#1}}
\newcommand{\tcs}[1]{\underline{#1}}

\newcommand{\tcg}[1]{\underline{#1}}

\definecolor{cbg}{cmyk}{0.8,0.4,0,0.012}
\definecolor{cbr}{cmyk}{0,0.75,0.75,0.1}

\newcommand{\clibd}{CLIBD\xspace}

\newcommand{\appref}[1]{\hyperref[#1]{Appendix~\ref*{#1}}}

\def\Snospace~{\S{}}

\definecolor{linkcolor}{HTML}{991408}  %
\definecolor{citecolor}{HTML}{2E7E2A}  %
\definecolor{filecolor}{HTML}{131877}  %
\definecolor{menucolor}{HTML}{727500}  %
\definecolor{runcolor} {HTML}{137776}  %
\definecolor{urlcolor} {HTML}{0a2bbf}  %
\hypersetup{colorlinks=true,linkcolor=linkcolor,citecolor=citecolor,filecolor=filecolor,menucolor=menucolor,runcolor=runcolor,urlcolor=urlcolor}

\title{BIOSCAN-5M: A Multimodal Dataset for\\ Insect Biodiversity}

\author{%
	Zahra Gharaee$^{3*}$,
	Scott C.~Lowe$^{5*}$, 
	ZeMing Gong$^{4*}$, 
	Pablo Millan Arias$^{3*}$,\\ 
	\textbf{Nicholas Pellegrino}$^{3}$,
	\textbf{Austin T.~Wang}$^{4}$,
	\textbf{Joakim Bruslund Haurum}$^{7}$, \\
	\textbf{Iuliia Zarubiieva}$^{2,5}$, 
	\textbf{Lila Kari}$^{3}$, \\
	\textbf{Dirk Steinke$^{1,2\dagger}$,
		Graham W.~Taylor$^{2,5\dagger}$,
		Paul Fieguth$^{3\dagger}$,
		Angel X.~Chang$^{4,6\dagger}$}
	\\
	$^{1}$Centre for Biodiversity Genomics, 
	$^{2}$University of Guelph, 
	$^{3}$University of Waterloo, \\
	$^{4}$Simon Fraser University,
	$^{5}$Vector Institute,
	$^{6}$Alberta Machine Intelligence Institute (Amii), \\
	$^{7}$Aalborg University and Pioneer Centre for AI\\
	{\url{https://biodiversitygenomics.net/5M-insects/}}
}

\newtoggle{arxiv}
\toggletrue{arxiv}%

\iftoggle{arxiv}{%
\newcommand{\maybesupplementref}[1]{\appref{#1}}
\newcommand{\maybesupplementreflong}[1]{\appref{#1}}
}{%
\newcommand{\maybesupplementref}[1]{supplement}
\newcommand{\maybesupplementreflong}[1]{the supplementary materials}
}

\begin{document}

\maketitle

\begin{abstract}

As part of an ongoing worldwide effort to comprehend and monitor insect biodiversity, this paper presents the BIOSCAN-5M Insect dataset to the machine learning community and establish several benchmark tasks.  
BIOSCAN-5M is a comprehensive dataset containing multi-modal information for over 5 million insect specimens, and it significantly expands existing image-based biological datasets by including taxonomic labels, raw nucleotide barcode sequences, assigned barcode index numbers, geographical, and size information.
We propose three benchmark experiments to demonstrate the impact of the multi-modal data types on the classification and clustering accuracy. 
First, we pretrain a masked language model on the DNA barcode sequences of the \mbox{BIOSCAN-5M} dataset, and demonstrate the impact of using this large reference library on species- and genus-level classification performance.
Second, we propose a zero-shot transfer learning task applied to images and DNA barcodes to cluster feature embeddings obtained from self-supervised learning, to investigate whether meaningful clusters can be derived from these representation embeddings. Third, we benchmark multi-modality by performing contrastive learning on DNA barcodes, image data, and taxonomic information. This yields a general shared embedding space enabling taxonomic classification using multiple types of information and modalities. The code repository of the BIOSCAN-5M Insect dataset is available at {\url{https://github.com/bioscan-ml/BIOSCAN-5M}}.

\end{abstract}
\def\thefootnote{$*$}\footnotetext{Joint first author. $^\dagger$Joint senior/last author.}
\renewcommand*{\thefootnote}{\arabic{footnote}}
\section{Introduction}
\label{sec:intro}

Biodiversity plays a multifaceted role in sustaining ecosystems and supporting human well-being. Primarily, it serves as a cornerstone for ecosystem stability and resilience, providing a natural defence against disturbances such as climate change and invasive species~\citep{cardinale2012biodiversity}. Additionally, biodiversity serves as a vital resource for the economy, supplying essentials like food, medicine, and genetic material~\citep{sala2000global}.
Understanding biodiversity is paramount for sustainable resource management, ensuring the availability of these resources for future generations~\citep{duraiappah2005ecosystems}. 
{To understand and monitor biodiversity, \citet{gharaee2023step} introduced the BIOSCAN-1M Insect dataset, which pairs DNA with images, as a stepping stone to developing AI tools for automatic classification of organisms.}

\begin{figure*}[tbh]
	\centering    
	\includegraphics[width=1\textwidth]{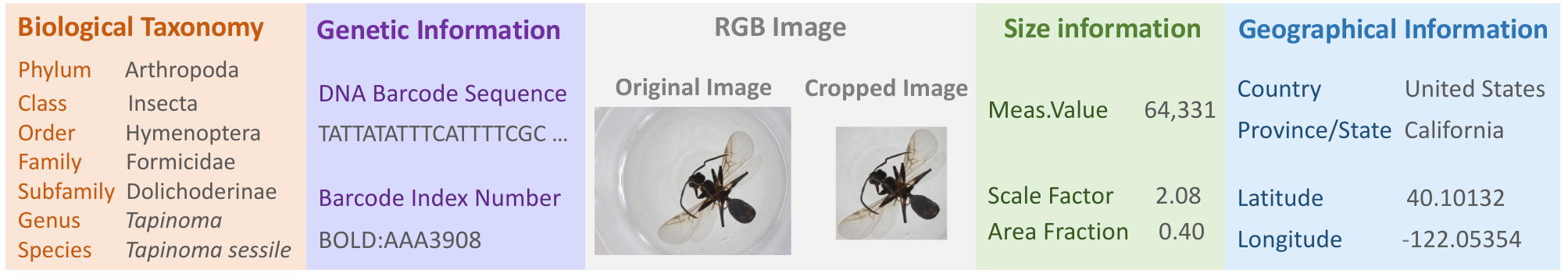}
 \caption{\textbf{Record attributes.} The BIOSCAN-5M dataset provides taxonomic labels, a DNA barcode sequence, barcode index number, a high-resolution image along with its cropped and resized versions, as well as size and geographic information for each sample.}
 \label{fig:dataset-overview}
\end{figure*}

{However, that work only investigated image classification down to the family level, focusing on the Diptera order, and did not fully utilize the multimodal nature of the dataset. 
In addition, BIOSCAN-1M was limited to specimen collected from just 3 countries and the \textit{Insecta} \texttt{class}.} 
Expanding upon BIOSCAN-1M, we introduce the BIOSCAN-5M dataset---a comprehensive repository of multi-modal information (see \autoref{fig:dataset-overview}) on over 5 million arthropod specimens (98\% insects), with 1.2 million labelled to \texttt{genus} or \texttt{species} taxonomic ranks. Compared to its predecessor, the BIOSCAN-5M dataset offers a significantly larger volume of high-resolution microscope images and DNA barcodes along with critical annotations, including taxonomic ranks, size, and geographical information.
{Additionally, we performed data cleaning to resolve inconsistencies and provide more reliable labels.}

The multimodal characteristics {of} BIOSCAN-5M are not only essential for biodiversity studies, but also facilitate further innovation in machine learning and AI. {In this paper, we conduct experiments that leverage the multimodal aspects of BIOSCAN-5M}, extending its application beyond the image-only modality used in~\citet{gharaee2023step}. 
Here, we train the masked language model (MLM) proposed in BarcodeBERT~\citep{millan2023barcodebert} on the DNA barcodes of the BIOSCAN-5M dataset and demonstrate the impact of using this large reference library on species- and genus-level classification. We achieve an accuracy higher than that of state-of-the-art models pretrained on more general genomic datasets, especially in the 1NN-probing task of assigning samples from unseen species to seen genera. %
Next, we perform a zero-shot transfer learning task~\citep{zsc} through zero-shot clustering representation embeddings obtained from encoders trained with self-supervised paradigms. This approach demonstrates the effectiveness of pretrained embeddings in clustering data, even in the absence of ground-truth. Finally, as in \clibd~\citep{gong2024bioscanclip}, we learn a shared embedding space across three modalities in the dataset---high-quality RGB images, textual taxonomic labels, and DNA barcodes---for fine-grained taxonomic classification.

\section{Related work}
\label{sec:background}
\subsection{Datasets for taxonomic classification} %

Biological datasets are essential for advancing our understanding of the natural world, with uses in genomics~\citep{cgarn:2013}, proteomics~\citep{kim:2014}, ecology~\citep{try:2011}, evolutionary biology~\citep{ensembl:2014}, medicine~\citep{jensen:2012}, and agriculture~\citep{lu2020survey,xu2023pink,galloway2017ciona17,he2024species196}. \autoref{tab:agg_datasets} compares biological datasets used for taxonomic classification. Many of these datasets feature fine-grained classes and exhibit a long-tailed class distribution, making the recognition task challenging for machine learning (ML) methods that do not account for these properties. While many datasets provide images, they do not include other attributes such as DNA barcode, or geographical locations. Most relevant to our work is BIOSCAN-1M Insect~\citep{gharaee2023step}, which introduced a dataset of 1.1\,M insect images paired with DNA barcodes and taxonomic labels.

DNA barcodes are short, highly descriptive DNA fragments that encode sufficient information for species-level identification. For example, a DNA barcode of an organism from Kingdom Animalia~\citep{hebert2003biological,braukmann2019metabarcoding} is a specific 648 bp sequence of the cytochrome c oxidase I (COI) gene from the mitochondrial genome, used to classify unknown individuals and discover new species~\citep{moritz2004dna}. DNA barcodes have been successfully applied to taxonomic identification and classification, ecology, conservation, diet analysis, and food safety~\citep{ruppert2019past,stoeck2018environmental}, offering faster and more accurate results than traditional methods~\citep{pawlowski2018future}. 
Barcodes can also be grouped together based on sequence similarity into clusters called Operational Taxonomic Units (OTUs)~\citep{sokal1963principles,blaxter2005defining}, each assigned a Barcode Index Number (BIN)~\citep{ratnasingham2013dna}.
In general, biological datasets may also incorporate other data such as labels for multi-level taxonomic ranks, which can offer valuable insights into the evolutionary relationships between organisms. However, datasets with hierarchical taxonomic annotations~\citep{he2024species196,ilyas2023cwd30,datasetplant,wu2019ip102,gharaee2023step} are relatively scarce.

\begin{table}[t]
	\caption{
    \textbf{Summary of fine-grained and long-tailed biological datasets.} %
    The ``Taxa'' column indicates the taxonomic scope of each dataset.
    The ``IR'' column is the class imbalance ratio, computed as the ratio of the number of samples in the largest category to the smallest category.}
	\label{tab:agg_datasets}
	\centering
	\resizebox{1.01\textwidth}{!}
	{%
		\begin{small}
			\begin{tabular}{@{}llcrrllrccccc@{}}
				\toprule
				{Dataset} &{Reference} & {Year} & {Images} & {IR} & {Taxa} & {Rank} & {Categories} & {Taxon} & {BIN} &{DNA} &{Geography} &{Size} \\ 
				\midrule

				LeafSnap&\cite{kumar2012leafsnap}&2012 & 31\,k & 8&Plants  & Species & 184 & \myxmark{} &\myxmark{} &\myxmark{} &\myxmark{} &\myxmark{}\\
    
                NA Birds&\cite{van2015building}&2015 & 48\,k & 15& Birds  & Species & 400 & \myxmark{} &\myxmark{} & \myxmark{}& \myxmark{}&\myxmark{} \\ 
            			
				Urban Trees&\cite{wegner2016cataloging}&2016 & 80\,k& 7 & Trees  & Species&18 & \myxmark{} &\myxmark{} & \myxmark{} &\myxmark{}&\myxmark{}\\ 

                DeepWeeds&\cite{olsen2019deepweeds}&2019 & 17\,k& 9 & Plants  & Species&9 & \myxmark{} &\myxmark{} & \myxmark{}&\mycmark{}&\myxmark{}\\ 
                
                IP102&\cite{wu2019ip102}&2019 & 75\,k&14  & Insects & Species&102 & \mycmark{} &\myxmark{} &\myxmark{} &\myxmark{}&\myxmark{}\\ 

                Pest24&\cite{pest24_wang2020}&2020& 25\,k& 494 & Insects  & Species & 24 & \myxmark{} &\myxmark{} & \myxmark{} &\myxmark{}&\myxmark{}\\ 
				
                {Pl@ntNet-300K}&\cite{garcin2021pl}&2021& 306\,k& 3,604 & Plants  & Species&1,000 & \myxmark{} &\myxmark{} & \myxmark{} &\myxmark{}&\myxmark{}\\ 
				
				iNaturalist (2021)&\cite{van2021benchmarking}&2021& 2,686\,k& 2 & All  & Species & 10,000 & \mycmark{}&\myxmark{} & \myxmark{} &\myxmark{}&\myxmark{}\\ 

				iNaturalist-Insect&\cite{van2021benchmarking}&2021& 663\,k& 2 & Insects  & Species&2,526 & \mycmark{}&\myxmark{} & \myxmark{}&\myxmark{}&\myxmark{}\\ 

                Species196-L&\cite{he2024species196}&2023& 19\,k& 351  & Various  & Mixed&196 & \mycmark{} &\myxmark{} & \myxmark{} &\myxmark{}&\myxmark{}\\

                CWD30&\cite{ilyas2023cwd30}&2023& 219\,k& 61  & Plants  & Species&30 & \mycmark{} &\myxmark{} & \myxmark{} &\myxmark{} &\myxmark{}\\

                BenthicNet&\cite{benthicnet}&2024& 1,429\,k& 22,394  & Aquatic  & Mixed & 791 & \mycmark{} &\myxmark{} & \myxmark{} &\mycmark{} &\myxmark{}\\

                Insect-1M&\cite{nguyen2023insect}&2024& 1,017\,k& N/A  & Arthropods  & Species&34,212 & \mycmark{} &\myxmark{} & \myxmark{} &\myxmark{} &\myxmark{}\\

                BIOSCAN-1M&\cite{gharaee2023step}&2023& 1,128\,k& 12,491  & Insects  & BIN* &90,918 & \mycmark{} &\mycmark{} & \mycmark{} & \myxmark{}&\myxmark{}\\ 
				\midrule
                \textbf{BIOSCAN-5M}&Ours&2024& 5,150\,k&35,458  & Arthropods  & BIN* &324,411 & \mycmark{} &\mycmark{} &\mycmark{} & \mycmark{}&\mycmark{}\\ 
				\bottomrule
				
			\end{tabular}
		\end{small}
		}
\begin{flushleft}
\scriptsize{* For datasets that include Barcode Index Numbers (BINs) annotations, we present BINs, which serve as a (sub)species proxy for organisms and offer a viable alternative to Linnean taxonomy.}
\end{flushleft}  
\vspace{-2mm}
\end{table}

\subsection{Self-supervised learning} 

Self-supervised learning (SSL) has recently gained significant attention for its ability to leverage vast amounts of unlabelled data, producing versatile feature embeddings for various tasks~\citep{balestriero2023cookbook}. This has driven the development of large-scale language models~\citep{brown2020language} and computer vision systems trained on billions of images~\citep{goyal2021self}. 
Advances in transformers pretrained with SSL at scale, known as foundation models~\citep{ji2021dnabert,zhou2023dnabert,dalla2023nucleotide,zhou2024dnabert,chia2022contrastive,gu2021domain}, have shown robust performance across diverse tasks. %

Recent work has leveraged these advances for taxonomic classification using DNA.
Since the introduction of the first DNA language model, DNABERT~\citep{ji2021dnabert}, which mainly focused on human data,  multiple models with different architectures and tokenization strategies have emerged~\citep{mock2022taxonomic,zhou2023dnabert,zhou2024dnabert,millan2023barcodebert,nguyen2024hyenadna} with some incorporating data from multiple species during pretraining and allowing for species classification~\citep{zhou2023dnabert,zhou2024dnabert,millan2023barcodebert}. These models are pretrained to be task-agnostic, and are expected to perform well after fine-tuning in downstream tasks. Yet, their potential application for taxonomic identification of arbitrary DNA sequences or DNA barcodes has not been extensively explored. One relevant approach, BERTax~\citep{mock2022taxonomic}, pretrained a BERT~\citep{dosovitskiy2020image} model for hierarchical taxonomic classification on broader ranks such as kingdom, phylum, and genus.  For DNA barcodes specifically, BarcodeBERT~\citep{millan2023barcodebert} was developed for species-level classification of insects, with assignment to genus for unknown species.

Although embeddings from SSL-trained feature extractors exhibit strong performance on downstream tasks post fine-tuning, their utility without fine-tuning remains underexplored. Previous studies~\citep{vaze2022generalized,zhou2022deep} suggest that SSL feature encoders produce embeddings conducive to clustering, albeit typically after fine-tuning. A recent study~\citep{zsc} has delved into whether SSL-trained feature encoders \emph{without} fine-tuning can serve as the foundation for clustering, yielding informative clusters of embeddings on real-world datasets unseen during encoder training.

\subsection{Multimodal Learning} 
There has been a growing interest in exploring multiple data modalities for biological tasks~\citep{ikezogwo2024quilt, lu2023towards, zhang2023biomedclip}. \citet{Badirli2021-zv} introduced a Bayesian zero-shot learning approach, leveraging DNA data to model priors for species classification based on images. Those authors also employed Bayesian techniques~\citep{badirli2023classifying}, combining image and DNA embeddings in a unified space to predict the genus of unseen species. 

Recent advances in machine learning allowed scalable integration of information across modalities. For example, CLIP~\citep{radford2021learning} used contrastive learning to encode text captions and images into a unified space for zero-shot classification. 
BioCLIP~\citep{stevens2023bioclip} used a similar idea to align images of organisms with their common names and taxonomic descriptions across a dataset of 10\,M specimens encompassing plants, animals, and fungi. \clibd~\citep{gong2024bioscanclip} used a contrastive loss to align the three modalities of RGB images, textual taxonomic labels, and DNA barcodes. By aligning these modalities, \clibd can use either images or DNA barcodes for taxonomic classification and learn from incomplete taxonomic labels, making it more flexible than BioCLIP~\citep{stevens2023bioclip}, which requires full taxonomic annotations for each specimen.

\section{Dataset} 
\label{sec:dataset}

The BIOSCAN-5M dataset is derived from~\citet{steinke2024dataset} and comprises 5,150,850 \texttt{arthropod} specimens, with insects accounting for about 98\% of the total. The diverse features of this dataset are described in this section. BIOSCAN-5M is a superset of the BIOSCAN-1M Insect dataset~\citep{gharaee2023step}, providing more samples and additional metadata such as geographical location.

\begin{figure*}[!h]
	\centering    
	\includegraphics[width=1\textwidth]{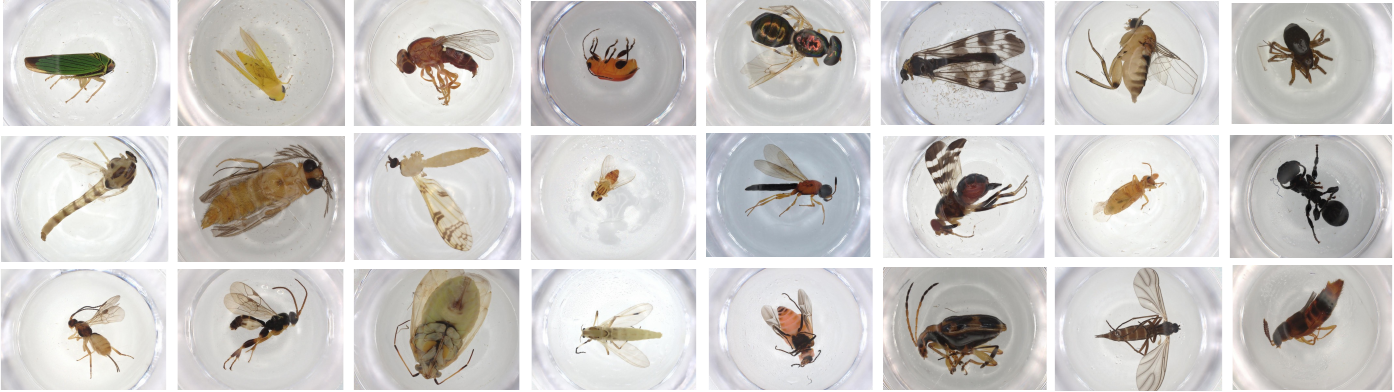}
    \vspace{-5mm}
	\caption{Samples of original full-size images of distinct organisms in the BIOSCAN-5M dataset.}
	\label{fig:image_samples}
\end{figure*}

\mypara{Images.}  
The BIOSCAN-5M dataset provides specimen images at 1024\texttimes768 pixels, captured using a Keyence VHX-7000 microscope.  \autoref{fig:image_samples} showcases the diversity in organism morphology across the dataset. The images are accessed via the \texttt{processid} field of the metadata as \{\texttt{processid}\}.jpg. Following BIOSCAN-1M Insect~\citep{gharaee2023step}, the images are cropped and resized to 341\texttimes256 pixels to facilitate model training. We fine-tuned DETR (End-to-End Object Detection with Transformers) for image cropping. For BIOSCAN-1M Insect, the cropping model was trained using 2\,k insect images. Building on the BIOSCAN-1M Insect cropping tool checkpoint, we fine-tuned the model for BIOSCAN-5M using the same 2\,k images and an additional 837 images that were not well-cropped previously. This fine-tuning process followed the same training setup, including batch size, learning rate, and other hyper parameter settings (see \maybesupplementref{s:image-processing} for details). The bounding box of the cropped region is provided as part of the dataset release. 

\mypara{Genetic-based indexing.} 
The genetic information of the BIOSCAN-5M dataset described in \autoref{sec:background} is represented as the raw nucleotide barcode sequence, under the \texttt{dna\_barcode} field, and the Barcode Index Number under \texttt{dna\_bin} field.
Independently, the field \texttt{processid} is a unique number assigned by BOLD~\citep{bold_systems} to each record, and \texttt{sampleid} is an identifier given by the collector.

\mypara{Biological taxonomic classification.} 
{Linnaean taxonomy is a hierarchical classification system instigated by~\citet{SystemaNaturae} for organizing living organisms which has been developed over several hundred years. It categorizes \texttt{species} based on shared characteristics and establishes a standardized naming convention. The hierarchy includes several taxonomic ranks, such as \texttt{domain}, \texttt{kingdom}, \texttt{phylum}, \texttt{class}, \texttt{order}, \texttt{family}, \texttt{genus}, and \texttt{species}, allowing for a structured approach to studying biodiversity and understanding the relationships between different organisms. }

The dataset samples undergo taxonomic classification using a hybrid approach involving an AI-assisted tool proposed by \citet{gharaee2023step} and human taxonomic experts. After DNA barcoding and sequence alignment, the taxonomic levels derived from both the AI tool and DNA sequencing are compared. Any discrepancies are then reviewed by human experts. Importantly, assignments to deeper taxonomic levels, such as \texttt{family} or lower, rely entirely on human expertise. Labels at seven taxonomic ranks are used to represent individual specimens, denoted by fields
\texttt{phylum}, \texttt{class}, \texttt{order}, \texttt{family}, \texttt{subfamily}, \texttt{genus}, and \texttt{species}.

\begin{table}[!ht]
	\begin{center}
    \resizebox{\textwidth}{!}{%
        \begin{tabular}{lrrrrrrrr}
					\toprule
					& \multicolumn{4}{c}{\textbf{BIOSCAN-5M} (Ours)} & \multicolumn{3}{c}{\textbf{BIOSCAN-1M~\citep{gharaee2023step}}}\\
					\cmidrule(lr){2-5} \cmidrule(lr){6-8}
					Attributes &  IR &  Categories &  Labelled &  Labelled (\%)  &  Categories &Labelled &  Labelled (\%)  \\
					\midrule
					\texttt{phylum}       &         1 &         1 & 5,150,850 & 100.0 &       1 & 1,128,313 & 100.0 \\
					\texttt{class}        &   719,831 &        10 & 5,146,837 &  99.9 &       1 & 1,128,313 & 100.0 \\
					\texttt{order}        & 3,675,317 &        55 & 5,134,987 &  99.7 &      16 & 1,128,313 & 100.0 \\
					\texttt{family}       &   938,928 &       934 & 4,932,774 &  95.8 &     491 & 1,112,968 &  98.6 \\
					\texttt{subfamily}    &   323,146 &     1,542 & 1,472,548 &  28.6 &     760 &   265,492 &  23.5 \\
					\texttt{genus}        &   200,268 &     7,605 & 1,226,765 &  23.8 &   3,441 &   254,096 &  22.5 \\
					\texttt{species}      &     7,694 &    22,622 &   473,094 &   9.2 &   8,355 &    84,397 &   7.5 \\
					\midrule
					\texttt{dna\_bin}     &    35,458 &   324,411 & 5,137,441 &  99.7 &  91,918 & 1,128,313 & 100.0 \\
					\texttt{dna\_barcode} &     3,743 & 2,486,492 & 5,150,850 & 100.0 & 552,629 & 1,128,313 & 100.0 \\
					\bottomrule
				\end{tabular}
    }
    \end{center} 		
    \caption{\textbf{Summary statistics} of dataset records by taxonomic rank: imbalance ratio (IR) between most and least common labels, number of unique labels, and number of labelled samples.}
    \label{tab:gen_stat}
\end{table}

In the source data, we found identical DNA nucleotide sequences labelled differently at some taxonomic levels, which was likely due to human error (e.g. typos) or disagreements in the taxonomic labelling. To address this, we checked and cleaned the taxonomic labels to address typos and ensure consistency across DNA barcodes (see \maybesupplementref{s:label-cleaning} for details). 
We note that some of the noisy species labels are placeholder labels that do not correspond to well-established scientific taxonomic species names.  
In our data, the placeholder \texttt{species} labels are identified by \texttt{species} labels that begin with a lowercase letter, contain a period, contain numerals, or contain ``malaise''. 

Statistics for BIOSCAN-5M are given in \autoref{tab:gen_stat} for the seven taxonomic ranks along with the BIN and DNA nucleotide barcode sequences. For each group, we report the number of categories, and the count and fraction labelled. We compute the class imbalance ratio (IR) as the ratio of the number of samples in the largest category to the smallest category, reflecting the class distribution within each group. For more detailed statistical analysis, see \maybesupplementreflong{s:dataset-stats}.

\begin{figure}[!h]
	\centering    
	\includegraphics[width=1\textwidth]{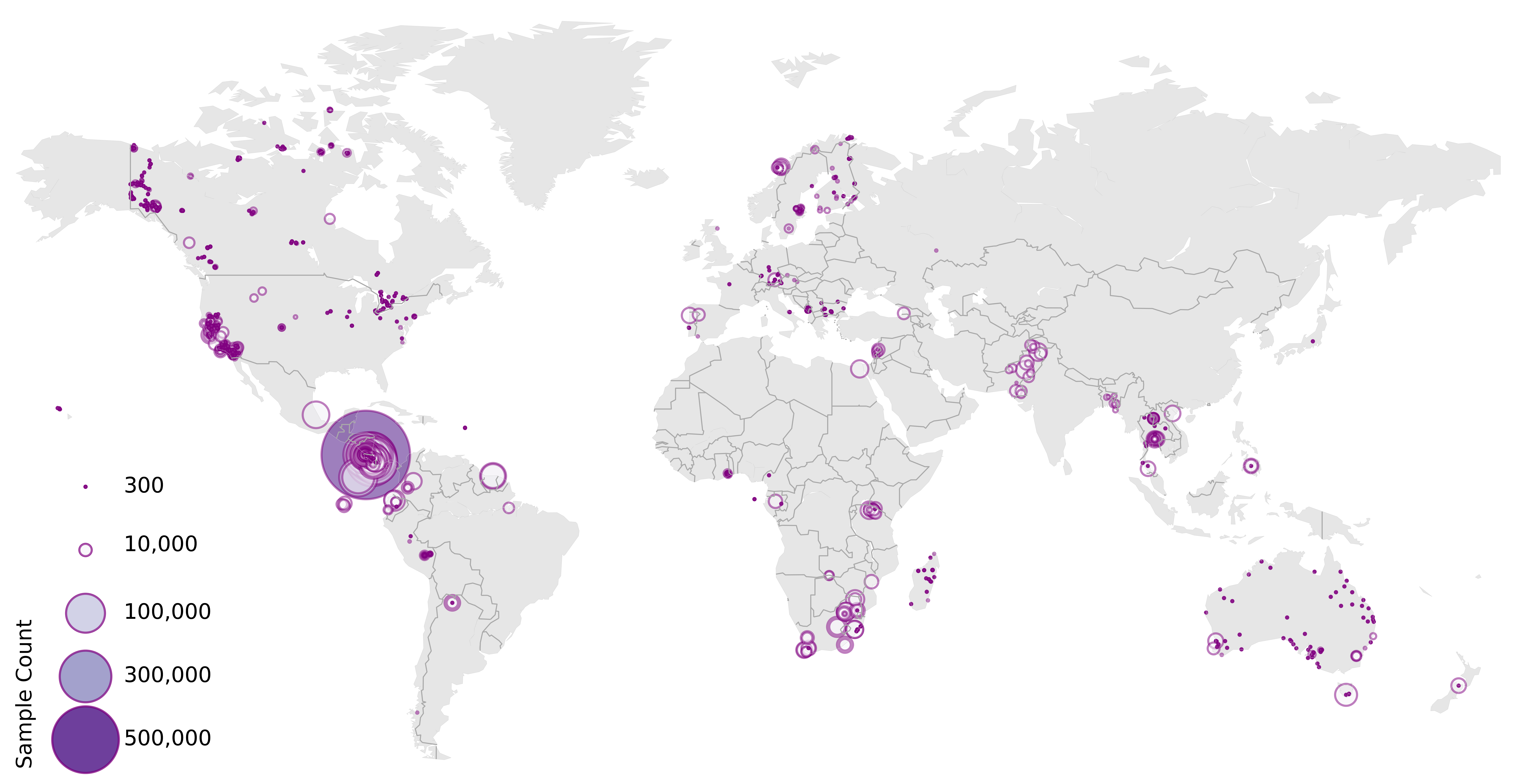}
 \caption{\textbf{Geographical locations} obtained from latitude and longitude coordinates of the regions where the samples of the BIOSCAN-5M dataset were collected.}
 \label{fig:geo-stat}
\end{figure}

\mypara{Geographic location.} 
The BIOSCAN-5M dataset includes geographic location information, detailing the country and province or state where each specimen is collected, along with the latitude and longitude coordinates of each collection site. This information is detailed in the fields \texttt{country}, \texttt{province\_state}, \texttt{coord-lat} and \texttt{coord-lon}. The distribution of specimen collection sites are shown on a world map in \autoref{fig:geo-stat}.

\mypara{Challenges.} The BIOSCAN-5M dataset faces two key challenges: First, there exists a sampling bias as a result of the locations where and the methods through which specimens are collected. Second, the number of labelled records sharply declines at deeper taxonomic levels, especially beyond the family rank, which makes fine-grained classification tasks more challenging.

\section{Benchmark experiments and results}
\label{sec:experiments}

In real-world insect biodiversity monitoring, it is common to encounter both species which are already known to science, and samples whose species is novel.
Thus, to excel in biodiversity monitoring, a model must correctly categorize instances of known species, and identify novel species outside the existing taxonomy, grouping together samples of the same new species.
In our experiments, we explore three methods which offer utility in these regards,
evaluated in two settings: closed-world and open-world.
In the closed-world setting, the task is to accurately identify species from a predefined set of existing labels.
In the open-world setting, the task is to group together samples of novel species.

\subsection{Data partitioning}
\label{sec:partitioning}

\mypara{Species sets.} We first partition records based on their species label into one of four categories, with all samples bearing the same species label being placed in the same species set.
\emph{Seen}: all samples whose species label is an established scientific name of a species.
\emph{Unseen}: labelled with an established scientific name for the genus, and a uniquely identifying placeholder name for the species.
\emph{Heldout}: labelled with a placeholder genus and species name.
\emph{Unknown}: samples without a species label (note: these may truly belong in any of the other three categories).

\begin{table}[bht]
	\caption{\textbf{Statistics and purpose of our data partitions.}}
    \vspace{-8pt}
	\centering
    \resizebox{\textwidth}{!}{%
    \begin{footnotesize}
    \begin{tabular}{lllrrr}
        \toprule
        Species set & Split                  & Purpose &  \# Samples & \# Barcodes & \# Species \\
        \midrule
        {unknown} & {\texttt{pretrain}}      & self- and semi-sup. training    &  4,677,756 & 2,284,232 & ---   \\
        {seen}    & {\texttt{train}}         & supervision; retrieval keys &    289,203 &  118,051 &11,846 \\
                  & {\texttt{val}}           & model dev; retrieval queries &     14,757 &    6,588 & 3,378 \\
                  & {\texttt{test}}          & final eval; retrieval queries&     39,373 &   18,362 & 3,483 \\
        {unseen}  & {\texttt{key\_unseen}}   & retrieval keys              &     36,465 &   12,166 &   914 \\
                  & {\texttt{val\_unseen}}   & model dev; retrieval queries &      8,819 &    2,442 &   903 \\
                  & {\texttt{test\_unseen}}  & final eval; retrieval queries&      7,887 &    3,401 &   880 \\
        {heldout} & {\texttt{other\_heldout}}& novelty detector training  &     76,590 &   41,250 & 9,862 \\
        \bottomrule
    \end{tabular}
    \end{footnotesize}
    }
	\label{tab:split_summary}
\end{table}

\mypara{Splits.} Using the above species sets, we establish partitions for our experiments (\autoref{tab:split_summary}).
The \emph{unknown} samples are all placed into a \texttt{pretrain} split for use in self-supervised pretraining and/or semi-supervised learning. As some DNA barcodes are common to multiple samples, for \emph{seen} and \emph{unseen} records we split the records by placing all samples with the same barcode in the same partition, to ensure there is no repetition of barcodes across splits.
For the closed-world setting, we use the \emph{seen} records to establish \texttt{train}, \texttt{val}, \texttt{test} splits.
To ensure that the \texttt{test} set is not too imbalanced in species distribution, we place samples in the \texttt{test} set with a flattened distribution.
We sample records from species with at least two unique barcodes and eight samples, and the number of samples placed in the \texttt{test} set scales linearly with the total number of samples for the species, until reaching a cap of 25 samples.
We sample 5\% of the remaining \emph{seen} data to form the \texttt{val} partition, but in this case match the imbalance of the overall dataset.
The remaining samples then form the \texttt{train} split, with a final split distribution of 84.2\,:\,4.3\,:\,11.5.
Following standard practice, the \texttt{val} set is for model evaluation during development and hyperparameter tuning, and the \texttt{test} set is for final evaluation.
In the retrieval setting, the \texttt{train} split should additionally be used as a database of \emph{keys} to retrieve over, and the \texttt{val} and \texttt{test} split as queries.
For additional details on the partitioning method and statistical comparisons between the partitions, please see \maybesupplementreflong{s:partitioning-extra}.

For the open-world scenario, we use a similar procedure to establish \texttt{val\_unseen} and \texttt{test\_unseen} over the \emph{unseen} records.
After creating \texttt{test\_unseen} with the same methodology as \texttt{test}, we sample 20\% of remaining \emph{unseen} species records to create \texttt{val\_unseen}. The remaining \emph{unseen} species samples form the \texttt{keys\_unseen} set. 
In the retrieval setting, \texttt{keys\_unseen} is used to form the database of \emph{keys} to retrieve, and the \texttt{val\_unseen} and \texttt{test\_unseen} splits act as queries.
The \emph{heldout} species samples form a final \texttt{other\_heldout} partition. As these species are in neither \emph{seen} nor \emph{unseen}, this split can be used to train a novelty detector without using any \emph{unseen} species.

\subsection{DNA-based taxonomic classification} %
\label{s:exp-dna}
In this section, we demonstrate the utility of the BIOSCAN-5M dataset for DNA-based taxonomic classification. 
Due to their standardized length, DNA barcodes are ideal candidates as input to CNN- and transformer-based architectures for supervised taxonomic classification. However, as noted by~\citet{millan2023barcodebert}, a limitation of this approach is the uncertainty in species-level labels for a substantial portion of the data. This uncertainty, partly due to the lack of consensus among researchers and the continuous discovery of new species, may render supervised learning suboptimal for this task. We address this issue by adopting a semi-supervised learning approach. Specifically, we train a model using self-supervision on unlabelled sequences from the \texttt{pretrain} split and the  \texttt{other\_heldout} split, followed by fine-tuning on sequences from the \texttt{train} split, which includes high-quality labels. The same pretrained model can be used to produce embeddings for sequences from unseen taxa to address tasks in the open-world setting. Consequently, we use these embeddings to perform non-parametric taxonomic classification at a higher (less specific) level in the taxonomic hierarchy for evaluation.

\mypara{Experimental setup.}
Although there has been a growing number of SSL DNA language models proposed in the recent literature, the results obtained by the recently proposed BarcodeBERT \citep{millan2023barcodebert} model empirically demonstrate that training on a dataset of DNA barcodes can outperform more sophisticated training schemes that use a diverse set of non-barcode DNA sequences, such as DNABERT \citep{ji2021dnabert} and DNABERT-2 \citep{zhou2023dnabert}. In this study, we selected BarcodeBERT as our reference model upon which to investigate the impact of pretraining on the larger and more diverse DNA barcode dataset BIOSCAN-5M.  See \autoref{a:exp-dna} for pretraining details.\looseness=-1

We compare our pretrained model against four pretrained transformer models: BarcodeBERT \citep{millan2023barcodebert}, DNABERT-2 \citep{zhou2023dnabert}, DNABERT-S \citep{zhou2024dnabert}, and the nucleotide transformer (NT) \citep{dalla2023nucleotide}; one state space model, HyenaDNA \citep{nguyen2024hyenadna}; and a CNN baseline following the architecture introduced by \citet{Badirli2021-zv}.

As an additional assessment of the impact of BIOSCAN-5M DNA data during pretraining, we use the different pretrained models as feature extractors and evaluate the quality of the embeddings produced by the models on two different SSL evaluation strategies \citep{balestriero2023cookbook}. We first implement genus-level 1-NN probing on sequences from unseen species, providing insights into the models’ abilities to generalize to new taxonomic groups. Finally, we perform species-level classification using a linear classifier trained on embeddings from the pretrained models. Note that for both probing tasks, all the embeddings produced by a single sequence are averaged across the token dimension to generate a token embedding for the barcode.

\mypara{Results.}
We leverage the different partitions of the data and make a distinction between the experiments in the closed-world and open-world settings. In the closed-world setting, the task is species-level identification of samples from species that have been seen during training (Fine-tuned accuracy, Linear probing accuracy). For reference, BLAST \citep{BLAST}, an algorithmic sequence alignment tool, achieves an accuracy of 99.78\% in the task (not included in \autoref{tab:dna-results} as it is not a machine learning model). In fine-tuning, our pretrained model with a 8-4-4 architecture achieves the highest accuracy with 99.28\%, while DNABERT-2 achieves 99.23\%, showing competitive performance. Overall, all models demonstrate strong performance in this task, showcasing the effectiveness of DNA barcodes in species-level identification. For linear probing accuracy, DNABERT-S outperforms others with 95.50\%, followed by our model (8-4-4) with 94.47\%. BarcodeBERT ($k$=4) and DNABERT-S also show strong performance with 91.93\% and 91.59\% respectively (see \autoref{tab:dna-results}). 
\begin{table}[!h]\centering
\caption{\textbf{Performance of DNA-based sequence models} in closed- and open-world settings. For the closed-world, we show the species-level accuracy (\%) for predicting seen species (\texttt{test}), for open-world the genus-level accuracy (\%) for \texttt{test\_unseen} species while using seen species to fit the model. Bold indicates highest accuracy, underlined denotes second highest.
}\label{tab:dna-results}
\resizebox{\linewidth}{!}
{
\begin{tabular}{lllrrrr}\toprule
\textbf{} &\textbf{} &\textbf{} &&\multicolumn{2}{c}{Seen: Species} &Unseen: Genus \\
\cmidrule(lr){5-6} \cmidrule(lr){7-7}
Model &Architecture &SSL-Pretraining & Tokens seen &Fine-tuned &Linear probe &1NN-Probe \\\midrule
CNN baseline &CNN &-- & -- & 97.70 & -- & \underline{29.88}\\
NT &Transformer &Multi-Species & 300\,B&98.99 & 52.41 &21.67 \\
DNABERT-2 &Transformer &Multi-Species& 512\,B &{\bf 99.23} & 67.81 & 17.99 \\
DNABERT-S &Transformer &Multi-Species & $\sim$1,000\,B & 98.99 & \tcf{95.50} &17.70 \\
HyenaDNA &SSM &Human DNA & 5\,B & 98.71 & 54.82& 19.26 \\
BarcodeBERT &Transformer &DNA barcodes & 5\,B & 98.52 & 91.93 & 23.15 \\
Ours (8-4-4) &Transformer &DNA barcodes & 7\,B &{\bf 99.28} & \tcs{94.47} &  {\bf 47.03}\\
\bottomrule
\end{tabular}
}
\end{table}

In the open-world setting, the task is to assign samples from unseen species to seen categories of a coarser taxonomic ranking (1NN-genus probing). In this task, BLAST achieves an accuracy of 58.74\% (not in the table), and our model (8-4-4) performs notably well with an 47.03\% accuracy, which is significantly higher than the other transformer models. The CNN baseline and HyenaDNA show lower accuracies of 29.88\% and 19.26\%, respectively. The use of DNA barcodes for pretraining in our models and BarcodeBERT demonstrates effectiveness in both seen and unseen species classification tasks. One limitation of the comparison is the difference in the dimension of the output space of the different models (128 for HyenaDNA, \vs 512 for NT and 768 for the BERT-based models).
The selection of our model (8-4-4) as the best-performing configuration was done after performing a hyperparameter search to determine the optimal value of $k$ for tokenization, as well as the optimal number of heads and layers in the transformer model. To do that, after pretraining, we fine-tuned the model for species-level identification and performed linear- and 1NN- probing on the \texttt{validation} split (see \autoref{tab:dna-model-search}). We finally note that our pretrained model outperforms BarcodeBERT, the other model trained exclusively trained on DNA barcodes, across all tasks.

\subsection{Zero-shot transfer-learning}
\label{s:exp-zsc}

Recently, \citet{zsc} proposed the task of \emph{zero-shot clustering}, investigating how well unseen datasets can be clustered using embeddings from pretrained feature extractors. \citet{zsc} found that BIOSCAN-1M images were best clustered taxonomically at the family rank while retaining high clustering performance at species and BIN labels. We replicate this analysis using BIOSCAN-5M and extend the modality space to include both image and DNA barcodes.

\mypara{Experimental setup.} 
We follow the experimental setup {of} \citet{zsc}. (1)~Take pretrained encoders; (2)~Extract feature vectors from the stimuli by passing them through an encoder; (3)~Reduce dimensions to 50 using UMAP \citep{mcinnes2018umap}; (4)~Cluster the reduced embeddings with Agglomerative Clustering (L2, Ward's method) \citep{ClusteringBook}; (5) Evaluate against the ground-truth annotations with Adjusted Mutual Information (AMI) score \citep{AMI}, measuring the percentage information explained relative to the entropy of the true labels.

For the image encoders, we consider ResNet-50 \citep{Resnet} and ViT-B \citep{VIT} models, each pretrained on ImageNet-1K \citep{ImageNet} using either cross-entropy supervision (X-ent.), or SSL methods (MAE: \citealp{he2022masked}; VICReg: \citealp{bardes2021vicreg}; DINO-v1: \citealp{caron2021emerging}; MoCo-v3: \citealp{chen2021mocov3}).
We also considered the CLIP \citep{radford2021learning} encoder, which was pretrained on an unspecified, large dataset of captioned images.
To cluster the DNA barcodes, we used recent pretrained models (see \autoref{s:exp-dna} and \appref{s:exp-dna-baseline}), which feature a variety of model architectures, pretraining datasets, and training methodologies:
BarcodeBERT \citep{millan2023barcodebert}, %
DNABERT-2 \citep{zhou2023dnabert},
DNABERT-S \citep{zhou2024dnabert}, %
the nucleotide transformer (NT) \citep{dalla2023nucleotide}, and
HyenaDNA \citep{nguyen2024hyenadna}.

We only cluster samples from the \texttt{test} and \texttt{test\_unseen} splits.
None of the image or DNA pretraining datasets overlap with BIOSCAN-5M, so all samples are ``unseen''.
However, we note that there is a greater domain shift from the image pretraining datasets than the DNA pretraining datasets.

\begin{figure}[tbh]
    \centering
    \includegraphics[width=0.48\linewidth]{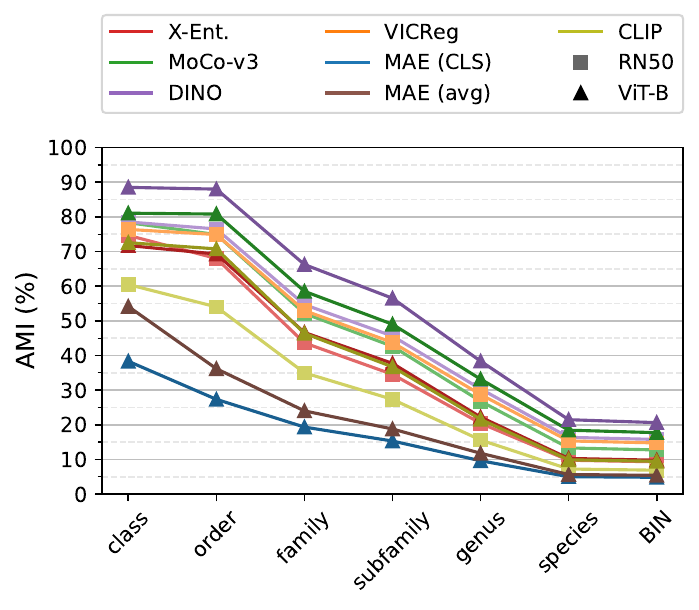}
    \quad
    \includegraphics[width=0.48\linewidth]{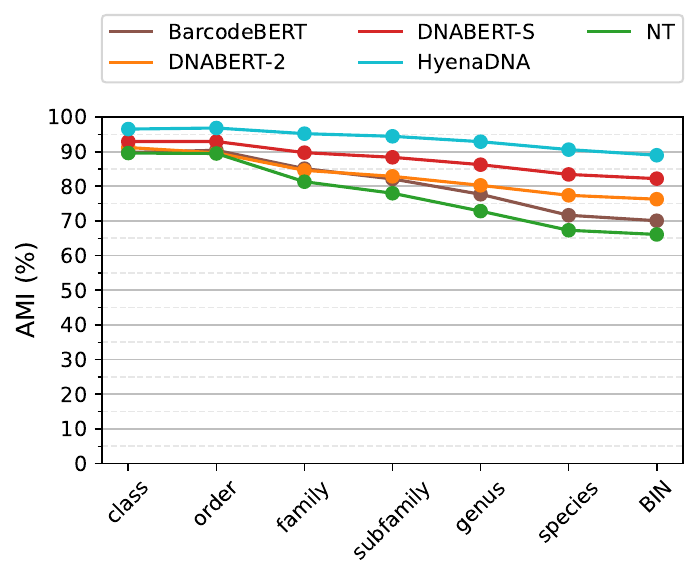}
    \caption{\textbf{Zero-shot clustering AMI (\%) performance} across taxonomic ranks. Left: Image encoders. Right: DNA encoders.}
    \label{fig:zsc-taxonomic-breakdown}
    \vspace{-5mm}
\end{figure}

\mypara{Results.}
Similar to~\citet{zsc}, we find (\autoref{fig:zsc-taxonomic-breakdown}) image clusterings agree with the taxonomic labels at coarse ranks (order: 88\%), but agreement decreases progressively at finer-grained ranks (species: 21\%); the best-performing image encoder was DINO, followed by other SSL methods VICReg and MoCo-v3, with (larger) ViT-B models outperforming ResNet-50 models.
We found the performance of the DNA encoders exceeded that of the image encoders across all taxonomic levels, with higher performance at coarse ranks (order: 97\%) and much shallower decline as granularity becomes finer (species: 91\%).
HyenaDNA provided the best performance, with 90\% agreement between its clusterings and both the GT species and DNA BIN annotations.
These results suggest that DNA barcodes are highly informative about species identity (which is unsurprising as it is the reason this barcode is used), and unseen samples can be readily grouped together using off-the-shelf DNA models.

We also considered the zero-shot clustering of the concatenated image and DNA representations, detailed in \appref{crossmodalclustering}.
Due to the high performance of the DNA features, adding image features to the embeddings decreased the performance compared to using DNA embeddings alone.
For additional details and analysis, see \autoref{a:exp-zsc}.

\subsection{Multimodal retrieval learning}
\label{s:exp-clip}

Lastly, we demonstrate the importance of a multimodal dataset through alignment of image, DNA, and taxonomic label embeddings {using \clibd~\citep{gong2024bioscanclip}} to improve taxonomic classification. By learning a shared embedding space across modalities, we can query between modalities and leverage the information across them to achieve better performance in downstream tasks. We are able to incorporate a diversity of samples into training toward taxonomic classification, even with incomplete taxonomic labels.

\mypara{Experimental setup.}
We follow the model architecture and experimental setup of {\clibd~\citep{gong2024bioscanclip}}.  
We start with pretrained encoders for each modality and 
{perform full-tuning}
with NT-Xent loss \citep{sohn2016improved}.  Our image encoder is a ViT-B \citep{VIT} pretrained on ImageNet-21k and fine-tuned on ImageNet-1k \citep{deng2009imagenet}. 
For DNA barcodes, we use BarcodeBERT \citep{millan2023barcodebert} with 5-mer tokenization, pretrained on 893~k DNA barcodes from the Barcode of Life Data system (BOLD)
\citep{bold_systems}, and for text, we use BERT-small \citep{turc2019well}.  
We train on our \texttt{pretrain} and \texttt{train} splits using the Adam \citep{kingma2014adam} optimizer for 20 epochs until convergence with a learning rate of 1\text{e}{-6}, batch size 2000.  Training took 29 hours on four 80GB A100 GPUs.
To evaluate the performance of our models, we report micro (see \autoref{a:exp-clip}) and macro (see \autoref{tab:bioscan_clip_results_macro}) top-1 accuracy for taxonomic classification at different levels.  To determine the taxonomic labels for a new query, we encode the sample image or DNA and find the closest matching embedding in a set of labelled samples (keys).  For efficient lookup, we use FAISS \citep{johnson2019billion} with exact search ({\small \texttt{IndexFlatIP}}).

We compared the performance for the initial pretrained (unimodal) encoders to our models fine-tuned on either the full \texttt{pretrain} and \texttt{train} partitions from BIOSCAN-5M, or on a random 1 million sample subset of these partitions.
The 1M image subset contained 20\% of the images, 27\% of the barcodes, and 47\% of the BINs of the 5\,M image training dataset.
We evaluated these using image-to-image, DNA-to-DNA, and image-to-DNA embeddings as queries and keys.

\mypara{Results.} We compare \clibd trained on {the full BIOSCAN-5M training set against models trained on a randomly selected subset of 1 million records} and the initial pretrained encoders before multimodal contrastive learning.  Our results, shown in \autoref{tab:bioscan_clip_results_macro}, demonstrate that our {full} model improves classification accuracy for same-modality queries and enables cross-modality queries. 
By aligning to DNA, our image embeddings are able to capture finer details. {We likewise see improvements in alignment among DNA embeddings.} 
{Additionally, we observe that increasing the training dataset size from 1 million to 5 million records  leads to better models with more accurate results across all studied taxa for both image and DNA modalities, indicating there are still benefits from dataset scale at this size.
By including the text modality, 
we further improve accuracy at the higher taxa levels. Interestingly, including the text modality results in slightly lower performance at the species level.  
This is likely due to the sparse availability of species labels in the training data, as only 9\% of records having species labels.}
For additional details and analysis, see \autoref{a:exp-clip}.

\begin{table}[tb]
\centering
\caption{
\textbf{Multimodal retrieval top-1 {macro} accuracy} (\%) on the test set for using different {amount of pre-training data (1 million vs 5 million records from BIOSCAN-5M)} and different combinations of aligned embeddings (image, DNA, text) during contrastive training. We show results for using image-to-image, DNA-to-DNA, and image-to-DNA query and key combinations. As a baseline, we show the results prior to contrastive learning (no alignment). We report the accuracy for seen and unseen species, and the harmonic mean (H.M.) between these (bold: highest acc.).
}
\resizebox{\linewidth}{!}
{
\begin{tabular}{@{}ll ccc rrr rrr rrr@{}}
\toprule
& & \multicolumn{3}{c}{Aligned embeddings} & \multicolumn{3}{c}{DNA-to-DNA} & \multicolumn{3}{c}{Image-to-Image} & \multicolumn{3}{c}{Image-to-DNA}\\
\cmidrule(){3-5} \cmidrule(l){6-8} \cmidrule(l){9-11}
\cmidrule(l){12-14}
Taxon & \# Records & Img & DNA & Txt & ~~Seen & Unseen & H.M. & ~~Seen & Unseen & H.M. & ~~Seen & Unseen & H.M. \\
\midrule
Order & --- & \myxmark & \myxmark & \myxmark  & 95.8 & 97.8 & 96.8 & 78.1 & 82.4 & 80.2 & 3.6 & 6.3 & 4.6 \\
 & 1M & \checkmark & \checkmark & \checkmark & \textbf{100.0} & \textbf{100.0} & \textbf{100.0} & 93.5 & 95.6 & 94.5 & 86.5 & 95.4 & 90.7 \\
 & 5M & \checkmark & \checkmark & \myxmark & \textbf{100.0} & \textbf{100.0} & \textbf{100.0} & \textbf{95.3} & 96.2 & 95.7 & \textbf{92.2} & 97.2 & \textbf{94.7} \\
 & 5M & \checkmark & \checkmark & \checkmark & \textbf{100.0} & \textbf{100.0} & \textbf{100.0} & 95.1 & \textbf{98.5} & \textbf{96.8} & 91.2 & \textbf{98.0} & 94.4 \\
\midrule
Family & --- & \myxmark & \myxmark & \myxmark  & 90.2 & 92.1 & 91.2 & 52.3 & 55.5 & 53.8 & 0.3 & 1.0 & 0.4\\
 & 1M & \checkmark & \checkmark & \checkmark & 98.3 & 99.3 & 98.8 & 86.8 & 89.9 & 88.3 & 65.8 & 73.7 & 69.5 \\
 & 5M & \checkmark & \checkmark & \myxmark & 99.4 & \textbf{100.0} & \textbf{99.7} & 91.0 & 92.7 & 91.8 & 80.5 & 83.6 & 82.0 \\
 & 5M & \checkmark & \checkmark & \checkmark & \textbf{99.5} & \textbf{100.0} & \textbf{99.7} & \textbf{91.7} & \textbf{94.2} & \textbf{93.0} & \textbf{80.9} & \textbf{84.6} & \textbf{82.7} \\
\midrule
Genus & --- & \myxmark & \myxmark & \myxmark  & 86.8 & 85.7 & 86.2 & 34.0 & 31.9 & 32.9 & 0.0 & 0.0 & 0.0 \\
 & 1M & \checkmark & \checkmark & \checkmark & 98.0 & 97.2 & 97.6 & 76.5 & 75.6 & 76.1 & 46.2 & 36.2 & 40.6 \\
 & 5M & \checkmark & \checkmark & \myxmark & \textbf{99.0} & 99.3 & \textbf{99.2} & 83.3 & 85.5 & 84.4 & \textbf{64.4} & 50.4 & \textbf{56.6} \\
 & 5M & \checkmark & \checkmark & \checkmark & 98.8 & \textbf{99.5} & \textbf{99.2} & \textbf{84.0} & \textbf{86.0} & \textbf{85.0} & 63.0 & \textbf{50.6} & 56.1 \\
\midrule
Species & --- & \myxmark & \myxmark & \myxmark  & 84.6 & 75.6 & 79.8 & 24.2 & 12.6 & 16.6 & 0.0 & 0.0 & 0.0 \\
  & 1M & \checkmark & \checkmark & \checkmark & 96.7 & 91.7 & 94.1 & 66.6 & 49.6 & 56.8 & 34.9 & 6.8 & 11.3 \\
 & 5M & \checkmark & \checkmark & \myxmark & \textbf{98.1} & 95.8 & \textbf{97.0} & 75.9 & \textbf{60.8} & \textbf{67.5} & \textbf{54.4} & \textbf{13.8} & \textbf{22.0} \\
 & 5M & \checkmark & \checkmark & \checkmark & 98.0 & \textbf{95.9} & \textbf{97.0} & \textbf{76.0} & 60.1 & 67.1 & 51.1 & 12.7 & 20.3 \\
\bottomrule
\end{tabular}
}
\label{tab:bioscan_clip_results_macro}
\vspace{-0mm}
\end{table}

\section{Conclusion}
\label{sec:conclusion}

We present the BIOSCAN-5M dataset, a valuable resource for the machine learning community containing over 5 million arthropod specimens. To highlight the dataset’s multimodal capabilities, we conducted three benchmark experiments that leverage images, DNA barcodes, and textual taxonomic annotations for fine-grained taxonomic classification and zero-shot clustering.

{An open problem for biodiversity monitoring systems is handling novel species.
To facilitate research in this space, our dataset includes partitions for both closed-world and open-world settings.
Furthermore, we provide three distinct benchmark tasks, each evaluated down to species-level, demonstrating the real-world applicability of BIOSCAN-5M’s multimodal features. These tasks include fine-grained taxonomic classification using DNA sequences, multimodal classification combining DNA, images, and taxonomic labels, and clustering of learned DNA and image embeddings.}

We believe that the BIOSCAN-5M dataset will serve as a catalyst for further machine learning research in biodiversity, fostering innovations that can enhance our understanding and preservation of the natural world. By providing a curated multi-modal resource, we aim to support further initiatives in the spirit of TreeOfLife-10M~\citep{stevens2023bioclip} and contribute to the broader goal of mapping and preserving global biodiversity. This dataset not only facilitates advanced computational approaches but also underscores the crucial intersection between technology and conservation science, driving forward efforts to protect our planet's diverse ecosystems for future generations.

\begin{ack}
\label{sec:acknowledgement}
We acknowledge the support of the Government of Canada’s New Frontiers in Research Fund (NFRF), [NFRFT-2020-00073].
This research is also supported by an NVIDIA Academic Grant.
This research was enabled in part by support provided by \hreffoot{https://calculquebec.ca}{Calcul Québec} and the \hreffoot{https://alliancecan.ca}{Digital Research Alliance of Canada}.
Resources used in preparing this research were provided, in part, by the Province of Ontario, the Government of Canada through CIFAR, and \hreffoot{https://vectorinstitute.ai/partnerships/current-partners/}{companies sponsoring} the Vector Institute.
Data collection was enabled by funds from the Walder Foundation, a New Frontiers in Research Fund (NFRF) Transformation grant, a Canada Foundation for Innovation's (CFI) Major Science Initiatives (MSI) Fund and CFREF funds to the Food from Thought program at the University of Guelph.
The authors also wish to acknowledge the team at the Centre for Biodiversity Genomics responsible for preparing, imaging, and sequencing specimens used for this study.  We also thank Mrinal Goshalia for assistance with the cropping tool and annotation of images.
\end{ack}

\iftoggle{arxiv}{%
\section*{Author contributions}

DS provided the original dataset and provided guidance on subsequent processing steps.
ZG curated the dataset, created primary metadata file and removed invalid images. 
ZG and SCL implemented data analytics and statistical processing pipelines. 
ZG calculated the specimen size information.
ZMG improved the image cropping tool and conducted experiments for cropping images.
SCL cleaned inconsistent taxonomic labels, with assistance from ZG, and they finalized the metadata file.
SCL partitioned the data, with assistance from ZMG.
PMA and SCL conducted the DNA-based taxonomic classification experiments.
SCL conducted the zero-shot transfer-learning clustering experiments, with assistance from PMA and NP.
ZMG and ATW conducted the multimodal retrieval learning experiments.
ZG and AXC packaged and released the dataset.
ZG, SCL, ZMG, PMA, NP, ATW, JBH, and IZ authored the manuscript text and figures.
AXC, PF, GWT, LK, JBH, and SCL provided guidance on experimental design.
All authors reviewed the manuscript.

}{}

\bibliographystyle{icml2024}  %
\bibliography{main}

\clearpage
\clearpage
\appendix
\section*{\Large Appendices}\label{sec:appendices}

In these appendices, we provide additional details about experimental designs; dataset construction,  preprocessing, partitioning, and distribution; and high-level overviews dataset.
The appendices are summarized below.

\begin{itemize}
\item \autoref{a:exp-dna}. Additional Experiments for \textit{DNA-based Taxonomic Classification} (extends \autoref{s:exp-dna}): pretraining details, architecture search, and linear probe training.

\item \autoref{a:exp-zsc}. Additional Experiments for \textit{Zero-Shot Clustering} (extends \autoref{s:exp-zsc}): clustering with deduplicated barcodes, and cross-modal clustering.

\item \autoref{a:exp-clip}. Additional Experiments for \textit{Multi-Modal Learning} (extends \autoref{s:exp-clip}): model training and inference details, top-1 micro accuracy, retrieval examples.

\iftoggle{arxiv}{%

\item \autoref{a:dataset}. Dataset summary.

\item \autoref{a:ethics}. Ethics and responsible use.

\item \autoref{a:availability}. Dataset availability and maintenance.

\item \autoref{a:licensing}. Licensing details.

\item \autoref{a:images}. RGB image cropping, preprocessing, and packaging.

\item \autoref{a:metadata}. Metadata format.

\item \autoref{a:comparison-BIOSCAN-1M}. Comparison between BIOSCAN-5M and BIOSCAN-1M.

\item \autoref{a:focus}. Focus and objectives discussion.

\item \autoref{s:dataset-stats}. Dataset feature statistics.

\item \autoref{a:category-distribution}. Taxonomic category distribution.

\item \autoref{a:barcode-statistics}. DNA barcode statistics.

\item \autoref{a:non-insect}. Insect vs non-insect organisms.

\item \autoref{a:limitations}. Limitations and challenges.

\item \autoref{a:data-processing}. Data processing: image processing, standardization of unassigned taxa, and taxonomic label cleaning.

\item \autoref{s:partitioning-extra}. Dataset partitioning (extends \autoref{sec:partitioning}): partitioning of species sets, data splits, and distributional splits between them.

}{%
}

\end{itemize}

\FloatBarrier
\section{DNA-based Taxonomic Classification --- Additional Experiments}
\label{a:exp-dna}

\begin{figure*}[tbh]
	\centering    
	\includegraphics[width=1\textwidth]{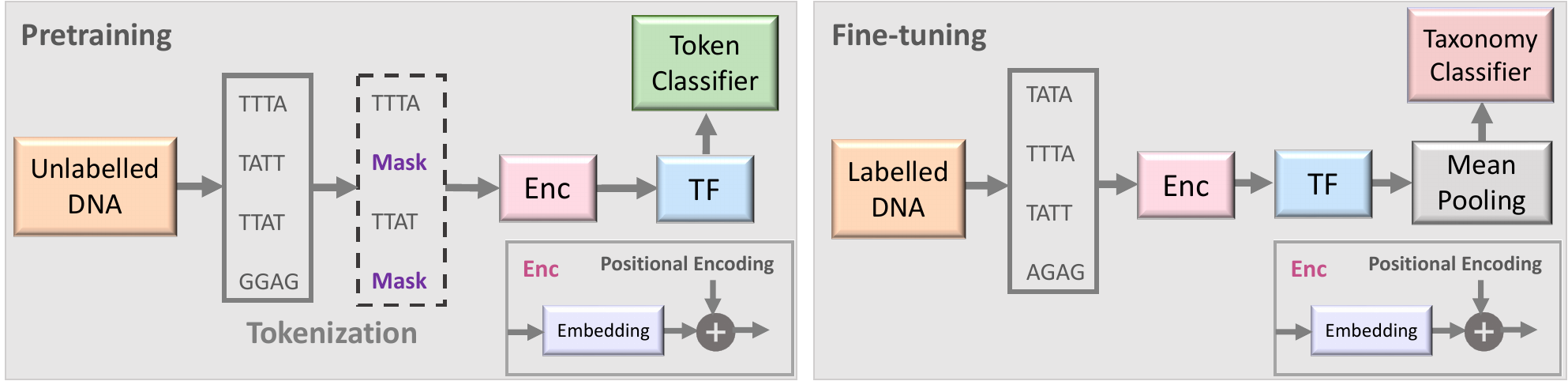}
	\caption{\textbf{DNA-based taxonomic classification methodology.} Two stages of the proposed semi-supervised learning set-up based on BarcodeBERT \citep{millan2023barcodebert}. (1) Pretraining: DNA sequences are tokenized using non-overlapping $k$-mers and 50\% of the tokens are masked for the MLM task. Tokens are encoded and fed into a transformer model. The output embeddings are used for token-level classification. (2) Fine-tuning: All DNA sequences in a dataset are tokenized using non-overlapping $k$-mer tokenization and all tokenized sequences, without masking, are passed through the pretrained transformer model.  Global mean-pooling is applied over the token-level embeddings and the output is used for taxonomic classification.}
	\label{fig:barcode_bert}
\end{figure*}

As described in the main text (\autoref{s:exp-dna}), we leverage all data splits in the BIOSCAN-5M dataset by adopting a semi-supervised learning approach. Specifically, we train a model using self-supervision on the unlabelled partition of the data, followed by fine-tuning on the train split. Our experimental setup is illustrated in \autoref{fig:barcode_bert}.

\subsection{Pretraining details}
We pretrain the model on the 2,283,900 unique DNA sequences from the \texttt{pretrained} partition and the 41,232 unique sequences from the \texttt{other\_heldout} partition, totalling 2,325,132 pretraining DNA samples. For all samples, trailing \verb|N| characters are removed and all sequences are truncated at 660 nucleotides. Note that leading \verb|N| characters are retained since they are likely to correspond to true unknown nucleotides in the barcode. The model was pretrained using the same MLM loss function and training configurations as in BarcodeBERT \citep{millan2023barcodebert}. Specifically, we use a non-overlapping $k$-mer-based tokenizer and a transformer model with 12 transformer layers, each having 12 attention heads. However, we included a random offset of at most $k$ nucleotides to each sequence as a data augmentation technique to enhance the sample efficiency. We use a learning rate of $2\times10^{-4}$, a batch size of 128, a OneCycle scheduler \citep{onecycle}, and the AdamW optimizer \citep{adamw}, training the model for 35 epochs. In addition to using the architecture reported in BarcodeBERT, we performed a parameter search to determine the optimal $k$-mer tokenization length and model size, parameterized by the number of layers and heads in the transformer model, in order to identify an optimal architecture configuration. After pretraining, we fine-tuned the model with cross-entropy supervision for species-level classification. The pre-training stage takes approximately 50 hours using four Nvidia A40 GPUs and the fine-tuning stage of the 4-12-12 models takes 2.5 hours in four Nvidia A40 GPUs.

\subsection{Baseline Models}
\label{s:exp-dna-baseline}
There has been a growing number of SSL DNA language models proposed in recent literature, most of which are based on the transformer architecture and trained using the MLM objective \citep{ji2021dnabert,zhou2023dnabert,zhou2024dnabert}. These models differ in the details of their model architecture, tokenization strategies, and training data but the underlying principles remain somewhat constant. An exception to this trend is the HyenaDNA \citep{nguyen2024hyenadna} model, which stands out by its use of a state space model (SSM) based on the Hyena architecture \citep{poli2023hyena} and trained for next token prediction. For evaluation, we utilized the respective pre-trained models from Huggingface ModelHub, specifically:

\begin{itemize}
\item DNABERT-2: \href{https://huggingface.co/zhihan1996/DNABERT-2-117M}{zhihan1996/DNABERT-2-117M}
\item DNABERT-S: \href{https://huggingface.co/zhihan1996/DNABERT-S}{zhihan1996/DNABERT-S}
\item NT: \href{https://huggingface.co/InstaDeepAI/nucleotide-transformer-v2-50m-multi-species}{InstaDeepAI/nucleotide-transformer-v2-50m-multi-species}
\item HyenaDNA: \href{https://huggingface.co/LongSafari/hyenadna-tiny-1k-seqlen}{LongSafari/hyenadna-tiny-1k-seqlen}
\end{itemize}
The BarcodeBERT implementation was taken from \url{https://github.com/Kari-Genomics-Lab/BarcodeBERT}.
All the models, including our pretrained models, were fine-tuned for 35 epochs with a batch size of 32 or 128 and a learning rate of $1\times10^{-4}$ per 64 samples in the batch with the OneCycle LR schedule \citep{onecycle}.

\subsection{Linear probe training}
A linear classifier is applied to the embeddings generated by all the pretrained models for species-level classification. The parameters of the model are learned using stochastic gradient descent with a constant learning rate of 0.01, momentum $\mu=0.9$ and weight $\lambda=1\times10^{-5}$.

For the hyperparameter search, shown in \autoref{tab:dna-model-search}, our linear probe is performed using the same methodology as the fine-tuning stage, except the encoder parameters are frozen.

\begin{table}[tbh]\centering
\caption{\textbf{Architecture search for DNA barcode encoder.} Search over the space of $k$-mer tokenization length and transformer architectures (number of layers and heads).
For fine-tuned and linear probe, we show the class-balanced accuracy (\%) on the closed-world \texttt{val} partition, and for 1-NN probe, we show the class-balanced accuracy on the \texttt{val\_unseen} partition. Bold: architecture with highest accuracy for the row. Underlined: second highest accuracy.}
\label{tab:dna-model-search}
\resizebox{\columnwidth}{!}{%
\begin{tabular}{lrrrrrrrrrrrr}
\toprule
& \multicolumn{4}{c}{4 layers, 4 heads}& \multicolumn{4}{c}{6 layers, 6 heads}& \multicolumn{4}{c}{12 layers, 12 heads} \\
\cmidrule(lr){2-5}\cmidrule(lr){6-9}\cmidrule(lr){10-13}
Evaluation   &{$k\!=\!2$} &{$k\!=\!4$} &{$k\!=\!6$} &{$k\!=\!8$} &{$k\!=\!2$} &{$k\!=\!4$} &{$k\!=\!6$} &{$k\!=\!8$} &{$k\!=\!2$} &{$k\!=\!4$} &{$k\!=\!6$} &{$k\!=\!8$} \\
\midrule
Fine-tuned   &93.8 &97.8 &\tcs{98.7} &\tcf{98.9} &92.4 &97.9 &49.4 &\tcs{98.7} &93.8 &98.1 & 0.0 & 0.0 \\
Linear probe &32.2 &\tcs{79.8} &76.4 &\tcf{97.1} &34.3 &58.9 & 8.9 &\tcs{79.7} &16.4 & 3.2 & 0.0 & 0.0 \\
1-NN         &43.1 &\tcf{50.7} &35.0 &\tcs{46.4} &46.2 &37.2 &23.4 &37.9 &29.1 &28.3 & 0.0 & 0.1 \\
\bottomrule
\end{tabular}
}
\end{table}

\FloatBarrier
\section{Zero-Shot Clustering --- Additional Experiments}
\label{a:exp-zsc}

\begin{figure}[thb]
    \centering    
    \includegraphics[width=\textwidth]{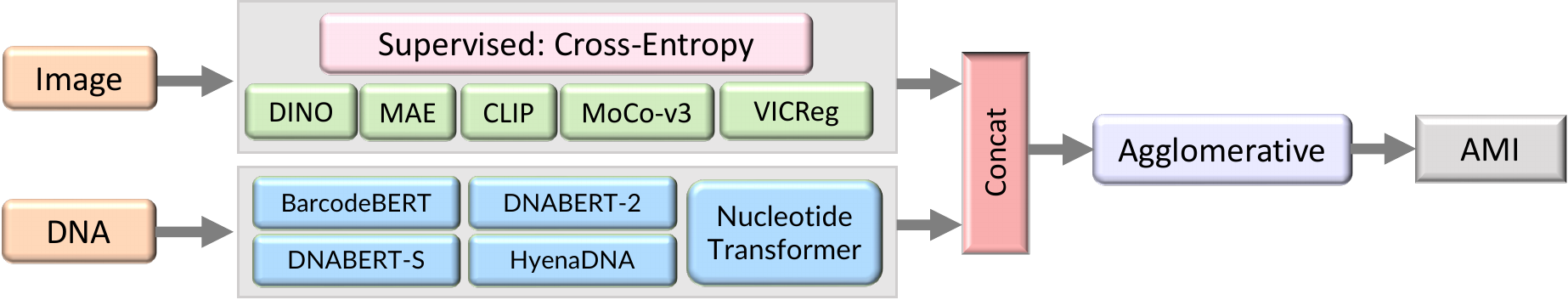}
    \caption{\textbf{Zero-shot clustering methodology.}
    Images and DNA are each passed through one of several pretrained encoders.
    These representations are clustered with Agglomerative Clustering.}
    \label{fig:zsc-method}
\end{figure}

As described in \autoref{s:exp-zsc}, we performed a series of zero-shot clustering experiments to establish how pretrained image and DNA models could handle the challenge of grouping together repeat observations of novel/unseen species.
Our methodology is illustrated in \autoref{fig:zsc-method}.

\subsection{Experiment resources}

All zero-shot clustering experiments were performed on a compute cluster with the job utilizing two CPU cores (2x Intel Xeon Gold 6148 CPU @ 2.40GHz) and no more than 20\,GB of RAM.
The typical runtime per experiment was around 4.5 hours.

\subsection{Accounting for Duplicated DNA Barcode Sequences}

In our main experiments, we found that the performance of DNA-embedding clusterings greatly outperformed that of image-embeddings.
However, it is worth considering that there are fewer unique DNA barcodes than images.
The mean number of samples per barcode is around two.
This provides clustering methods using DNA with an immediate advantage as some stimuli compare as equal and are trivially grouped together, irrespective of the encoder.

\begin{figure}[thb]
    \centering
    \includegraphics[width=0.47\linewidth]{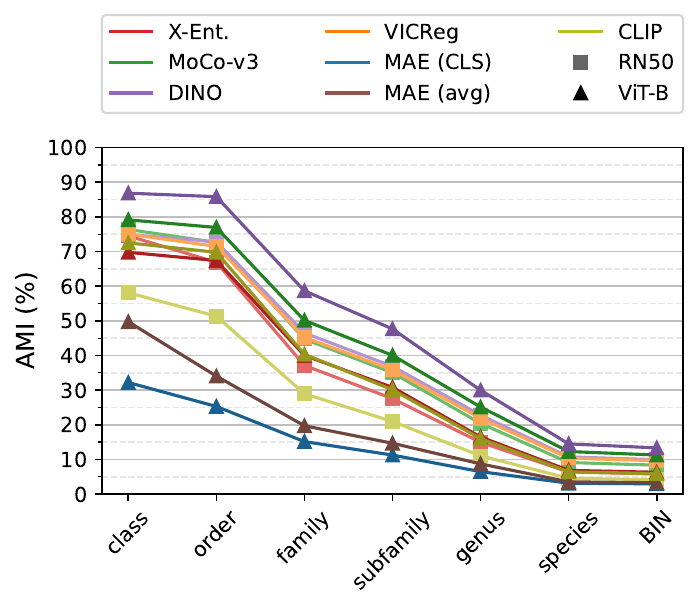}
    \quad
    \includegraphics[width=0.47\linewidth]{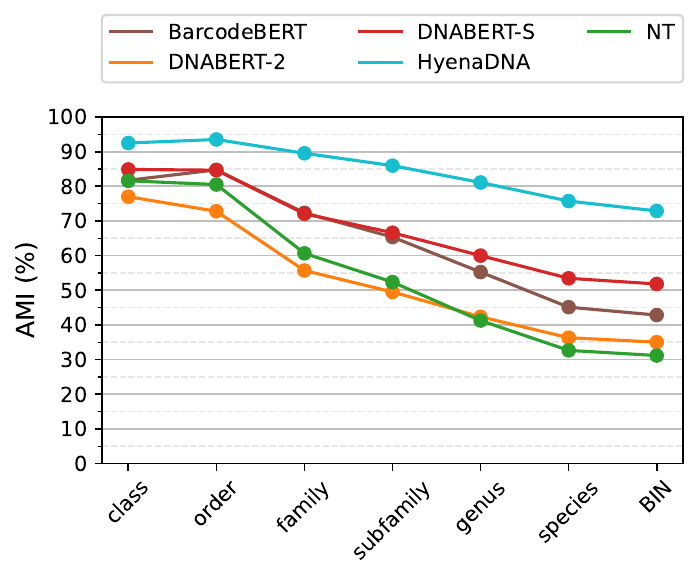}
    \caption{\textbf{Zero-shot clustering} AMI (\%) performance across taxonomic ranks on \texttt{test} and \texttt{test\_unseen} data, with \textbf{one sample per barcode}.}
    \label{fig:zsc-taxonomic-breakdown-dedup}
\end{figure}

To account for this, we repeated our analysis with only one sample per barcode.
Our results, shown in \autoref{fig:zsc-taxonomic-breakdown-dedup}, indicate that both image- and DNA-based clusterings are reduced in performance when the number of samples per barcode is reduced to one.
This is explained in part by the fact that many species will be reduced to a single observation, which is challenging for clusterers to handle.
We found that the performance of most DNA encoders fell by more than the image encoders when the number of samples per barcode was reduced to one.
However, the DNA embeddings still produce clusterings in better agreement with the taxonomic labels than the image embeddings.
In particular, the best-performing DNA encoder, HyenaDNA, still attained 75\% agreement with the ground truth labels at the species-level clustering.

\begin{table}[thb]
\caption{\textbf{Cross-modal zero-shot clustering} AMI (\%) performance, on \texttt{test} and \texttt{test\_unseen} data, with \textbf{one sample per barcode}.}
\label{tab:zsc-cross}
\centering
\small
\label{tab:AMI:AgglomerativeClustering}
\begin{tabular}{llrrrrrr}
\toprule
&             & \multicolumn{1}{c}{} & \multicolumn{5}{c}{DNA encoder}\\
\cmidrule(l){4-8}
Architecture & Image encoder& \rotatebox{90}{\textit{Image-only}}& \rotatebox{90}{  BarcodeBERT  }& \rotatebox{90}{   DNABERT-2   }& \rotatebox{90}{   DNABERT-S   }& \rotatebox{90}{   HyenaDNA    }& \rotatebox{90}{      NT       }\\
\midrule
---
& \textit{DNA-only}       &   --   &${  47} $ &${  52} $ &$\tcs{  63} $ &$\tcf{  81} $ &${  36} $ \\
\midrule
ResNet-50
& X-Ent.                  &$   5  $ &$  30  $ &$  26  $ &${  32 }$ &$   9  $ &$  12  $ \\
& MoCo-v3                 &$   8  $ &$  29  $ &$  23  $ &$  27  $ &$  11  $ &$  11  $ \\
& DINO                    &$\tcg{11}$ &${  31 }$ &${  28 }$ &${  31 }$ &${  15 }$ &${  14 }$ \\
& VICReg                  &$  10  $ &$  30  $ &$  26  $ &$  30  $ &$  13  $ &$  13  $ \\
& CLIP                    &$   6  $ &$  25  $ &$  21  $ &$  25  $ &$   9  $ &$   9  $ \\
\midrule
ViT-B
& X-Ent.                  &$   7  $ &$  33  $ &$  35  $ &$  42  $ &$  13  $ &$  14  $ \\
& MoCo-v3                 &$\tcs{13}$ &$  38  $ &$  43  $ &$  49  $ &$  21  $ &$  20  $ \\
& DINO                    &$\tcf{15}$ &${{  38}}$ &${{  45}}$ &$\tcs{{  51}}$ &${{  23}}$ &${{  21}}$ \\
& MAE (CLS)               &$   5  $ &$  33  $ &$  33  $ &$  40  $ &$  10  $ &$  13  $ \\
& MAE (avg)               &$   3  $ &$  29  $ &$  26  $ &$  32  $ &$   7  $ &$   9  $ \\
& CLIP                    &$   7  $ &$  34  $ &$  37  $ &$  44  $ &$  14  $ &$  16  $ \\
\bottomrule
\end{tabular}
\end{table}

\subsection{Cross-modal embedding clustering}
\label{crossmodalclustering}

We additionally considered the effect of clustering the embeddings from both modalities at once, achieved by concatenating an image embedding and a DNA embedding to create a longer feature vector per sample.
As shown in \autoref{tab:zsc-cross}, we find that combining image features with DNA features results in a worse performance at species-level clustering.

In preliminary experiments (not shown) we found that the magnitude of the vectors greatly impacted the performance, as large image embeddings would dominate DNA embeddings with a smaller magnitude. We considered standardizing the embeddings before concatenation with several methods (L2-norm, element-wise z-score, average z-score) and found element-wise z-score gave the best performance, a step which we include in these results.
Even with this, the performance falls when we add image embeddings to the DNA embeddings.
We note that the best DNA-only encoder, HyenaDNA, has the largest drop in performance, which we hypothesize is because it has the shortest embedding dimensions of 128-d compared with NT (512-d) and the BERT-based models (768-d).

\FloatBarrier
\section{Multi-Modal Learning --- Additional Experiments}
\label{a:exp-clip}

\begin{figure}[thb]
	\centering    
	\includegraphics[width=1.\textwidth]{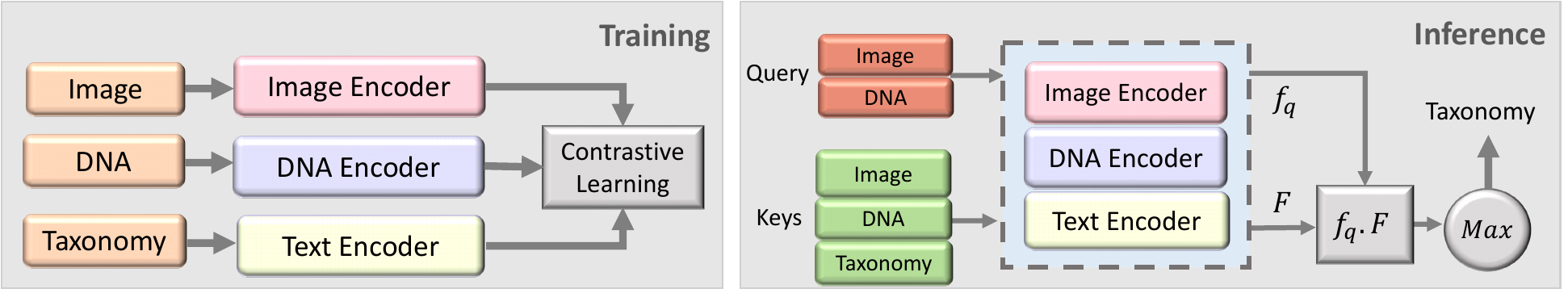}
	\caption{\textbf{Multi-modal learning methodology.} Our experiments using \clibd~\citep{gong2024bioscanclip} are conducted in two steps. (1)~Training: Multiple modalities, including RGB images, textual taxonomy, and DNA sequences, are encoded separately, and trained using a contrastive loss function. (2)~Inference: Image vs DNA embedding is used as a query, and compared to the embeddings obtained from a database of image, DNA and text (keys). The cosine similarity is used to find the closest key embedding, and the corresponding taxonomic label is used to classify the query.}
	\label{fig:bioscan_clip}
\end{figure}

As described in \autoref{s:exp-clip}, we trained a multimodal model with an aligned embedding space across the images, DNA, and taxonomic labels.
Our methodology is illustrated in \autoref{fig:bioscan_clip}.

\begin{table}[thb]
\centering
\caption{
\textbf{Multimodal retrieval top-1 \emph{micro} accuracy} (\%) on the test set for using {different amount of training data (1 million vs 5 million records from BIOSCAN-5M)} and different combinations of aligned embeddings (image, DNA, text) during contrastive training. We show results for using image-to-image, DNA-to-DNA, and image-to-DNA query and key combinations. As a baseline, we show the results prior to contrastive learning (no alignment). We report the accuracy for seen and unseen species, and the harmonic mean (H.M.) between these (bold: highest acc.).
}
\resizebox{\linewidth}{!}
{
\begin{tabular}{@{}ll ccc rrr rrr rrr@{}}
\toprule
& & \multicolumn{3}{c}{Aligned embeddings} & \multicolumn{3}{c}{DNA-to-DNA} & \multicolumn{3}{c}{Image-to-Image} & \multicolumn{3}{c}{Image-to-DNA}\\
\cmidrule(){3-5} \cmidrule(l){6-8} \cmidrule(l){9-11}
\cmidrule(l){12-14}
Taxon & \# Records & Img & DNA & Txt & ~~Seen & Unseen & H.M. & ~~Seen & Unseen & H.M. & ~~Seen & Unseen & H.M. \\
\midrule
Order & --- & \myxmark & \myxmark & \myxmark  & 98.9 & 99.3 & 99.1 & 94.2 & 97.0 & 95.6 & 18.3 & 14.7 & 16.3 \\
 & 1M & \checkmark & \checkmark & \checkmark & \textbf{100.0} & \textbf{100.0} & \textbf{100.0} & 99.3 & 99.6 & 99.5 & 98.7 & 99.2 & 98.9 \\
 & 5M & \checkmark & \checkmark & \myxmark & \textbf{100.0} & \textbf{100.0} & \textbf{100.0} & \textbf{99.5} & \textbf{99.7} & \textbf{99.6} & \textbf{99.4} & 99.5 & \textbf{99.5} \\
 & 5M & \checkmark & \checkmark & \checkmark & \textbf{100.0} & \textbf{100.0} & \textbf{100.0} & \textbf{99.5} & \textbf{99.7} & \textbf{99.6} & 99.3 & \textbf{99.6} & \textbf{99.5} \\
\midrule
Family & --- & \myxmark & \myxmark & \myxmark  & 96.5 & 97.3 & 96.9 & 72.9 & 76.0 & 74.4 & 1.7 & 1.9 & 1.8 \\
 & 1M & \checkmark & \checkmark & \checkmark & 99.8 & 99.8 & 99.8 & 95.5 & 96.8 & 96.2 & 90.6 & 89.1 & 89.9 \\
 & 5M & \checkmark & \checkmark & \myxmark & \textbf{99.9} & \textbf{100.0} & 99.9 & 96.8 & 97.9 & 97.4 & 94.0 & 93.1 & 93.5 \\
 & 5M & \checkmark & \checkmark & \checkmark & \textbf{99.9} & \textbf{100.0} & \textbf{100.0} & \textbf{97.0} & \textbf{98.3} & \textbf{97.7} & \textbf{94.6} & \textbf{94.4} & \textbf{94.5} \\
\midrule
Genus & --- & \myxmark & \myxmark & \myxmark & 94.0 & 93.5 & 93.7 & 47.8 & 47.0 & 47.4 & 0.2 & 0.0 & 0.1  \\
 & 1M & \checkmark & \checkmark & \checkmark & 99.3 & 98.8 & 99.0 & 86.0 & 85.9 & 86.0 & 68.1 & 52.3 & 59.2 \\
 & 5M & \checkmark & \checkmark & \myxmark & \textbf{99.6} & \textbf{99.8} & \textbf{99.7} & 90.6 & 91.6 & 91.1 & \textbf{79.5} & 65.0 & 71.5 \\
 & 5M & \checkmark & \checkmark & \checkmark & \textbf{99.6} & \textbf{99.8} & \textbf{99.7} & \textbf{91.0} & \textbf{92.1} & \textbf{91.5} & 79.3 & \textbf{66.3} & \textbf{72.2} \\
\midrule
Species & --- & \myxmark & \myxmark & \myxmark  & 91.6 & 84.8 & 88.1 & 31.9 & 19.1 & 23.9 & 0.0 & 0.0 & 0.0 \\
 & 1M & \checkmark & \checkmark & \checkmark & 98.3 & 95.0 & 96.6 & 75.1 & 57.5 & 65.1 & 47.9 & 10.4 & 17.0 \\
 & 5M & \checkmark & \checkmark & \myxmark & \textbf{98.9} & 97.4 & 98.2 & 82.7 & \textbf{68.3} & \textbf{74.8} & \textbf{64.2} & \textbf{18.7} & \textbf{29.0} \\
 & 5M & \checkmark & \checkmark & \checkmark & \textbf{98.9} & \textbf{97.7} & \textbf{98.3} & \textbf{82.8} & 67.6 & 74.4 & 61.7 & 17.8 & 27.7 \\
\bottomrule
\end{tabular}
}
\label{tab:bioscan_clip_results_micro}
\vspace{-0mm}
\end{table}

  \begin{figure}[t]
    \centering
    \includegraphics[width=\linewidth]{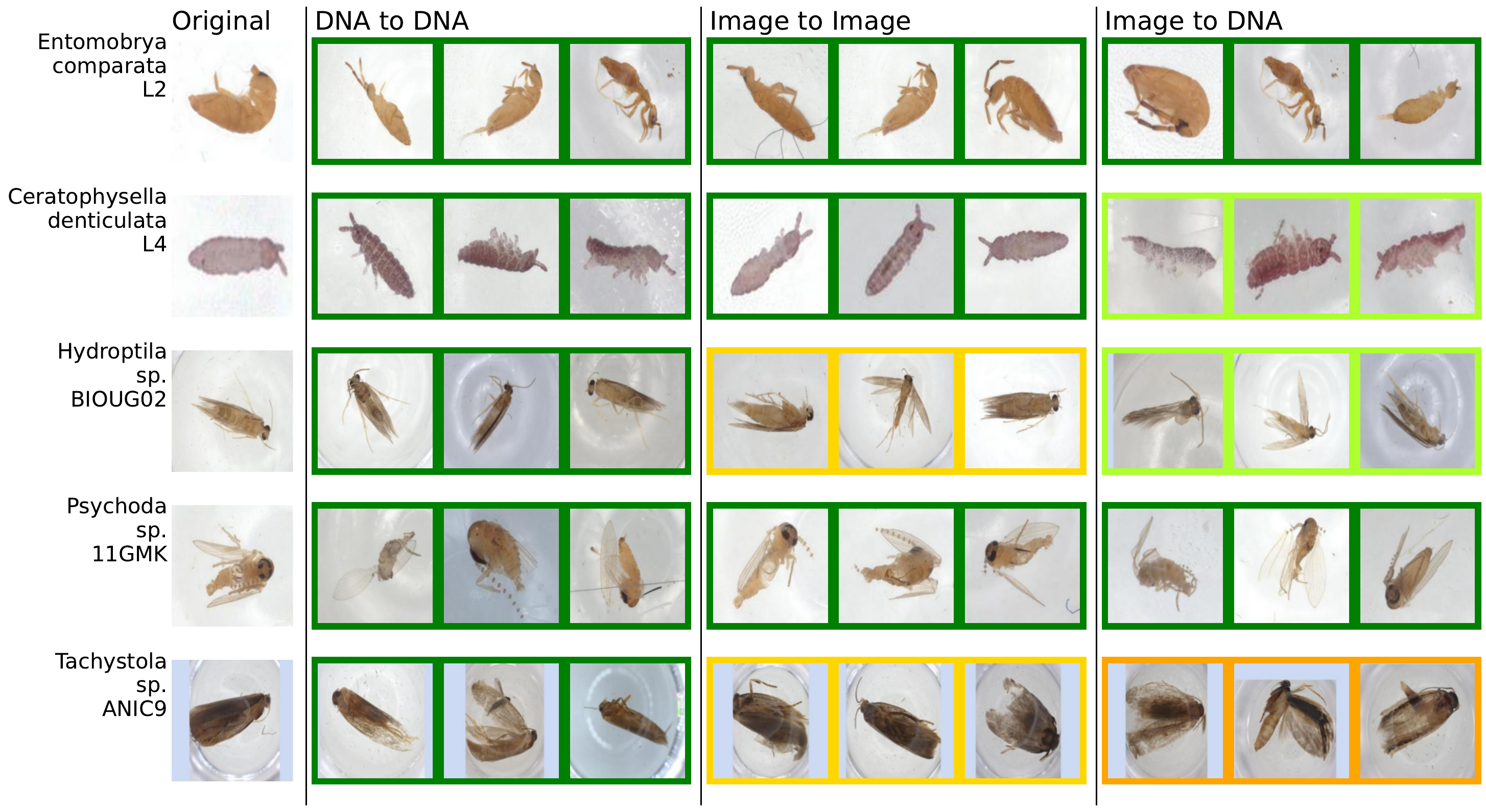}
   \vspace{0.1cm}
    \caption{\textbf{Example query-key pairs.} Top-3 nearest specimens from the unseen validation-key dataset retrieved based on the cosine-similarity for DNA-to-DNA, image-to-image, and image-to-DNA retrieval. Box colour indicates whether the retrieved samples had the same species (green), genus (light-green), family (yellow), or order (orange) as the query.
    }
    \label{fig:bioscan_clip_retrieval_examples}
    \vspace{-5mm}
  \end{figure}

\subsection{Model training and inference}
We illustrate our model training and inference methodology in \autoref{fig:bioscan_clip}. {For our multimodal model, we start with pretrained encoders for image, DNA, and taxonomic labels.}  
We use contrastive learning to fine-tune the image, DNA, and text encoders. During inference, we compare the embedding of the query image or DNA input to a key database of embeddings from images, DNA, or taxonomy labels using cosine similarity, and we predict the query's taxonomy based on the taxonomy of the closest retrieved key embeddings.

Our fine-tuned model checkpoints are openly available. For access and usage details, please see
\href{https://github.com/bioscan-ml/clibd?tab=readme-ov-file#pretrained-embeddings-and-models}{\texttt{https://github.com/bioscan-ml/clibd}}.

\subsection{Additional experiments}
{In the main paper, we reported the macro accuracy of our models.  In \autoref{tab:bioscan_clip_results_micro}, we} report the micro accuracy to compare performance when averaged over individual samples rather than classes. The results show similar trends to the macro accuracy (\autoref{fig:bioscan_clip}), with the model trained on the BIOSCAN-5M dataset performing best for broader taxa, especially in image-to-image and image-to-DNA inference setups. Results are more mixed at the species level due in part to the challenge of species classification, highlighting the importance of further research at this fine-grained level.

\subsection{Retrieval examples} 
\autoref{fig:bioscan_clip_retrieval_examples} shows image retrieval examples using images as queries and DNA as keys. These demonstrate the ability of the model to classify taxonomy based on retrieval and the visual similarities of the retrieved images corresponding to the most closely matched DNA embeddings.

\clearpage
\iftoggle{arxiv}{}{%
\section*{NeurIPS Paper Checklist}

\begin{enumerate}

\item {\bf Claims}
    \item[] Question: Do the main claims made in the abstract and introduction accurately reflect the paper's contributions and scope?
    \item[] Answer: \answerYes{} %
    \item[] Justification: Our claims in the abstract and introduction accurately reflect the paper's contributions and scope.
    \item[] Guidelines:
    \begin{itemize}
        \item The answer NA means that the abstract and introduction do not include the claims made in the paper.
        \item The abstract and/or introduction should clearly state the claims made, including the contributions made in the paper and important assumptions and limitations. A No or NA answer to this question will not be perceived well by the reviewers. 
        \item The claims made should match theoretical and experimental results, and reflect how much the results can be expected to generalize to other settings. 
        \item It is fine to include aspirational goals as motivation as long as it is clear that these goals are not attained by the paper. 
    \end{itemize}

\item {\bf Limitations}
    \item[] Question: Does the paper discuss the limitations of the work performed by the authors?
    \item[] Answer: \answerYes{} %
    \item[] Justification: We discussed the major limitations of the work in \autoref{sec:experiments}: Benchmark experiments and results. Additional information about the limitations of our work are detailed in the supplemental material.
    \item[] Guidelines:
    \begin{itemize}
        \item The answer NA means that the paper has no limitation while the answer No means that the paper has limitations, but those are not discussed in the paper. 
        \item The authors are encouraged to create a separate "Limitations" section in their paper.
        \item The paper should point out any strong assumptions and how robust the results are to violations of these assumptions (e.g., independence assumptions, noiseless settings, model well-specification, asymptotic approximations only holding locally). The authors should reflect on how these assumptions might be violated in practice and what the implications would be.
        \item The authors should reflect on the scope of the claims made, e.g., if the approach was only tested on a few datasets or with a few runs. In general, empirical results often depend on implicit assumptions, which should be articulated.
        \item The authors should reflect on the factors that influence the performance of the approach. For example, a facial recognition algorithm may perform poorly when image resolution is low or images are taken in low lighting. Or a speech-to-text system might not be used reliably to provide closed captions for online lectures because it fails to handle technical jargon.
        \item The authors should discuss the computational efficiency of the proposed algorithms and how they scale with dataset size.
        \item If applicable, the authors should discuss possible limitations of their approach to address problems of privacy and fairness.
        \item While the authors might fear that complete honesty about limitations might be used by reviewers as grounds for rejection, a worse outcome might be that reviewers discover limitations that aren't acknowledged in the paper. The authors should use their best judgment and recognize that individual actions in favor of transparency play an important role in developing norms that preserve the integrity of the community. Reviewers will be specifically instructed to not penalize honesty concerning limitations.
    \end{itemize}

\item {\bf Theory Assumptions and Proofs}
    \item[] Question: For each theoretical result, does the paper provide the full set of assumptions and a complete (and correct) proof?
    \item[] Answer: \answerNA{} %
    \item[] Justification: Our paper does not include theoretical results.
    \item[] Guidelines:
    \begin{itemize}
        \item The answer NA means that the paper does not include theoretical results. 
        \item All the theorems, formulas, and proofs in the paper should be numbered and cross-referenced.
        \item All assumptions should be clearly stated or referenced in the statement of any theorems.
        \item The proofs can either appear in the main paper or the supplemental material, but if they appear in the supplemental material, the authors are encouraged to provide a short proof sketch to provide intuition. 
        \item Inversely, any informal proof provided in the core of the paper should be complemented by formal proofs provided in appendix or supplemental material.
        \item Theorems and Lemmas that the proof relies upon should be properly referenced. 
    \end{itemize}

    \item {\bf Experimental Result Reproducibility}
    \item[] Question: Does the paper fully disclose all the information needed to reproduce the main experimental results of the paper to the extent that it affects the main claims and/or conclusions of the paper (regardless of whether the code and data are provided or not)?
    \item[] Answer: \answerYes{} %
    \item[] Justification: All the information needed to reproduce the main experimental results of our paper to the extent that it affects the main claims and/or conclusions of the paper is included in \autoref{sec:experiments}: Benchmark experiments and results.
    \item[] Guidelines:
    \begin{itemize}
        \item The answer NA means that the paper does not include experiments.
        \item If the paper includes experiments, a No answer to this question will not be perceived well by the reviewers: Making the paper reproducible is important, regardless of whether the code and data are provided or not.
        \item If the contribution is a dataset and/or model, the authors should describe the steps taken to make their results reproducible or verifiable. 
        \item Depending on the contribution, reproducibility can be accomplished in various ways. For example, if the contribution is a novel architecture, describing the architecture fully might suffice, or if the contribution is a specific model and empirical evaluation, it may be necessary to either make it possible for others to replicate the model with the same dataset, or provide access to the model. In general. releasing code and data is often one good way to accomplish this, but reproducibility can also be provided via detailed instructions for how to replicate the results, access to a hosted model (e.g., in the case of a large language model), releasing of a model checkpoint, or other means that are appropriate to the research performed.
        \item While NeurIPS does not require releasing code, the conference does require all submissions to provide some reasonable avenue for reproducibility, which may depend on the nature of the contribution. For example
        \begin{enumerate}
            \item If the contribution is primarily a new algorithm, the paper should make it clear how to reproduce that algorithm.
            \item If the contribution is primarily a new model architecture, the paper should describe the architecture clearly and fully.
            \item If the contribution is a new model (e.g., a large language model), then there should either be a way to access this model for reproducing the results or a way to reproduce the model (e.g., with an open-source dataset or instructions for how to construct the dataset).
            \item We recognize that reproducibility may be tricky in some cases, in which case authors are welcome to describe the particular way they provide for reproducibility. In the case of closed-source models, it may be that access to the model is limited in some way (e.g., to registered users), but it should be possible for other researchers to have some path to reproducing or verifying the results.
        \end{enumerate}
    \end{itemize}

\item {\bf Open access to data and code}
    \item[] Question: Does the paper provide open access to the data and code, with sufficient instructions to faithfully reproduce the main experimental results, as described in supplemental material?
    \item[] Answer: \answerYes{} %
    \item[] Justification: Our paper provides open access to the data (see supplemental material) and code (see our \hreffoot{https://github.com/bioscan-ml/BIOSCAN-5M}{GitHub repo}, linked to in the abstract), with sufficient instructions to faithfully reproduce the main experimental results described in \autoref{sec:experiments} and our Appendices.
    \item[] Guidelines:
    \begin{itemize}
        \item The answer NA means that paper does not include experiments requiring code.
        \item Please see the NeurIPS code and data submission guidelines (\url{https://nips.cc/public/guides/CodeSubmissionPolicy}) for more details.
        \item While we encourage the release of code and data, we understand that this might not be possible, so “No” is an acceptable answer. Papers cannot be rejected simply for not including code, unless this is central to the contribution (e.g., for a new open-source benchmark).
        \item The instructions should contain the exact command and environment needed to run to reproduce the results. See the NeurIPS code and data submission guidelines (\url{https://nips.cc/public/guides/CodeSubmissionPolicy}) for more details.
        \item The authors should provide instructions on data access and preparation, including how to access the raw data, preprocessed data, intermediate data, and generated data, etc.
        \item The authors should provide scripts to reproduce all experimental results for the new proposed method and baselines. If only a subset of experiments are reproducible, they should state which ones are omitted from the script and why.
        \item At submission time, to preserve anonymity, the authors should release anonymized versions (if applicable).
        \item Providing as much information as possible in supplemental material (appended to the paper) is recommended, but including URLs to data and code is permitted.
    \end{itemize}

\item {\bf Experimental Setting/Details}
    \item[] Question: Does the paper specify all the training and test details (e.g., data splits, hyperparameters, how they were chosen, type of optimizer, etc.) necessary to understand the results?
    \item[] Answer: \answerYes{} %
    \item[] Justification: \autoref{sec:experiments} contains detailed information about all experiments conducted in the paper.
    \item[] Guidelines:
    \begin{itemize}
        \item The answer NA means that the paper does not include experiments.
        \item The experimental setting should be presented in the core of the paper to a level of detail that is necessary to appreciate the results and make sense of them.
        \item The full details can be provided either with the code, in appendix, or as supplemental material.
    \end{itemize}

\item {\bf Experiment Statistical Significance}
    \item[] Question: Does the paper report error bars suitably and correctly defined or other appropriate information about the statistical significance of the experiments?
    \item[] Answer: \answerNo{} %
    \item[] Justification: No claims are made in this paper on the statistical significance of the results. In particular, only single runs of experiments were performed and thus neither error bounds nor statistical tests were performed. Our experiments were predominantly performed by adapting pretrained models; since only one repetition of each pretrained model is available, performing experiments with multiple seeds would provide a misleading indication of the variability as it only reflects a minority of the variance between the architectures and training processes prescribed by each of the pretrained models. Nonetheless, efforts were made regarding our validation strategy, utilizing a well thought out approach to the train/validation/test split described in \autoref{sec:dataset}. All experiments are conducted with specified experimental conditions, which are described in detail in \autoref{sec:experiments}.
    \item[] Guidelines:
    \begin{itemize}
        \item The answer NA means that the paper does not include experiments.
        \item The authors should answer "Yes" if the results are accompanied by error bars, confidence intervals, or statistical significance tests, at least for the experiments that support the main claims of the paper.
        \item The factors of variability that the error bars are capturing should be clearly stated (for example, train/test split, initialization, random drawing of some parameter, or overall run with given experimental conditions).
        \item The method for calculating the error bars should be explained (closed form formula, call to a library function, bootstrap, etc.)
        \item The assumptions made should be given (e.g., Normally distributed errors).
        \item It should be clear whether the error bar is the standard deviation or the standard error of the mean.
        \item It is OK to report 1-sigma error bars, but one should state it. The authors should preferably report a 2-sigma error bar than state that they have a 96\% CI, if the hypothesis of Normality of errors is not verified.
        \item For asymmetric distributions, the authors should be careful not to show in tables or figures symmetric error bars that would yield results that are out of range (e.g. negative error rates).
        \item If error bars are reported in tables or plots, The authors should explain in the text how they were calculated and reference the corresponding figures or tables in the text.
    \end{itemize}

\item {\bf Experiments Compute Resources}
    \item[] Question: For each experiment, does the paper provide sufficient information on the computer resources (type of compute workers, memory, time of execution) needed to reproduce the experiments?
    \item[] Answer: \answerYes{} %
    \item[] Justification: We detailed information about the resources used to run experiments in \autoref{s:exp-clip}, \autoref{a:exp-dna}, and \autoref{a:exp-zsc}.
    \item[] Guidelines:
    \begin{itemize}
        \item The answer NA means that the paper does not include experiments.
        \item The paper should indicate the type of compute workers CPU or GPU, internal cluster, or cloud provider, including relevant memory and storage.
        \item The paper should provide the amount of compute required for each of the individual experimental runs as well as estimate the total compute. 
        \item The paper should disclose whether the full research project required more compute than the experiments reported in the paper (e.g., preliminary or failed experiments that didn't make it into the paper). 
    \end{itemize}
    
\item {\bf Code Of Ethics}
    \item[] Question: Does the research conducted in the paper conform, in every respect, with the NeurIPS Code of Ethics \url{https://neurips.cc/public/EthicsGuidelines}?
    \item[] Answer: \answerYes{} %
    \item[] Justification: The research conducted in the paper conform, in every respect, with the NeurIPS Code of Ethics.
    \item[] Guidelines:
    \begin{itemize}
        \item The answer NA means that the authors have not reviewed the NeurIPS Code of Ethics.
        \item If the authors answer No, they should explain the special circumstances that require a deviation from the Code of Ethics.
        \item The authors should make sure to preserve anonymity (e.g., if there is a special consideration due to laws or regulations in their jurisdiction).
    \end{itemize}

\item {\bf Broader Impacts}
    \item[] Question: Does the paper discuss both potential positive societal impacts and negative societal impacts of the work performed?
    \item[] Answer: \answerNA{} %
    \item[] There is no direct societal impact of the work performed. This work specifically relates to global biodiversity in arthropods. However, the hope for positive \emph{ecological} impact is discussed. 
    \item[] Guidelines:
    \begin{itemize}
        \item The answer NA means that there is no societal impact of the work performed.
        \item If the authors answer NA or No, they should explain why their work has no societal impact or why the paper does not address societal impact.
        \item Examples of negative societal impacts include potential malicious or unintended uses (e.g., disinformation, generating fake profiles, surveillance), fairness considerations (e.g., deployment of technologies that could make decisions that unfairly impact specific groups), privacy considerations, and security considerations.
        \item The conference expects that many papers will be foundational research and not tied to particular applications, let alone deployments. However, if there is a direct path to any negative applications, the authors should point it out. For example, it is legitimate to point out that an improvement in the quality of generative models could be used to generate deepfakes for disinformation. On the other hand, it is not needed to point out that a generic algorithm for optimizing neural networks could enable people to train models that generate Deepfakes faster.
        \item The authors should consider possible harms that could arise when the technology is being used as intended and functioning correctly, harms that could arise when the technology is being used as intended but gives incorrect results, and harms following from (intentional or unintentional) misuse of the technology.
        \item If there are negative societal impacts, the authors could also discuss possible mitigation strategies (e.g., gated release of models, providing defenses in addition to attacks, mechanisms for monitoring misuse, mechanisms to monitor how a system learns from feedback over time, improving the efficiency and accessibility of ML).
    \end{itemize}
    
\item {\bf Safeguards}
    \item[] Question: Does the paper describe safeguards that have been put in place for responsible release of data or models that have a high risk for misuse (e.g., pretrained language models, image generators, or scraped datasets)?
    \item[] Answer: \answerNA{} %
    \item[] Justification: To the best of our knowledge, our paper and our published dataset pose no such risks.
    \item[] Guidelines:
    \begin{itemize}
        \item The answer NA means that the paper poses no such risks.
        \item Released models that have a high risk for misuse or dual-use should be released with necessary safeguards to allow for controlled use of the model, for example by requiring that users adhere to usage guidelines or restrictions to access the model or implementing safety filters. 
        \item Datasets that have been scraped from the Internet could pose safety risks. The authors should describe how they avoided releasing unsafe images.
        \item We recognize that providing effective safeguards is challenging, and many papers do not require this, but we encourage authors to take this into account and make a best faith effort.
    \end{itemize}

\item {\bf Licenses for existing assets}
    \item[] Question: Are the creators or original owners of assets (e.g., code, data, models), used in the paper, properly credited and are the license and terms of use explicitly mentioned and properly respected?
    \item[] Answer: \answerYes %
    \item[] Justification: The original owners of code, dataset and model used in our paper are properly cited and credited. The license and terms of use explicitly mentioned in the supplemental material and properly respected.
    \item[] Guidelines:
    \begin{itemize}
        \item The answer NA means that the paper does not use existing assets.
        \item The authors should cite the original paper that produced the code package or dataset.
        \item The authors should state which version of the asset is used and, if possible, include a URL.
        \item The name of the license (e.g., CC-BY 4.0) should be included for each asset.
        \item For scraped data from a particular source (e.g., website), the copyright and terms of service of that source should be provided.
        \item If assets are released, the license, copyright information, and terms of use in the package should be provided. For popular datasets, \url{paperswithcode.com/datasets} has curated licenses for some datasets. Their licensing guide can help determine the license of a dataset.
        \item For existing datasets that are re-packaged, both the original license and the license of the derived asset (if it has changed) should be provided.
        \item If this information is not available online, the authors are encouraged to reach out to the asset's creators.
    \end{itemize}

\item {\bf New Assets}
    \item[] Question: Are new assets introduced in the paper well documented and is the documentation provided alongside the assets?
    \item[] Answer: \answerYes %
    \item[] Justification: All new assets introduced in our paper including, data, code, and models are well documented with documents included in the paper (the links of the dataset landing page as well as the \hreffoot{https://github.com/bioscan-ml/BIOSCAN-5M}{Github repository}) and supplemental material.
    \item[] Guidelines:
    \begin{itemize}
        \item The answer NA means that the paper does not release new assets.
        \item Researchers should communicate the details of the dataset/code/model as part of their submissions via structured templates. This includes details about training, license, limitations, etc. 
        \item The paper should discuss whether and how consent was obtained from people whose asset is used.
        \item At submission time, remember to anonymize your assets (if applicable). You can either create an anonymized URL or include an anonymized zip file.
    \end{itemize}

\item {\bf Crowdsourcing and Research with Human Subjects}
    \item[] Question: For crowdsourcing experiments and research with human subjects, does the paper include the full text of instructions given to participants and screenshots, if applicable, as well as details about compensation (if any)? 
    \item[] Answer: \answerNA{} %
    \item[] Justification: Our paper does not involve crowdsourcing nor research with human subjects.
    \item[] Guidelines:
    \begin{itemize}
        \item The answer NA means that the paper does not involve crowdsourcing nor research with human subjects.
        \item Including this information in the supplemental material is fine, but if the main contribution of the paper involves human subjects, then as much detail as possible should be included in the main paper. 
        \item According to the NeurIPS Code of Ethics, workers involved in data collection, curation, or other labor should be paid at least the minimum wage in the country of the data collector. 
    \end{itemize}

\item {\bf Institutional Review Board (IRB) Approvals or Equivalent for Research with Human Subjects}
    \item[] Question: Does the paper describe potential risks incurred by study participants, whether such risks were disclosed to the subjects, and whether Institutional Review Board (IRB) approvals (or an equivalent approval/review based on the requirements of your country or institution) were obtained?
    \item[] Answer: \answerNA{} %
    \item[] Justification: Our paper does not involve crowdsourcing nor research with human subjects.
    \item[] Guidelines:
    \begin{itemize}
        \item The answer NA means that the paper does not involve crowdsourcing nor research with human subjects.
        \item Depending on the country in which research is conducted, IRB approval (or equivalent) may be required for any human subjects research. If you obtained IRB approval, you should clearly state this in the paper. 
        \item We recognize that the procedures for this may vary significantly between institutions and locations, and we expect authors to adhere to the NeurIPS Code of Ethics and the guidelines for their institution. 
        \item For initial submissions, do not include any information that would break anonymity (if applicable), such as the institution conducting the review.
    \end{itemize}

\end{enumerate}

}
\iftoggle{arxiv}{%
\section{Dataset summary}
\label{a:dataset}
BIOSCAN-5M is a large-scale, multimodal dataset comprising over 5 million specimens, 98\% of which are insects. Unlike existing image-based datasets, BIOSCAN-5M integrates taxonomic labels, raw nucleotide barcode sequences, assigned barcode index numbers, geographical data, and specimen size information, making it one of the most comprehensive biological datasets available. \autoref{fig:dataset-summary-graphic} illustrates the data modalities available in the dataset, for one sample record. For more details about the dataset, please refer to \autoref{sec:dataset} in the main text and Appendices \ref{a:ethics}--\ref{s:partitioning-extra} below.
\begin{figure*}[!h]
	\centering    
	\includegraphics[width=\textwidth]{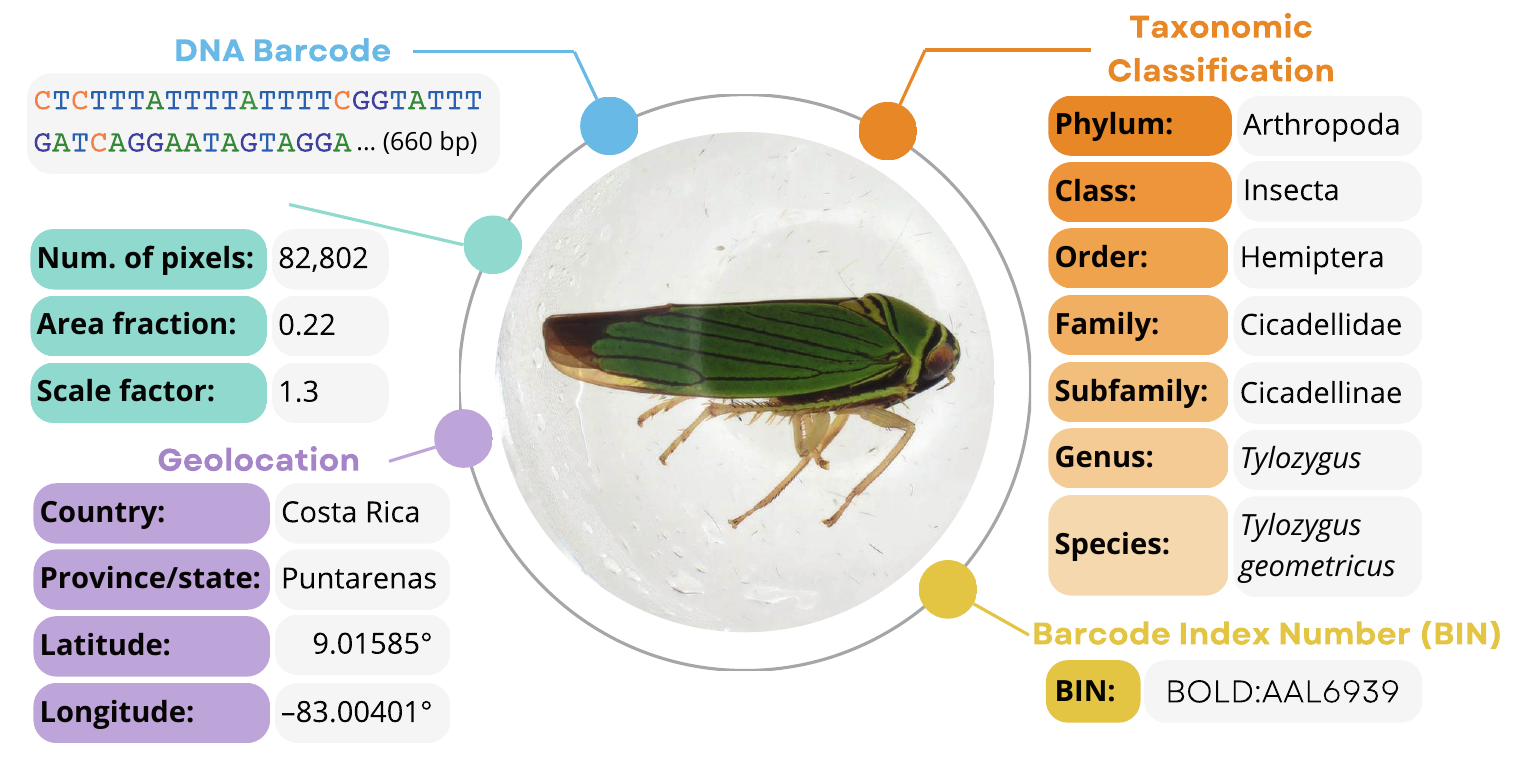}
 \caption{The BIOSCAN-5M Dataset provides taxonomic labels, a DNA barcode sequence, barcode index number, a high-resolution image along with its cropped and resized versions, as well as size and geographic information for each sample.}
 \label{fig:dataset-summary-graphic}
\end{figure*}

\section{Ethics and responsible use}
\label{a:ethics}

The BIOSCAN project was instigated by the International Barcode of Life (iBOL) Consortium, which has collected a large dataset of manually-labelled images of organisms \citep{bold_systems,steinke2024dataset}. 
As part of our project, we conducted a thorough review to identify any potential ethical issues related to the inclusion of our data sources. After careful evaluation, we did not find any ethical concerns. Therefore, we confirm that this work adheres to all relevant ethical standards and guidelines.

\section{Dataset availability and maintenance }
\label{a:availability}
To explore more about the BIOSCAN-5M dataset, kindly visit the following landing page:\\ \url{https://biodiversitygenomics.net/5M-insects/}.

The BIOSCAN-5M dataset and all its contents are available in a \href{https://drive.google.com/drive/u/0/folders/1Jc57eKkeiYrnUBc9WlIp-ZS_L1bVlT-0}{GoogleDrive Folder}. The Google Drive folder serves as the primary repository for the BIOSCAN-5M dataset, ensuring ongoing maintenance and the potential addition of new content as necessary. It will be gradually updated to address any data issues that may arise.

The Google Drive folder contains the following dataset contents:
\begin{itemize}
	
	\item \textbf{BIOSCAN\_5M\_IMAGES}: This directory contains images:
	\begin{itemize}
		\item \texttt{BIOSCAN\_5M\_original\_full}: The original full-size images.
		\item \texttt{BIOSCAN\_5M\_original\_256}: The original images resized to 256 pixels on their shorter side.
		\item \texttt{BIOSCAN\_5M\_cropped}: The cropped images.
		\item \texttt{BIOSCAN\_5M\_cropped\_256}: The cropped images resized to 256 pixels on their shorter side.
	\end{itemize}
	
	\item \textbf{BIOSCAN\_5M\_METADATA}: This directory contains metadata:
	\begin{itemize}
		\item \texttt{BIOSCAN\_5M\_Insect\_Dataset\_metadata\_MultiTypes.zip}: A zip file containing both CSV and JSON formats of the metadata file.
	\end{itemize}
	\item \textbf{BIOSCAN\_5M\_CropTool}: This directory contains our cropping tool components:
	\begin{itemize}
		\item \texttt{bounding\_box/BIOSCAN\_5M\_Insect\_bbox.tsv}: A TSV file that includes bounding box information obtained from our cropping tool.
		\item \texttt{checkpoint/BIOSCAN\_5M\_Insect\_cropping\_tool.ckpt}: The model checkpoint used to crop the original full-size images, which generated the cropped images of the BIOSCAN-5M dataset.
	\end{itemize}
\end{itemize}

Additionally, the dataset is released on several platforms, including \href{https://zenodo.org/records/11973457}{Zenodo}, \href{https://www.kaggle.com/datasets/zahragharaee/bioscan-5m}{Kaggle}, and \href{https://huggingface.co/datasets/Gharaee/BIOSCAN-5M}{HuggingFace}.

We provide a code repository for dataset manipulation, which supports tasks like reading images and metadata, cropping images, statistical processing, dataset splitting into pretrain, train, and evaluation, as well as running benchmark experiments presented in the BIOSCAN-5M paper. To access the BIOSCAN-5M code repository, please visit \url{https://github.com/bioscan-ml/BIOSCAN-5M}.

Additionally, we provide a Python package for working with the BIOSCAN-5M dataset, designed in the style of torchvision's \texttt{VisionDataset} class, which can be installed with \verb|pip install bioscan-dataset|.
For usage details, please visit \url{https://bioscan-dataset.readthedocs.io/}.

\section{Licensing}
\label{a:licensing}

\autoref{tab:ethical} shows all the copyright associations related to the BIOSCAN-5M dataset with the corresponding names and contact information.  

\begin{table}[h]
	\centering
	\caption{Copyright associations related to the BIOSCAN-5M dataset.}
    \label{tab:ethical}
	\begin{center}
		\resizebox{\linewidth}{!}{
        \begin{tabular}{ll}
            \toprule
            \textbf{Copyright Associations} & \textbf{Name \& Contact}  \\
            \midrule
            Image Photographer      & CBG Robotic Imager          \\
            Copyright Holder        & CBG Photography Group          \\
            Copyright Institution   & Centre for Biodiversity Genomics (email: CBGImaging@gmail.com) \\
            Copyright License & Creative Commons Attribution 3.0 Unported (\href{https://creativecommons.org/licenses/by/3.0/}{CC BY 3.0}) \\
            Copyright Contact     & collectionsBIO@gmail.com  \\
            Copyright Year       & 2021 \\
            \bottomrule
		\end{tabular}
  }
	\end{center}
\end{table}

The authors state that they bear all responsibility in case of violation of usage rights.

\section{RGB images}
\label{a:images}

The BIOSCAN-5M dataset comprises resized and cropped images, as introduced in BIOSCAN-1M Insect \citep{gharaee2023step}. We have provided various packages of the BIOSCAN-5M dataset, detailed in \autoref{tab:pack}, each tailored for specific purposes.
    \begin{itemize}
    \item original\_full: The raw images of the dataset, typically 1024\texttimes768 pixels.
    \item cropped: Images after cropping with our cropping tool (see \autoref{s:image-processing}).
    \item original\_256: Original images resized to 256 on their shorter side (most 341\texttimes256 pixels).
    \item cropped\_256: Cropped images resized to 256 on their shorter side.
\end{itemize}
Among these, the original\_256 and cropped\_256 packages are specifically provided for experimentation as they are small and easy to work with. Therefore, using our predefined split partitions, we provide per-split experimental packages in addition to the packages with all the original\_256 and cropped\_256 images.
\begin{table}[!h]
\begin{center}
\begin{small}
    \begin{tabular}{lllrr}
        \toprule
        \textbf{Image set} & \textbf{Package} &  \textbf{Partition(s)} & \textbf{Size (GB)} & \textbf{\# Parts}\\
        \midrule
        original\_full & BIOSCAN\_5M\_original\_full.zip        &  All & 200\phantom{.0} & 5\\
        \addlinespace
        cropped        & BIOSCAN\_5M\_cropped.zip               &  All & 77.2 & 2\\
        \addlinespace
        original\_256 & BIOSCAN\_5M\_original\_256.zip          &  All & 35.2 & 1\\
                      & BIOSCAN\_5M\_original\_256\_pretrain.zip&  Pretrain & 31.7 & 1\\
                      & BIOSCAN\_5M\_original\_256\_train.zip   &  Train & 2.1 & 1\\
                      & BIOSCAN\_5M\_original\_256\_eval.zip    &  Evaluation & 1.4 & 1\\
        \addlinespace
        cropped\_256  & BIOSCAN\_5M\_cropped\_256.zip           &  All & 36.4 & 1\\
                      & BIOSCAN\_5M\_cropped\_256\_pretrain.zip &  Pretrain & 33.0 & 1\\
                      & BIOSCAN\_5M\_cropped\_256\_train.zip    &  Train & 2.1 & 1\\
                      & BIOSCAN\_5M\_cropped\_256\_eval.zip     &  Evaluation & 1.4 & 1\\
        \bottomrule
    \end{tabular}
\end{small}	
\end{center}			
\caption{Various downloadable packages of the images comprising the BIOSCAN-5M dataset.%
}\label{tab:pack}
\end{table}

Accessing the dataset images is facilitated by the following directory structure used to organize the dataset images:

{
\small
\quad\texttt{bioscan5m/images/[imgtype]/[split]/[chunk]/[{processid}.jpg]}
}

where \texttt{[imgtype]} can be \texttt{original\_full}, \texttt{cropped}, \texttt{original\_256}, or \texttt{cropped\_256}. The \texttt{[split]} values can be \texttt{pretrain}, \texttt{train}, \texttt{val}, \texttt{test}, \texttt{val\_unseen}, \texttt{test\_unseen}, \texttt{key\_unseen}, or \texttt{other\_heldout}. Note that the \texttt{val}, \texttt{test}, \texttt{val\_unseen}, \texttt{test\_unseen}, \texttt{key\_unseen}, and \texttt{other\_heldout} splits are within the evaluation partition of the original\_256 and cropped\_256 image packages.

The \texttt{[chunk]} is determined by using the first one or two characters of the MD5 checksum (in hexadecimal) of the \texttt{processid}. This method ensures that the chunk name is purely deterministic and can be computed directly from the \texttt{processid}. As a result, the \texttt{pretrain} split organizes files into 256 directories by using the first two letters of the MD5 checksum of the \texttt{processid}. For the \texttt{train} and \texttt{other\_heldout} splits, files are organized into 16 directories using the first letter of the MD5 checksum. The remaining splits do not use chunk directories since each split has less than 50\,k images.

\section{Metadata file}
\label{a:metadata}

To enrich the metadata of our published dataset, we provide integrated structured metadata conforming to Web standards. Our dataset's metadata file is titled \textbf{BIOSCAN\textunderscore5M\textunderscore Insect\textunderscore Dataset\textunderscore metadata}. We provide two versions of this file: one in CSV format (\textbf{.csv}) and the other in JSON-LD format (\textbf{.jsonld}). 
Accessing the dataset metadata files is facilitated by the following directory structure used to organize the dataset images:

{
\small
\quad\texttt{bioscan5m/metadata/[type]/BIOSCAN\_5M\_Insect\_Dataset\_metadata.[type\_extension]}
}

In this structure, \texttt{[type]} refers to the file type of the metadata file, which can be either CSV or JSON-LD. The \texttt{[type\_extension]} indicates the corresponding file extensions, which are \texttt{csv} for CSV files and \texttt{jsonld} for JSON-LD files.

\autoref{tab:attr_desc} outlines the fields of the metadata file and the description of their contents.
\begin{table}[!h]
\begin{center}
\begin{small}

\resizebox{\linewidth}{!}{
    \begin{tabular}{@{}rlll@{}}
        \toprule
        & \textbf{Field} &      \textbf{Description} & \textbf{Type}\\
        \midrule
        1&\texttt{processid}                  & A unique number assigned
by BOLD (International Barcode of Life Consortium). & String \\
        2&\texttt{sampleid}                   & A unique identifier given by the collector. & String \\
        3&\texttt{taxon}                      & Bio.info: Most specific taxonomy rank. & String \\
        4&\texttt{phylum}                     & Bio.info: Taxonomic classification label at phylum rank. & String \\
        5&\texttt{class}                      & Bio.info: Taxonomic classification label at class rank. & String \\
        6&\texttt{order}                      & Bio.info: Taxonomic  classification label at order rank. & String \\
        7&\texttt{family}                     & Bio.info: Taxonomic  classification label at family rank. & String \\
        8&\texttt{subfamily}                  & Bio.info: Taxonomic  classification label at subfamily rank. & String \\
        9&\texttt{genus}                      & Bio.info: Taxonomic  classification label at genus rank. & String \\
        10&\texttt{species}                   & Bio.info: Taxonomic  classification label at species rank. & String \\
        11&\texttt{dna\_bin}                  & Bio.info: Barcode Index Number (BIN). & String \\
        12&\texttt{dna\_barcode}              & Bio.info: Nucleotide barcode sequence. & String \\
        13&\texttt{country}                   & Geo.info: Country associated with the site of collection. & String \\
        14&\texttt{province\_state}           & Geo.info: Province/state associated with the site of collection. & String \\
        15&\texttt{coord-lat}                 & Geo.info: Latitude (WGS 84; decimal degrees) of the collection site. & Float \\
        16&\texttt{coord-lon}                 & Geo.info: Longitude (WGS 84; decimal degrees) of the collection site. & Float \\    
        17&\texttt{image\_measurement\_value} & Size.info: Number of pixels occupied by the organism. & Integer \\
        18&\texttt{area\_fraction}            & Size.info: Fraction of the original image the cropped image comprises. & Float \\
        19&\texttt{scale\_factor}             & Size.info: Ratio of the cropped image to the cropped\_256 image. & Float \\
        20&\texttt{inferred\_ranks}           & An integer indicating at which taxonomic ranks the label is inferred. & Integer \\
        21&\texttt{split}                     & Split set (partition) the sample belongs to. & String \\
        22&\texttt{index\_bioscan\_1M\_insect}& An index to locate organism in BIOSCAN-1M Insect metadata.& Integer \\
        23&\texttt{chunk}                     & The packaging subdirectory name (or empty string) for this image. & String \\
        \bottomrule
    \end{tabular}
}
\end{small}
\end{center}			
\caption{Metadata fields for the BIOSCAN-5M dataset.}
\label{tab:attr_desc}
\end{table}

\section{Comparison between BIOSCAN-5M and BIOSCAN-1M}
\label{a:comparison-BIOSCAN-1M}
{
The six key differences between BIOSCAN-1M and BIOSCAN-5M are as follows:
\begin{enumerate}
	\item \textbf{Increased data volume:} BIOSCAN-5M contains five times as many samples as BIOSCAN-1M.
	\item \textbf{Greater data diversity:} BIOSCAN-5M is collected from a broader range of geographic locations (3 countries in BIOSCAN-1M; 47 countries in BIOSCAN-5M) and encompasses a wider variety of insect life (1 class and 16 orders in BIOSCAN-1M; 10 classes and 55 orders in BIOSCAN-5M).
	\item \textbf{Enhanced post-processing:} The taxonomic labels in BIOSCAN-5M underwent a rigorous data cleaning pipeline to identify and resolve inconsistencies in the original data, resulting in more reliable labels compared to those in BIOSCAN-1M.
	\item \textbf{Geographic and specimen size data:} This information is available in BIOSCAN-5M but not in BIOSCAN-1M. 
	\item \textbf{Comprehensive partitioning support:} BIOSCAN-5M offers robust support for both closed-world and open-world tasks, whereas BIOSCAN-1M only supports closed-world partitioning.
	\item \textbf{Enhanced benchmarking experiments:} BIOSCAN-1M included a baseline with an image-only model evaluated at order and family ranks. In contrast, BIOSCAN-5M features three baselines that leverage the multimodal aspects of the dataset (including DNA barcode sequences, textual taxonomic labels, and RGB images), allowing for performance exploration in both closed- and open-world settings.
\end{enumerate}
}

\section{Focus and objectives}
\label{a:focus}
{
We have released dataset splits for closed-world and open-world settings, using labelled species data for evaluation and reserving unlabelled data for pretraining. Our splitting approach and configurations offer valuable resources to the ML community. BIOSCAN-5M experiments evaluate down to the species level. Additionally, we benchmark three distinct tasks to showcase BIOSCAN-5M's multimodal utility in real-world applications: fine-grained taxonomic classification with DNA sequences, classification using DNA, images, and taxonomic labels, and clustering of DNA and image embeddings.}

\subsection{Leveraging unlabelled and multimodal data for enhanced taxonomic classification}
{%
It's important to note that taxonomic classification from images presents greater challenges compared to DNA barcodes, as illustrated by our clustering experiments; thus, paired data can be valuable even when unlabelled. Additionally, data not labelled at the species level remains useful for pretraining, highlighting the crucial role of unlabelled data in model development. In BIOSCAN-5M, we employ BERT-style masked sequence modelling to pretrain and encode DNA sequences, complemented by contrastive learning to align image and DNA embeddings. This pretraining approach enhances the model's ability to generalize across various applications. 
}
\subsection{Lack of utilization of geographic and size information in models}
{
In BIOSCAN-5M, we focus on biological (taxonomic labels) and genetic (DNA barcode sequences and BIN) data for fine-grained taxonomic classification, intentionally excluding geographic and size information from our experiments. Our rationale is that while geographic and size data can help rule out certain species (e.g., knowing a sample was collected in North America excludes species not found there, and knowing a sample's size eliminates species that do not grow that large), they alone do not provide sufficient information for accurate species classification. In contrast, image and genetic data are often sufficient for accurate species-level predictions.}

{
We believe that models incorporating geographic and size data will need to do so alongside image and genetic data. Therefore, models using only image and genetic information serve as valuable baselines for future work that combines these data types. Given the complexities of integrating geographic and size data into our models, we prioritized establishing a broad range of image and genetic baselines in this study and plan to explore the incorporation of geographic and size data in future research. We anticipate that effective use of this additional information will enhance model performance and look forward to the community's advancements in this area.
}

\section{Dataset feature statistics}
\label{s:dataset-stats}
This section provides additional information regarding the dataset, including a detailed statistical analysis of its diverse multimodal data types and processing methods.

\subsection{Geographical information}
The detailed statistical analysis of the geographical locations where the organisms were collected is presented in \autoref{tab:geo_stat}. This table indicates the number of distinct regions represented by country, province or state, along with their corresponding latitude and longitude. Additionally, \autoref{tab:geo_stat} provides the count of labelled versus unlabelled records, as well as the class imbalance ratio (IR) for each location group within the dataset. 

\begin{table}[!h]
	\begin{center}
		\begin{small}
			\resizebox{\linewidth}{!}{				
\begin{tabular}{lrrrrrr}
\toprule
\textbf{Geo locations} &  \textbf{Categories} &  \textbf{Labelled} &  \textbf{Labelled (\%)} &  \textbf{Unlabelled} &  \textbf{Unlabelled (\%)} &  \textbf{IR} \\
\midrule
\texttt{country}         &           47 &           5,150,842 &        100.00 &                    8 &            0.00 &               325,631.6 \\
\texttt{province\_state} &          102 &           5,058,718 &         98.21 &               92,132 &            1.79 &             1,243,427.0 \\ 
\texttt{coord-lat}       &        1,394 &           5,149,019 &         99.96 &                1,831 &            0.04 &               556,352.0 \\
\texttt{coord-lon}       &        1,489 &           5,149,019 &         99.96 &                1,831 &            0.04 &               618,931.0 \\
\midrule
Location (lat, lon)      &        1,650 &           5,149,019 &         99.96 &                1,831 &            0.04 &               520,792.0 \\
\bottomrule
\end{tabular}
}
\end{small}
\end{center} 		
\caption{The statistics for the columns indicating geographical locations where the specimens are collected.}\label{tab:geo_stat}
\end{table}

The latitude and longitude coordinates indicate that the dataset comprises 1,650 distinct regions with unique geographical locations shown by \autoref{tab:geo_stat}. However, a significant portion of the organisms—approximately 73.36\%—were collected from the top 70 most populated regions, which represent only 4.24\% of the total regions identified by their coordinates.

\begin{figure*}[!ht]
	\centering    
	\includegraphics[width=1\textwidth]{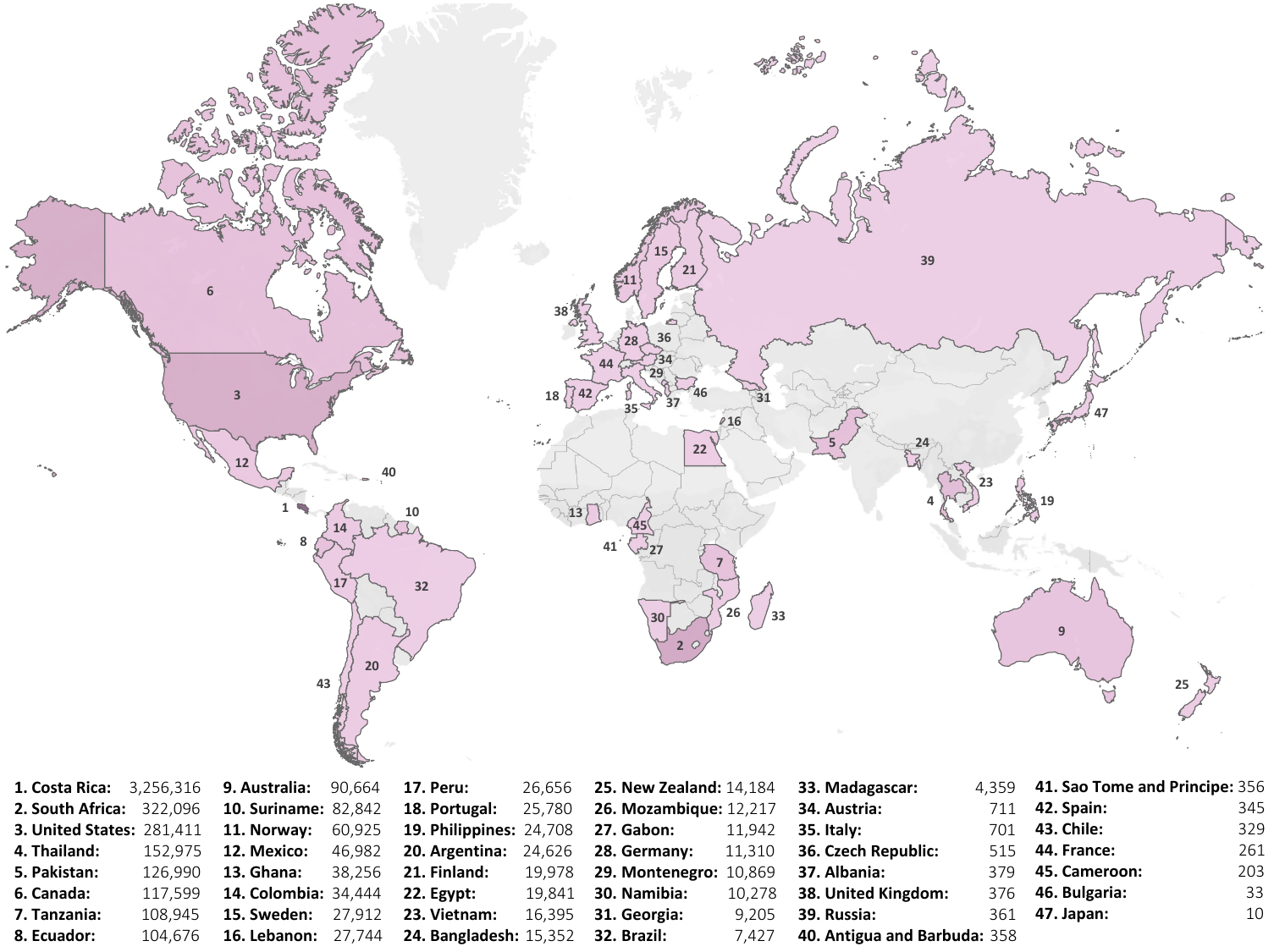}
	\caption{Global distribution of sample collection efforts. The countries are ranked by the number of samples collected.}
	\label{fig:geo-country}
\end{figure*}

\autoref{fig:geo-country} shows the distinct countries where the organisms were collected on the world map. The majority of the organisms, over 62\%, were collected from Costa Rica.

\subsection{Size information}
Monitoring organism size is crucial as it can signal shifts in various factors affecting their lives, including food access, nutrition, and climate change \citep{sheridan2011shrinking}. For instance, in humans, limited access to nutrition correlates with a decrease in average height over generations \citep{steckel1995stature}, reflecting environmental and economic changes. Tracking organism size offers insights into environmental shifts vital for biodiversity conservation \citep{hickling2006distributions}.

\mypara{Pixel count.}
The raw dataset provides information about each organism's size by quantifying the total number of pixels occupied by the organism. This information is provided in the \texttt{image\_measurement\_value} field. Since the image capture settings are consistent for all images, irrespective of scale, as indicated by the organism's distance to the camera, the number of pixels occupied by the organism should approximate its size. Less than 1\% of samples of the BIOSCAN-5M dataset do not have this information.

To provide a clearer understanding of the content in the \texttt{image\_measurement\_value} field, \autoref{fig:pixels} displays examples of original images along with their corresponding masks, highlighting the total number of pixels occupied by an organism.
To determine the real-world size of the organism based on the number of pixels, it is also important to have the pixel to metric scaling factor.  For the original full sized images, most of the images are captured using a Keyence imaging system with a known pixel to millimetre scaling.  See \appref{s:image-processing} for details on the pixel scale and how to determine it for cropped and resized images.

\begin{figure}[!h]
	\centering    
	\includegraphics[width=\textwidth]{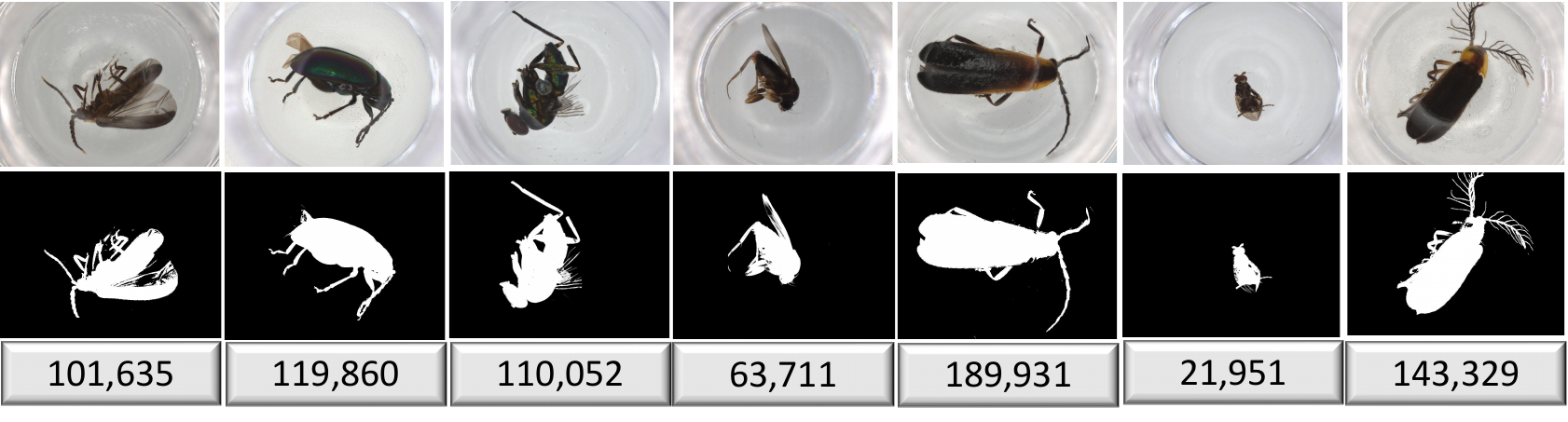}
	\caption{Examples of original images of the BIOSCAN-5M dataset, along with their respective total number of pixels (size) that occupy the image. The top row shows original images and the bottom row shows masks.}
	\label{fig:pixels}
\end{figure}

\section{Taxonomic category distribution}
\label{a:category-distribution}
\autoref{fig:stat_class} illustrates the taxonomic \texttt{class} distribution within the rank \texttt{order}. For example, of the 99.9\% of organisms labelled at the \texttt{class} level, approximately 71\% are classified within the \texttt{order} \textit{Diptera} of the \texttt{class} \textit{Insecta}.

\begin{figure}[!ht]
	\centering

\includegraphics[width=\textwidth]{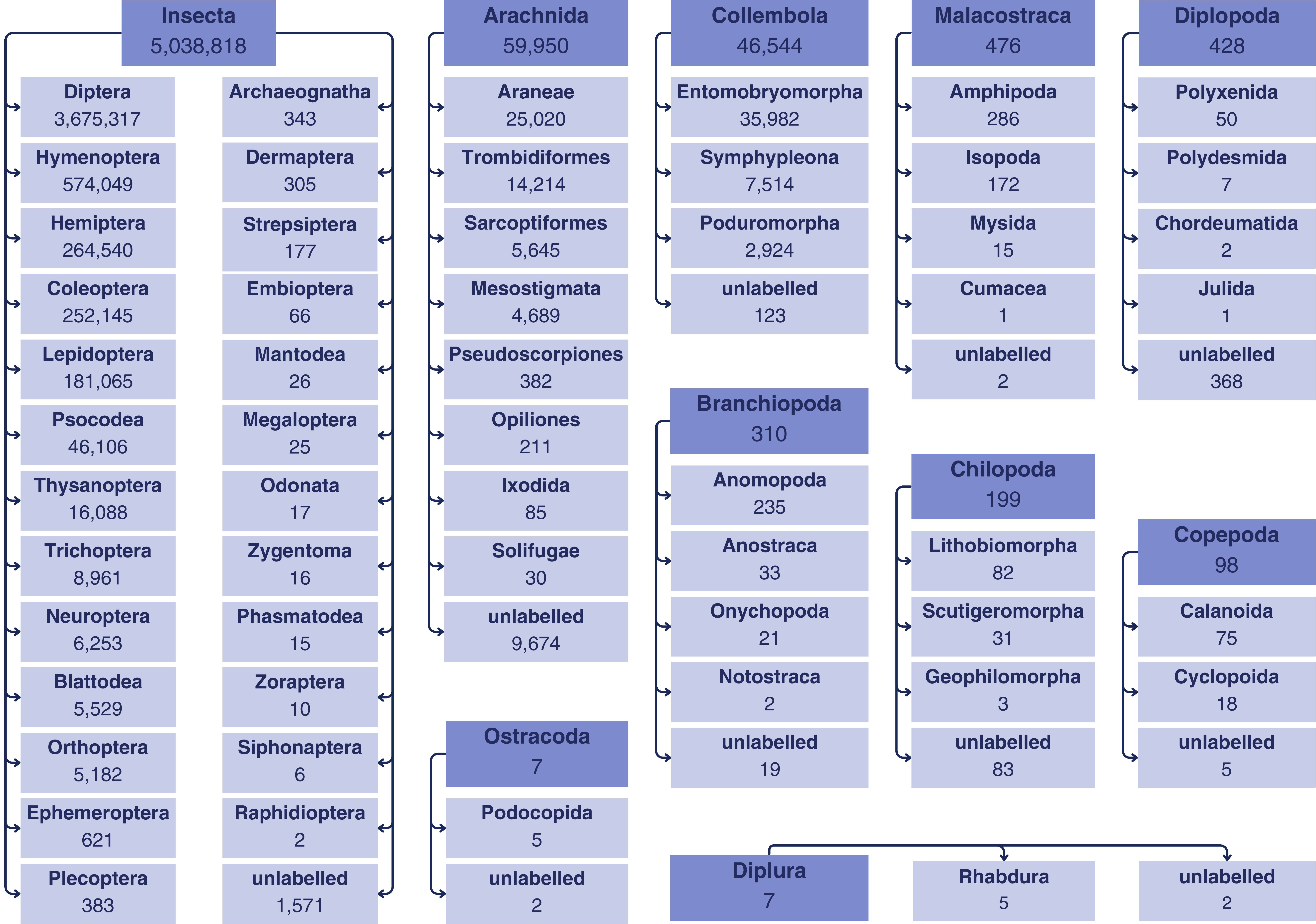}
	\caption{Distribution of taxonomic ranks in the BIOSCAN-5M dataset. Each darker cell represents a taxonomic \texttt{class}, while the lighter cells within each \texttt{class} represent the corresponding taxonomic \texttt{orders}. The numbers indicate the records belonging to each \texttt{class} and \texttt{order}. The \textit{unlabeled} category denotes records assigned to a \texttt{class} but not to any specific \texttt{order}.}
	\label{fig:stat_class}
\end{figure}

For detailed insights into the class distribution within the major categories of the BIOSCAN-5M dataset, \autoref{tab:cat_dits} presents comprehensive statistics. This table provides the total number of categories across 7 taxonomic group levels and BINs, highlighting both the most and least densely populated ones. Additionally, it includes calculated means, medians, and standard deviations of the population vectors of all subcategories of each attribute.
\begin{table}[!h]
\begin{center}
\begin{small}
\resizebox{14.0cm}{!}{
    \begin{tabular}{@{}lrlrlrrrr@{}}
    \toprule
    &   &    \multicolumn{2}{c}{\textbf{Most populated}} &     \multicolumn{2}{c}{\textbf{Least populated}} &        &      &         \\
    \cmidrule(lr){3-4} \cmidrule(lr){5-6}
    
    \textbf{Attributes} & \textbf{Categories} &   \textbf{Name} & \textbf{Size} &     \textbf{Name} & \textbf{Size} & \textbf{Mean}      & \textbf{Median}     &    \textbf{Std. Dev.}
    \\
    \midrule
    \texttt{phylum}    &        1 &         Arthropoda &     5,150,850 &                Arthropoda & 5,150,850.0 &  &  &          \\
    \texttt{class}     &       10 &            Insecta &     5,038,818 &                 Ostracoda &         7 &   514,683.7 &       369.0 & 1,508,192.8 \\
    \texttt{order}     &       55 &            Diptera &     3,675,317 &                   Cumacea &         1 &    93,363.4 &       172.0 &   495,969.5 \\
    \texttt{family}    &      934 &      Cecidomyiidae &       938,928 &             Pyrgodesmidae &         1 &     5,281.3 &        63.5 &    45,321.1 \\
    \texttt{subfamily} &    1,542 &        Metopininae &       323,146 &               Bombyliinae &         1 &       953.7 &        23.0 &     9,092.8 \\
    \texttt{genus}     &    7,605 &          Megaselia &       200,268 &           chalMalaise9590 &         1 &       161.3 &         6.0 &     2,492.2 \\
    \texttt{species}   &   22,622 & Psychoda sp. 11GMK &         7,694 & Microcephalops sp. China3 &         1 &        20.9 &         2.0 &       139.5 \\
    \midrule
    \texttt{dna\_bin}  &  324,411 &       BOLD:AEO1530 &        35,458 &              BOLD:ADT1070 &         1 &        15.8 &         2.0 &       146.4 \\
    \bottomrule
    \end{tabular}
}
\end{small}
\end{center} 		
\caption{BIOSCAN-5M taxonomic and BIN categories distribution. For each attribute, we show the value which occurs most often in the dataset and the least populated value (in the event of a tie, we show an exemplar selected at random).}
\label{tab:cat_dits}
\end{table}

\section{DNA barcode statistics}
\label{a:barcode-statistics}
{This section presents the DNA barcode statistics and analysis for the BIOSCAN-5M dataset. We provide several different statistics to show how the diversity of DNA barcodes varies across the different taxonomic levels.  
In \autoref{tab:dna_stats}, we report the number of distinct barcodes, the Shannon diversity index (e.g. entropy), and the average pairwise distances between barcodes at different taxonomic ranks.
The different analysis all show that at higher levels of taxa, there are more distinct barcodes, and that at the genus and species level, the lexical distance between different barcodes are much smaller than at the higher levels of taxa.  
Below we provide more details on how these statistics are computed.}

\subsection{Identical DNA barcodes: Statistical insights from the BIOSCAN-5M dataset}
{%
We compute and show in \autoref{tab:dna_stats} the statistics for identical DNA barcode sequences across taxonomic ranks, including the total number of distinct barcodes per rank, as well as the average, median, and standard deviation of barcodes counts across subgroups within each rank.}

{%
Based on the statistics in \autoref{tab:dna_stats}, the total number of identical DNA barcode sequences within each subgroup of a specific taxonomic rank is lower than the total number of DNA sequences corresponding to the labelled samples in that subgroup. This indicates that some samples share identical DNA barcodes, possibly due to sequencing limitations.  Since DNA barcodes are merely short snippets, they alone do not fully capture the unique genetic characteristics of individual samples.
}
\begin{table}[!ht]
	\begin{center}
		\resizebox{\textwidth}{!}{%
			\begin{tabular}{@{}lrrrrrrrr@{}}
				\toprule
				& &    \multicolumn{5}{c}{\textbf{Unique Barcodes}} &     \multicolumn{2}{c}{\textbf{Pairwise Distance}} \\
				\cmidrule(lr){3-7} \cmidrule(lr){8-9}
				
				\textbf{Attributes}& \textbf{Categories} & \textbf{Total} &   \textbf{Mean} & \textbf{Median} & \textbf{Std. Dev.}& \textbf{Avg SDI}&  \textbf{Mean}  & \textbf{Std. Dev.}  		\\
				\midrule
				\texttt{phylum}     &1     & 2,486,492  &           &        &          &19.78& 158        & 42 \\
				\texttt{class}      &10    & 2,482,891  & 248,289   & 177    & 725,237  & 8.56 & 166     & 103 \\
				\texttt{order}      &55    & 2,474,855  & 44,997    & 57     & 225,098  & 7.05 & 128      & 53 \\
				\texttt{family}     &934   & 2,321,301  & 2,485     & 46     & 19,701   & 5.42 & 90       & 46 \\
				\texttt{subfamily}  &1,542 & 657,639    & 426       & 17     & 3,726    & 4.28 & 78       & 51 \\
				\texttt{genus}      &7,605 & 531,109    & 70        & 5      & 1,061    & 2.63 & 50       & 39 \\
				\texttt{species}    &22,622& 202,260    & 9         & 2      & 37       & 1.46 & 17       & 18 \\
				\bottomrule
			\end{tabular}
		}
	\end{center}
	\caption{
		{The DNA barcode statistics for various taxonomic ranks in the BIOSCAN-5M dataset. We indicate the total number of unique barcodes for the samples labelled to a given rank, and the mean, median, and standard deviation of the number of unique barcodes within the subgroupings at that rank. We also show the average across subgroups of the Shannon Diversity Index (SDI) for the DNA barcodes, measured in bits. We report the mean and standard deviation of pairwise DNA barcode sequence distances, aggregated across subgroups for each taxonomic rank.}}
	\label{tab:dna_stats}
\end{table}

\subsection{Analyzing genetic diversity with the Shannon Index}
{
\mypara{Shannon Diversity Index (SDI).}
The Shannon Diversity Index (SDI)~\citep{shannon1948mathematical}, which measures the entropy within a group, is an effective metric for measuring genetic diversity as it considers both barcode richness (the number of distinct barcodes) and evenness (the distribution of samples among those barcodes). A high prevalence of identical barcodes leads to lower evenness and, consequently, a reduced SDI, indicating limited diversity and redundancy in genetic makeup.
}

Incorporating duplicated barcodes allows the SDI to capture the prevalence of specific barcodes within the subgroup. 
If certain barcodes are common across samples, the index may reflect a dominant genetic signature, resulting in a lower SDI and suggesting reduced diversity. 
Conversely, a greater presence of distinct barcodes with even distributions yields a higher SDI, indicating a more diverse subgroup structure. This dual focus on richness and evenness underscores the SDI's value in assessing genetic diversity and elucidating the genetic relationships within a subgroup.

{We compute the Shannon Diversity Index (SDI) for each subgroup, $T$, within a taxonomic rank as}
\begin{equation}
\text{SDI}_T = -\sum_{i=1}^{N} p_i \log_{2}(p_i),
\end{equation}
{where \(N\) is the number of unique DNA barcodes within a subgroup, and \(p_i\) is the fraction of samples in subgroup $T$ which have the $i$-th barcode.
}

{
In \autoref{tab:dna_stats}, we report the average SDI (Avg SDI) for each taxonomic rank by computing $\text{SDI}_T$ for each subgroup and then averaging these values across all subgroups within the respective rank. 
From the \autoref{tab:dna_stats}, the Avg SDI values indicate a high level of biodiversity at the \texttt{phylum} ($19.78$) and \texttt{class} ($8.56$) levels, suggesting a rich community with a wide variety of taxa. 
However, as we move down the taxonomic hierarchy, the index values decline significantly, reaching the lowest point at the \texttt{species} level ($1.46$). This pattern suggests that while there is a diverse range of \texttt{phyla} and \texttt{classes}, the distribution of \texttt{species} within these groups is uneven, indicating the presence of a few dominant \texttt{species} or \texttt{genera}.

}

\subsection{Pairwise distance analysis of identical DNA barcodes}

{%
\mypara{Damerau-Levenshtein Distance.} The {Damerau-Levenshtein distance}~\citep{damerau1964technique} is a string-edit distance metric that measures the minimum number of operations required to transform one string into another. It is an extension of the standard Levenshtein distance~\citep{levenshtein1966binary}, which counts the number of single-character edits needed for transformation. The key difference is that the Damerau-Levenshtein distance also accounts for transpositions, i.e., when two adjacent characters are swapped. In the context of our DNA barcoding, it measures how similar or different two DNA sequences are by counting how many single-character changes (insertions, deletions, substitutions, or transpositions) are needed to make one sequence identical to another.
}

We report the average Damerau-Levenshtein pairwise distance between unique DNA barcodes at different taxonomic ranks in \autoref{tab:dna_stats}.
To compute the statistics for the pairwise distances, we take each subgroup at every taxonomic rank, and only consider subgroups with sufficient number of distinct barcodes.
For a given subgroup, if the total number of unique DNA barcode sequences is fewer than 4, the subgroup is not considered. If the total exceeds 1,000, up to 1,000 sequences are randomly sampled; otherwise, all sequences are included.

To compute the distances between barcodes, the sampled DNA barcode sequences are first aligned using the MAFFT alignment technique~\citep{katoh2013mafft}. Next, the pairwise distances between aligned DNA barcodes are computed using the Damerau-Levenshtein metric, with a total of $n \times \frac{(n-1)}{2}$ comparisons (where $n$ is the number of DNA barcodes). The mean and standard deviation of these distance values are then computed within each subgroup and subsequently aggregated using the mean function across subgroups at each taxonomic rank.

{
The statistics (\autoref{tab:dna_stats}, right columns) indicate that as we progress from higher to lower taxonomic ranks (e.g., from \texttt{phylum} and \texttt{class} to \texttt{genus} and \texttt{species}), both the mean and standard deviation of pairwise genetic distances decrease. This reduction indicates that genetic differences between organisms become smaller as we move down the taxonomic hierarchy, meaning organisms at lower ranks are more genetically similar to each other compared to those at higher ranks. For instance, \texttt{species} within the same \texttt{genus} tend to have much more similar DNA sequences than \texttt{families} within an \texttt{order} or \texttt{orders} within a \texttt{class}. This pattern aligns with the hierarchical structure of biological classification, where organisms are grouped based on increasing genetic relatedness as we move to finer taxonomic levels.
}

{
At the same time, the larger standard deviations observed at higher taxonomic ranks, such as \texttt{class} and \texttt{order}, reflect greater variability in genetic distances, suggesting a broader range of genetic diversity at these levels. Conversely, at lower ranks, such as \texttt{genus} and \texttt{species}, the smaller mean and standard deviation of pairwise distances highlight closer genetic relationships. However, these reduced distances can pose challenges for classification since the differences between closely related species become subtle. 
}

{
This emphasizes the need for finer genetic markers or additional traits beyond pairwise distances to accurately distinguish between organisms, especially at the \texttt{species} level, where genetic distinctions can be minimal. Incorporating multimodal data, such as combining DNA sequences with images, can help address this challenge by providing complementary information. While DNA sequences offer insights into genetic differences, images capture morphological traits that may not be reflected in the genetic data. This multimodal approach can enhance classification accuracy, particularly when distinguishing between closely related species.
}

\begin{figure}[!ht]
	\centering    
	\includegraphics[width=\textwidth]{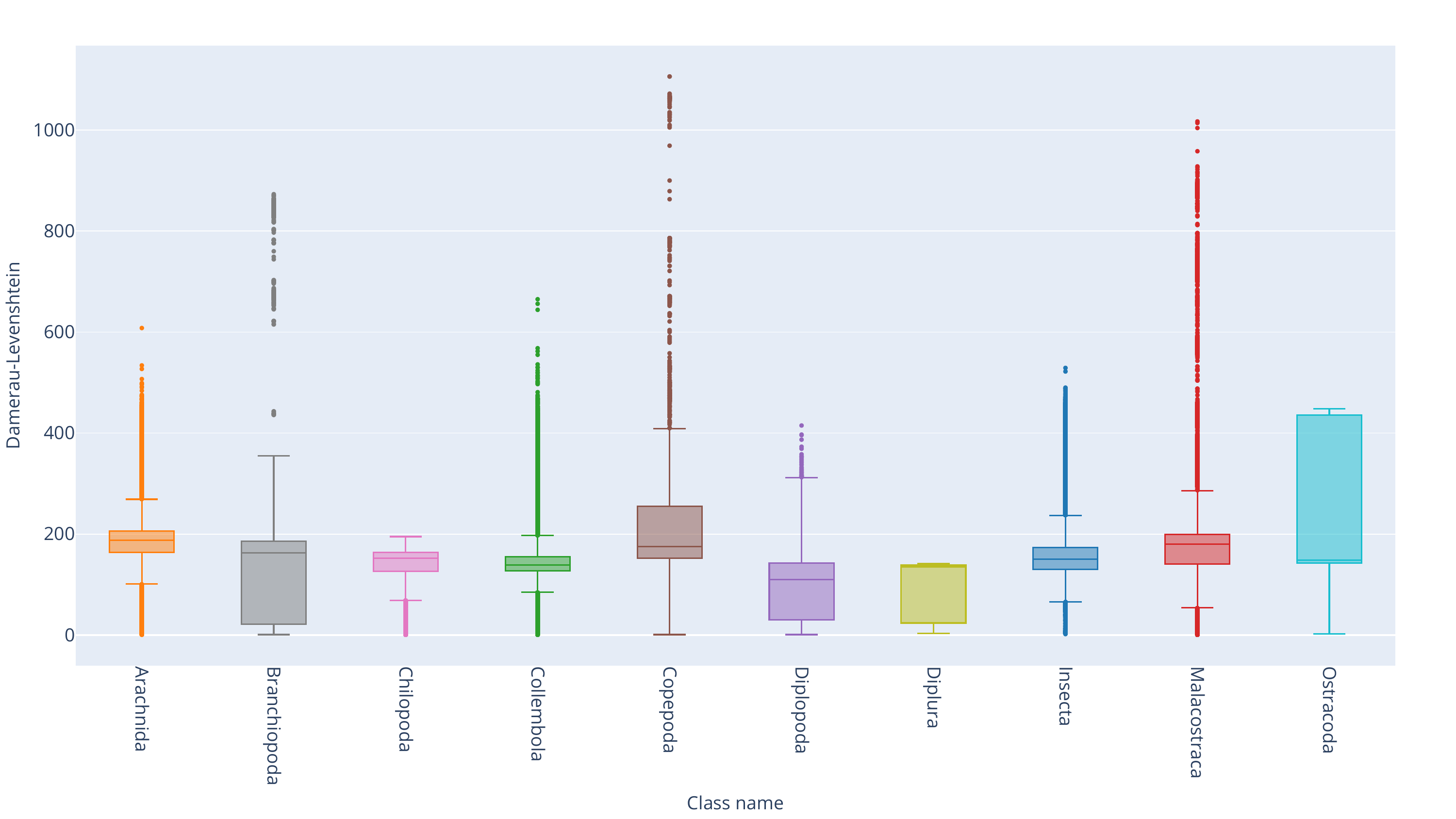}
	\caption{\textbf{Distribution of pairwise distances of subgroups of class.}
		The x-axis shows the subgroup categories sorted alphabetically.}
	\label{fig:dist_distr_class}
\end{figure}

\begin{figure}[!ht]
	\centering    
	\includegraphics[width=\textwidth]{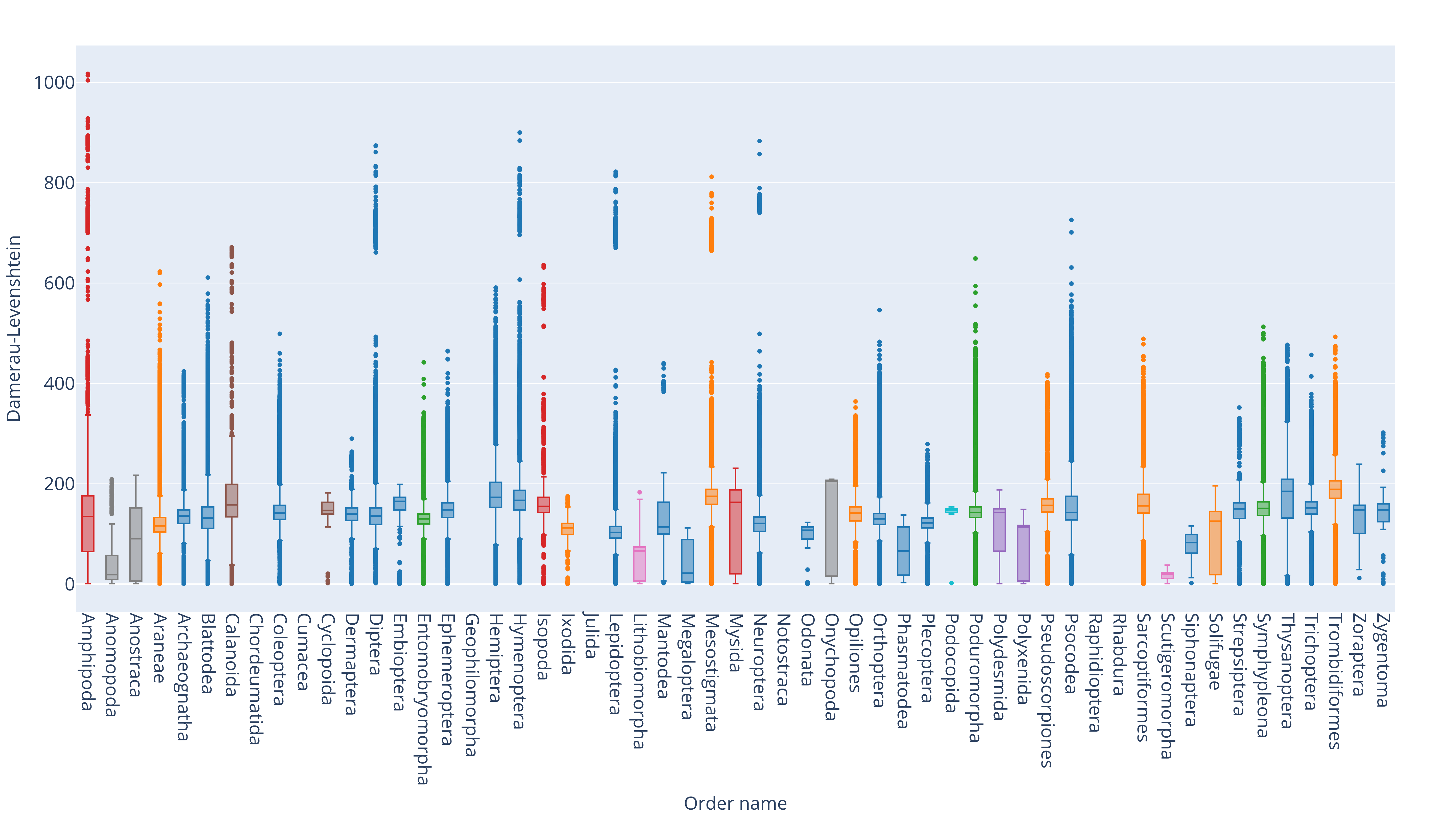}
	\caption{\textbf{Distribution of pairwise distances of subgroups of order.} The x-axis shows the subgroup categories sorted alphabetically.}
	\label{fig:dist_distr_order}
\end{figure}
\begin{figure}[!ht]
	\centering    
	\includegraphics[width=\textwidth]{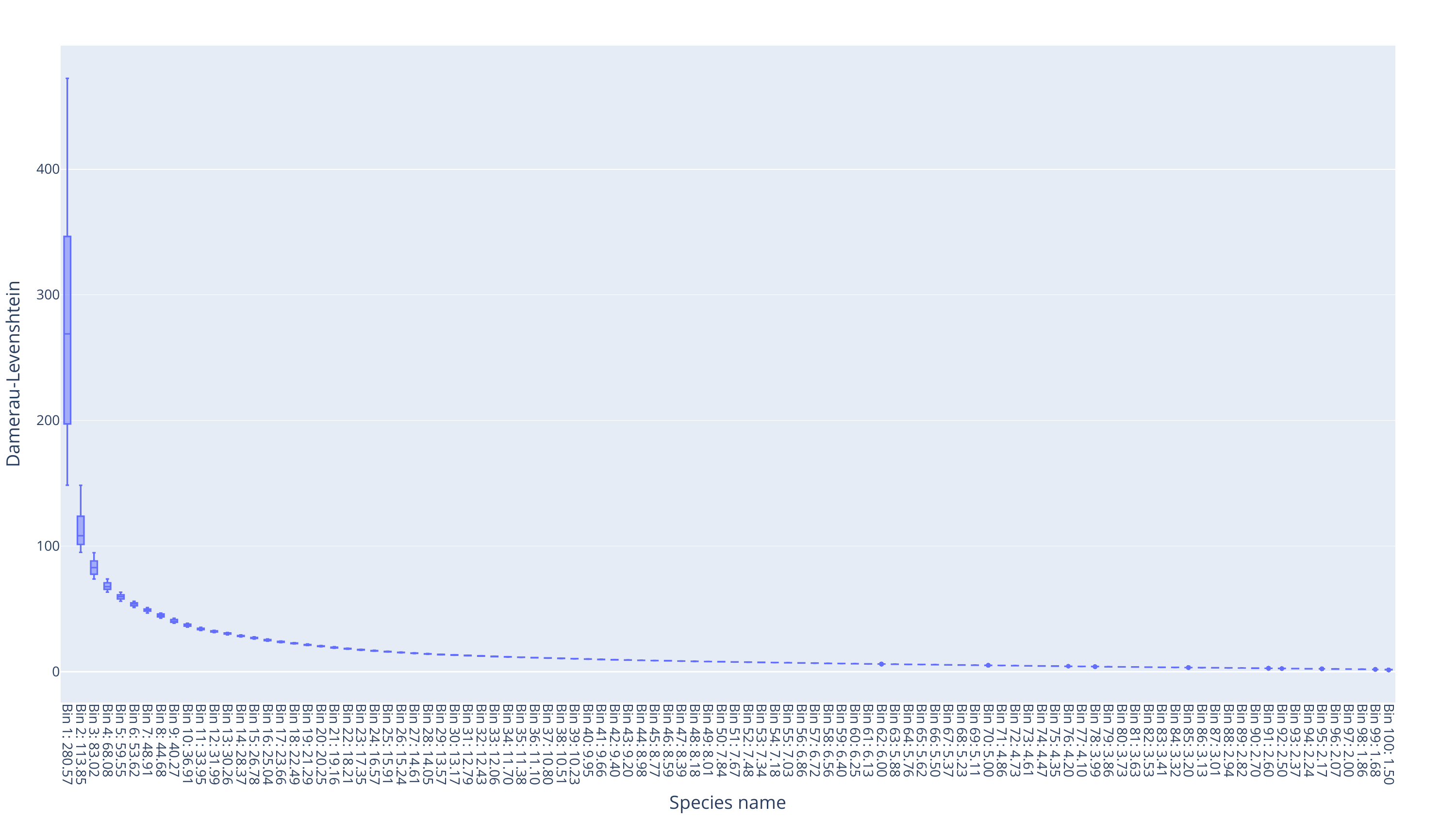}
	\caption{\textbf{Distribution of pairwise distances of subgroups of species.}
		Among the species, there are 8,372 distinct subgroups with sufficient identical barcodes for calculating pairwise distances, which makes visualization challenging. To address this, the groups are sorted in descending order based on their mean distances and partitioned into 100 bins. These bins are used to plot the distribution of pairwise distances within the species rank. The mean distance of each bin is displayed along the x-axis.}
	\label{fig:dist_distr_species}
\end{figure}

\autoref{fig:dist_distr_class}, \autoref{fig:dist_distr_order} and \autoref{fig:dist_distr_species} provide a visual representation of the statistics of pairwise distances computed in \autoref{tab:dna_stats} for taxonomic ranks \texttt{class}, \texttt{order}, and \texttt{species}, respectively. The Interquartile Range (IQR) is a measure of statistical dispersion that describes the range within which the central 50\% of the pairwise distances lies. It is calculated as the difference between the third quartile (\(Q_3\)) and the first quartile (\(Q_1\)) of the data,
\[
\text{IQR} = Q_3 - Q_1,
\]
where \(Q_1\) is the 25th percentile of the data, and \(Q_3\) is the 75th percentile. The line inside the box represents the median (\(Q_2\)) of the data. The height of the box illustrates the IQR. The lines extending from the box (whiskers) indicate the range of the data outside the IQR, typically extending up to 1.5 times the IQR from the quartiles, which help identify the spread of the data.

A small IQR (e.g., Collemboda in \autoref{fig:dist_distr_class}) indicates that the pairwise distances among DNA barcode sequences within the group are tightly clustered around the median, suggesting that the sequences are similar to one another. This homogeneity may imply that the groups consist of closely related species or individuals with minimal genetic divergence, possibly due to a recent common ancestor.

Conversely, a large IQR (e.g., Ostracoda in \autoref{fig:dist_distr_class}) signifies significant variability in the pairwise distances among sequences within a group, indicating a wider range of genetic diversity. This heterogeneity suggests that the groups may encompass genetically diverse species or populations with notable evolutionary divergence. Additionally, the presence of a large IQR may point to potential outliers—sequences that differ substantially from the majority—which could warrant further investigation to understand the underlying genetic variations.

If the whiskers are long while the IQR is small (e.g., Malacostraca in \autoref{fig:dist_distr_class}), it implies that there are outlier values or a wider distribution of data points beyond the central cluster, highlighting the presence of variability in the dataset that may be worth investigating further.

If the median \(Q_2\) is closer to \( Q_1 \) (e.g., Copepoda in \autoref{fig:dist_distr_class}), the distribution is positively skewed, with most data points concentrated at the lower end and fewer but larger values at the higher end. Conversely, if the median is closer to \( Q_3 \) (e.g., Branchiopoda in \autoref{fig:dist_distr_class}), the distribution is negatively skewed, with more values at the higher end and fewer, smaller values at the lower end.

Note that in all taxonomic ranks except for \texttt{species}, a random selection of 1,000 records is made for subgroups with more than 1,000 samples. For the \texttt{species} rank, all subgroups with a large number of records are included in the pairwise distance calculations. Some taxonomic ranks contain extremely large subgroups, such as \textit{Arthropoda} in \texttt{phylum} and \textit{Insecta} in \texttt{class}, each with over 2 million unique DNA records. Consequently, the 1,000 selected records may not fully represent the pairwise distances within the large subgroups. Due to computational limitations—since 1,000 records result in about 500\,k unique pairwise distance computation—we adhere to this rule of selecting a random subset of 1,000 records.

\section{Insect vs non-insect organisms}
\label{a:non-insect}
Focusing on \textit{Insecta} as the most populous group at the \texttt{class} level, we present its detailed statistical records for DNA, BIN, and various taxonomic ranks in \autoref{tab:gen_stat_insect}.

\begin{figure}[!h]
	\centering    
	\includegraphics[width=1\textwidth]{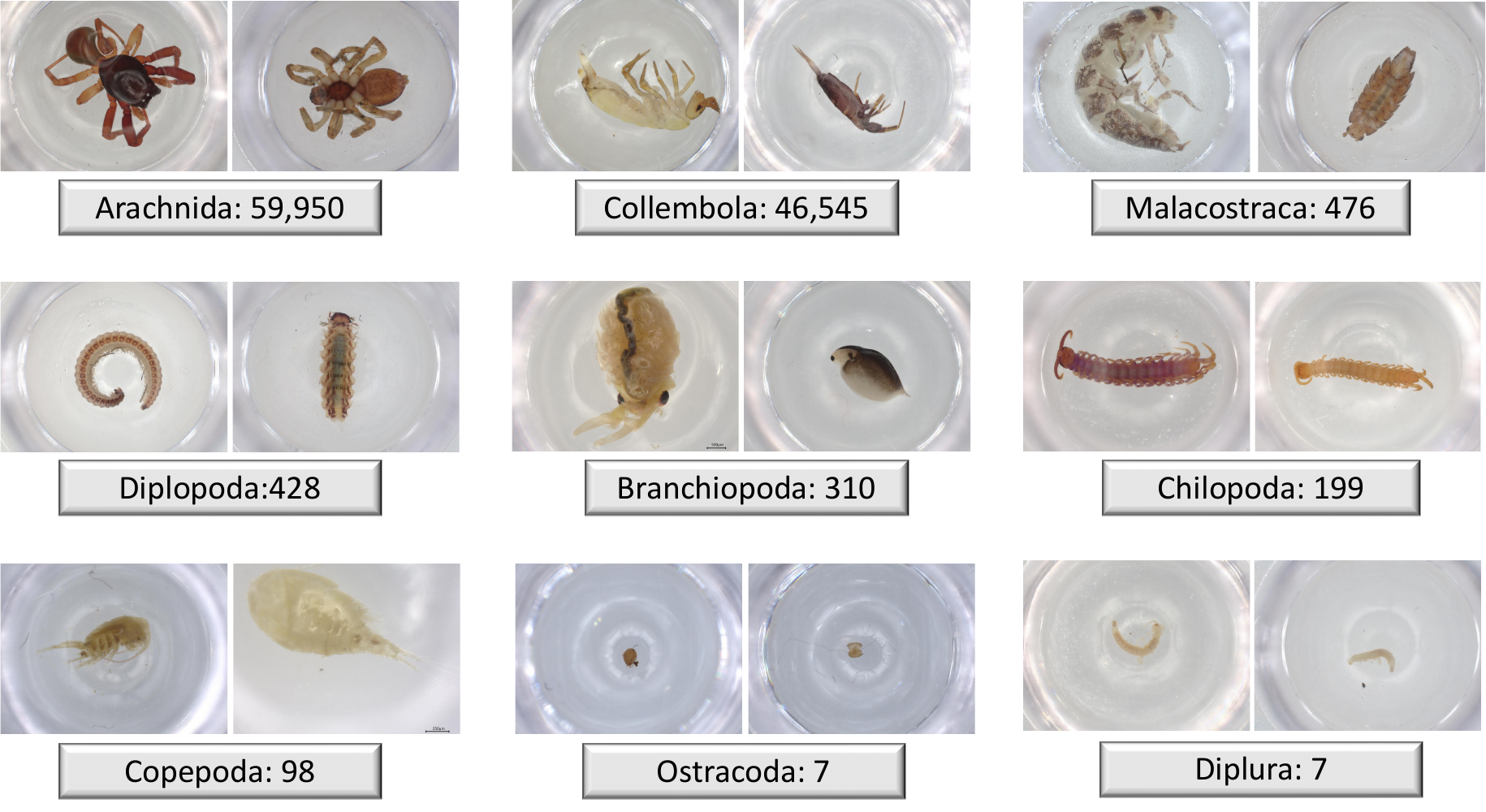}
	\caption{Examples of original images of non-insect organisms from the BIOSCAN-5M dataset. Below each image, the \texttt{class} name and its population within the BIOSCAN-5M dataset are displayed. }
	\label{fig:non_insect}
\end{figure}

\autoref{fig:stat_class} shows the class distribution within the taxonomic rank \texttt{class}, with 99.9\% of organisms labelled at this level, of which 97.8\% belong to the \texttt{class} \textit{Insecta}.
\autoref{fig:non_insect} displays original images of non-insect taxonomic classes from the BIOSCAN-5M dataset, which includes a total of 137,479 organisms.
\begin{table}[!ht]
\begin{center}
\begin{small}
\resizebox{14.0cm}{!}{
\begin{tabular}{lrrrrrrr}
\toprule
\textbf{Attributes} &  \textbf{Categories} &  \textbf{Labelled} &  \textbf{Labelled (\%)} &  \textbf{Unlabelled} &  \textbf{Unlabelled (\%)} & \textbf{IR}  \\
\midrule
\texttt{order}       & \num{      25} & \num{5037247} & \num{ 99.97} & \num{   1571} & \num{           0.03} & \num{1837658} \\
\texttt{family}      & \num{     681} & \num{4853383} & \num{ 96.32} & \num{ 185435} & \num{           3.68} & \num{ 938928} \\
\texttt{subfamily}   & \num{    1305} & \num{1431962} & \num{ 28.42} & \num{3606856} & \num{          71.58} & \num{ 323146} \\
\texttt{genus}       & \num{    6897} & \num{1188043} & \num{ 23.58} & \num{3850775} & \num{          76.42} & \num{ 200268} \\
\texttt{species}     & \num{   21512} & \num{ 450215} & \num{  8.93} & \num{4588603} & \num{          91.07} & \num{   7694} \\
\texttt{taxon}       & \num{   26603} & \num{5038818} & \num{100.00} & \num{      0} & \num{           0.00} & \num{ 925520} \\
\midrule
\texttt{dna\_bin}    & \num{  311743} & \num{5025921} & \num{ 99.74} & \num{  12897} & \num{           0.26} & \num{  35458} \\
\texttt{dna\_barcode}& \num{ 2423704} & \num{5038818} & \num{100.00} & \num{      0} & \num{           0.00} & \num{   3743} \\
\bottomrule
\end{tabular}
}
\end{small}	
\end{center}			
\caption{Detailed statistical records for DNA, BIN and taxonomic ranks within \texttt{class} \textit{Insecta} of the BIOSCAN-5M dataset.}
\label{tab:gen_stat_insect}
\end{table}

\section{Limitations and challenges}
\label{a:limitations}
\subsection{Fine-grained classification}
The BIOSCAN-5M dataset offers detailed biological features for each organism by annotating images with multi-grained taxonomic ranks. The class imbalance ratio (IR) across taxonomic groups reveals significant disparities in sample sizes between the majority class (with the most samples) and minority classes (with fewer samples). Notably, among the 55 distinct \texttt{orders}, \textit{Diptera} accounts for approximately 71\% of the total organisms. \autoref{fig:species} illustrates various \texttt{species} within the \texttt{order} \textit{Diptera}, highlighting the high similarity among images of distinct categories, which poses additional challenges for downstream image classification tasks.

\begin{figure*}[!h]
	\centering    
	\includegraphics[width=1.\textwidth]{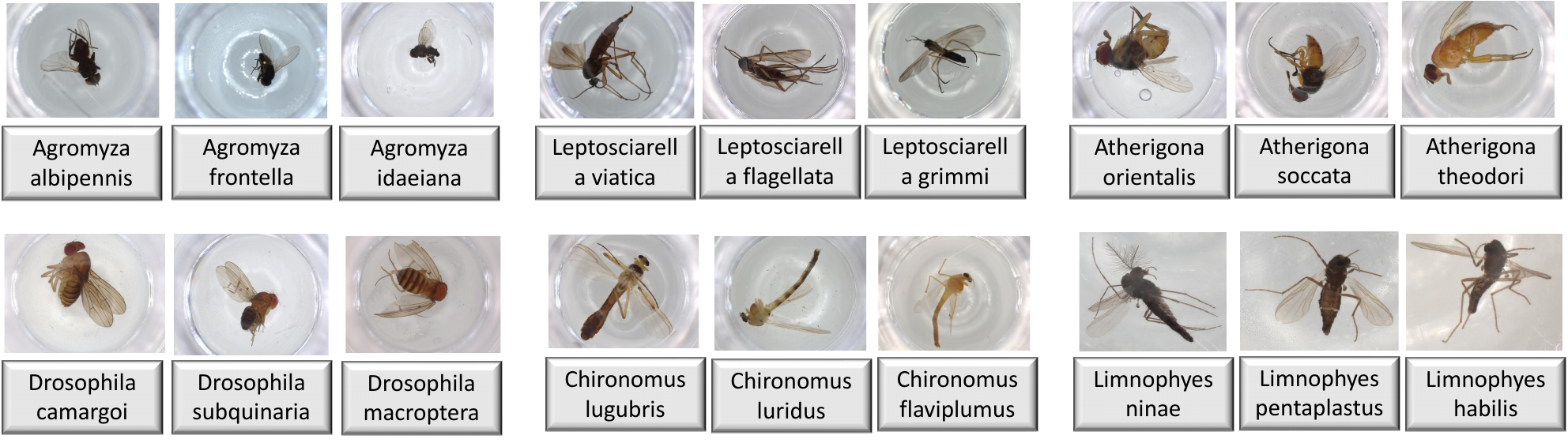}
	\caption{Sample images of distinct \texttt{species} from the \texttt{order} \textit{Diptera}, which comprises about 71\% of BIOSCAN-5M dataset. High similarity between samples of different \texttt{species} highlights significant image classification challenges.}
	\label{fig:species}
\end{figure*}

\subsection{Accessing ground-truth labels}
The BIOSCAN-5M dataset exposes a limitation regarding labelling. The number of labelled records sharply declines as we delve deeper into taxonomic ranking groups, particularly when moving towards finer-grained taxonomic ranks beyond the \texttt{family} level. In fact, over 80\% of the organisms lack taxonomic labels for ranks such as \texttt{subfamily}, \texttt{genus} and \texttt{species}. This circumstance poses a significant challenge for conducting taxonomic classification tasks. However, this limitation also opens doors to opportunities for research in various domains. The abundance of unlabelled data presents avenues for exploration in clustering, unsupervised, semi-supervised, and self-supervised learning paradigms, allowing for innovative approaches to data analysis and knowledge discovery.

\subsection{Sampling Bias}
\label{a:sampling-bias}
The BIOSCAN-5M dataset also exposes a sampling bias as a result of the locations where and the methods through which organisms were collected, as depicted by \autoref{fig:geo-country}.

\section{Data processing}
\label{a:data-processing}
To optimize our benchmark experiments using the BIOSCAN-5M dataset, we implemented two critical pre-processing steps on the raw dataset samples. These steps were necessary to enhance the efficiency and accuracy of our downstream tasks.

The first step involved image cropping and resizing. Due to the high resolution and large size of images in the dataset, processing the original images is both time-consuming and computationally expensive. Additionally, the area around the organism in each image is redundant for our feature extraction. To address these issues, we cropped the images to focus on the region of interest, specifically the area containing the organism. This step eliminated unnecessary background, reducing the data size and focusing the analysis on the relevant parts of the images. After cropping, we resized the images to a standardized resolution, further reducing the computational load and ensuring uniformity across all image samples.

The second step addressed inconsistencies in the taxonomic labels. In the raw dataset, we encountered identical DNA nucleotide sequences labelled differently at certain taxonomic levels, likely due to human error (e.g., typos) or disagreements in taxonomic naming conventions. Such discrepancies posed significant challenges for our classification tasks involving images and DNA barcodes. To address this, we implemented a multi-step cleaning process for the taxonomic labels. We identified and flagged inconsistent labels associated with identical DNA sequences and corrected typographical errors to ensure accurate and consistent naming. 

We present additional details of our pre-processing steps in the following section.

\subsection{Image processing details}
\label{s:image-processing}
The BIOSCAN-5M dataset contains resized and cropped images following the process in BIOSCAN-1M Insect \citep{gharaee2023step}. We resized images to 256\,px on the smaller dimension. 
As in BIOSCAN-1M, we opt to conduct experiments on the cropped and resized images due to their smaller size, facilitating efficient data loading from disk.

\mypara{Cropping.}
Following BIOSCAN-1M \citep{gharaee2023step}, we develop our cropping tool by fine-tuning a DETR \citep{carion2020end} model with a ResNet-50 \citep{he2016deep} backbone (pretrained on MSCOCO, \citealp{lin2014microsoft}) on a small set of 2,837 insect images annotated using the \hreffoot{https://aidemos.cs.toronto.edu/annotation-suite/}{Toronto Annotation Suite}.

For BIOSCAN-1M, the DETR model was fine-tuned using 2,000 insect images (see Section 4.2 of \citealp{gharaee2023step} for details). While the BIOSCAN-1M cropping tool worked well in general, there are some images for which the cropping was poor. Thus, we took the BIOSCAN-1M cropping tool checkpoint, and further fine-tuned the model for BIOSCAN-5M using the same 2,000 images and an additional 837 images that were not well-cropped previously. We followed the same training setup and hyperparameter settings as in BIOSCAN-1M and fine-tuned DETR on one RTX2080 Ti with batch size 8 and a learning rate of 0.0001. 

\begin{table}[tbh]
\caption{We compare the performance of the DETR model we used for cropping that was trained with the extra 837 images (NWC-837) that were previously not well-cropped to the model used for BIOSCAN-1M. We report the Average Precision (AP) and Average Recall (AR) computed on an additional validation set consisting of 100 images that were not-well cropped previously (NWC-100-VAL), as well as the images (IP-100-VAL + IW-150-VAL) used to evaluate the cropping tool's model used in BIOSCAN-1M.  Our updated model performs considerably better on NWC-100-VAL, while given comparable performance on the original validation set of images.}
\label{tab:exp-cropping-results}
\centering
\resizebox{\linewidth}{!}
{
\begin{tabular}{llrrrr}
\toprule
& & \multicolumn{2}{c}{NWC-100-VAL} & \multicolumn{2}{c}{IP-100-VAL + IW-150-VAL} \\ 
\cmidrule(lr){3-4}\cmidrule(lr){5-6}
Dataset & Training data  & AP{[}0.75{]} & AR{[}0.50:0.95{]} & AP{[}0.75{]} & AR{[}0.50:0.95{]} \\ 
\midrule
BIOSCAN-1M & IP-1000 + IW-1000 & 0.257 &  0.485 & \textbf{0.922} & \textbf{0.894}\\
BIOSCAN-5M & IP-1000 + IW-1000 + NWC-837 & \textbf{0.477} & \textbf{0.583} & 0.890 & 0.886\\
\bottomrule
\end{tabular}
}
\end{table}

\autoref{tab:exp-cropping-results} shows that our model with additional data achieves better cropping performance on an evaluation set of 100 images that were previously poorly cropped (NWC-100-VAL). Before cropping, we increase the size of the predicted bounding box by a fixed ratio $R=1.4$ relative to the tight bounding box to capture some of the image background. If the bounding box extends beyond the image's edge, we pad the image with maximum-intensity pixels to align with the white background. These processes are the same as used by \citet{gharaee2023step}. After cropping, we save the cropped-out bounding box.

\mypara{Resizing.}
After cropping the image, we resize the image to 256 pixels on its smaller side while maintaining the aspect ratio ($r=\frac{w}{h}$). 
As nearly all original images are 1024\texttimes768 pixels, our resized images are (nearly all) 341\texttimes256 pixels.

\mypara{Area fraction.}
The \texttt{area\_fraction} field in the metadata file indicates the proportion of the original image represented by the cropped image. This factor is calculated using the bounding box information predicted by our cropping tool and serves as an indicator of the organism's size. \autoref{fig:frac_crop} displays the bounding boxes detected by our cropping model, which we used to crop images in the BIOSCAN-5M dataset. The area fraction factor is calculated as follows:
\begin{equation}
	f_{a} = \frac{w_{c} \, h_{c}}{w\, h}
\end{equation}

\begin{figure}[!h]
	\centering    
	\includegraphics[width=0.8\textwidth]{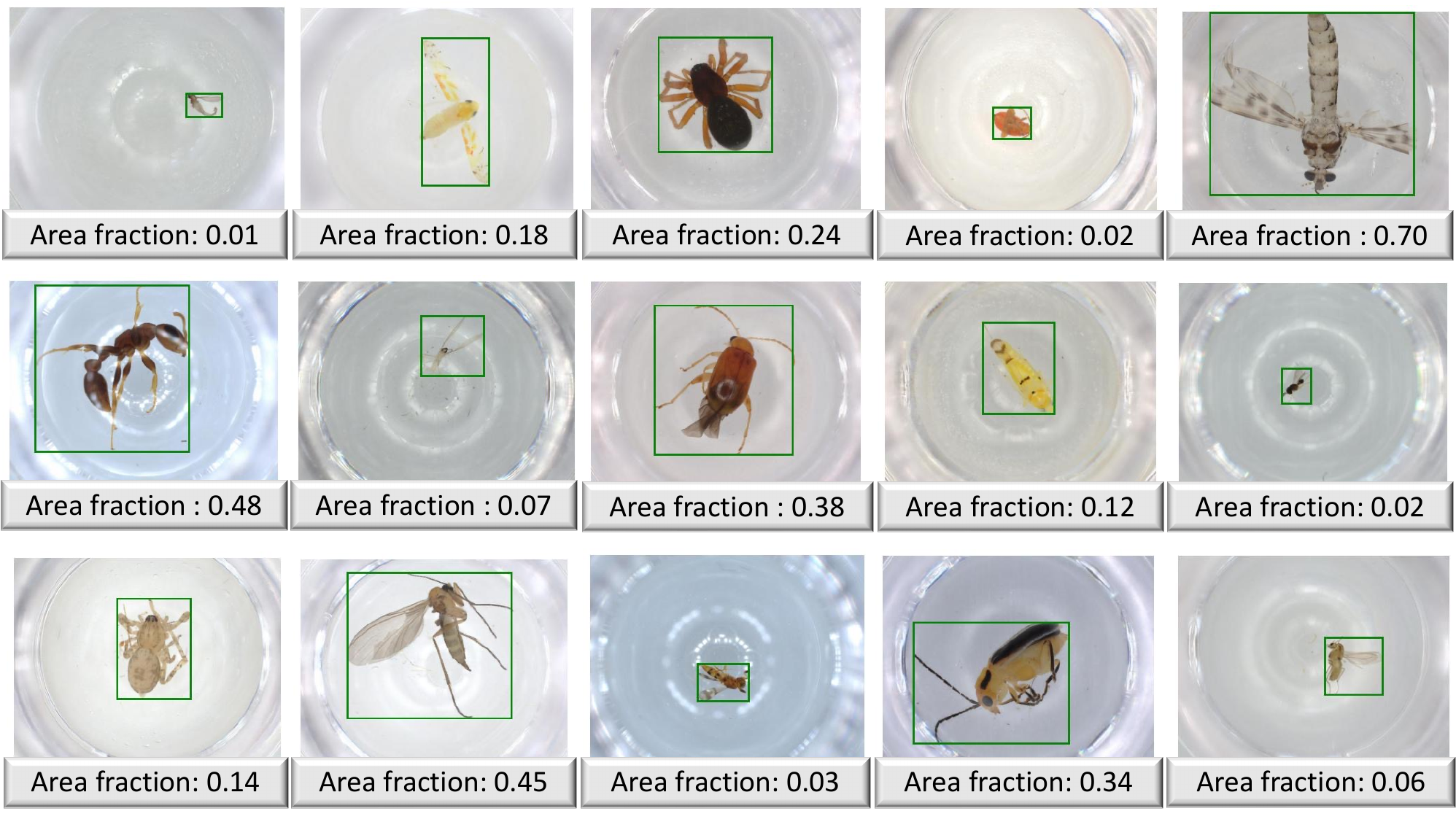}
	\caption{Examples of original images of organisms of the BIOSCAN-5M dataset with the bounding boxes detected by our cropping module. The area fraction value below each image shows how much of the original image is included in the crop. }
	\label{fig:frac_crop}
\end{figure}

\mypara{Scale factor.} 
When capturing images of physical objects, such as medical scans or biological samples, it is essential to ensure that measurements derived from these images accurately represent the real objects. To compute real-world sizes from captured images, a consistent relationship between pixel size and physical size is necessary. Therefore, we introduced the \texttt{scale\_factor} field in the metadata file, which defines the ratio between the cropped image (\texttt{cropped}) and the cropped and resized image (\texttt{cropped\_256}).

Assuming the original images ($I$) have constant dimensions, width ($w$) and height ($h$), the cropped images ($I_c$) are extracted using bounding box information from our cropping tool and have varying widths ($w_c$) and heights ($h_c$) proportional to the size of the organism. The resized images ($I_r$) are adjusted so that the shorter dimension, either width ($w_r$) or height ($h_r$), is set to a constant size of 256 px, while the other dimension is scaled proportionally to maintain the aspect ratio, resulting in a dimension greater than 256 px.

We calculated the scale-factor ($f_s$) as follows:
\begin{equation}
    f_s = \frac{\min(w_{c}, h_{c})}{256}
\end{equation}
If we define the pixel scale as the number of millimeters per pixel, then the pixel scale of the cropped and resized image (\texttt{cropped\_256}) is equivalent to the pixel scale of the original image multiplied by the scale factor:
\begin{equation}\label{eq:scale_factor}
    \text{pixel\_scale}_{\text{cropped\_256}} = \text{pixel\_scale}_{\text{original}} \times f_s
\end{equation}

Note the pixel scale of the original image remains unchanged during the cropping process, as cropping only involves cutting out areas around the region of interest (the organism) without scaling the image.

The original images were captured using a Keyence VHX-7000 Digital Microscope system imaging system at a resolution of 2880\texttimes2160 pixels. These images were then resized to a resolution of 1024\texttimes768 pixels to obtain the original images (\texttt{original\_full}) of the BIOSCAN-5M dataset. Each pixel in the raw images represents a physical space of 2.95\,\textmu m by 2.95\,\textmu m. Using  this pixel scale and the scale factor obtained from \autoref{eq:scale_factor}, we can estimate the size of the object in the real-world.

\subsection{HDF5 file}
\label{a:hdf5-file}
To load data efficiently during the training of the \clibd baseline, we also generated a 190\,GB HDF5 file to store images and related metadata from the BIOSCAN-5M dataset. This file is structured to allow rapid access and processing of large-scale data.

At the top level, the file consists of a \emph{group} of the following \emph{datasets} representing different partitions of BIOSCAN-5M. Each partition includes keys or queries for one or all of the splits (pretraining, validation, or test).

For more information or to download the HDF5 file, please see the instructions at the \clibd GitHub repo: \url{https://github.com/bioscan-ml/clibd}.

\subsection{Taxa of unassigned placement}
\label{a:unassigned-taxa}

Some taxonomic labels had ``holes'' in them due to the complexities of the definition of taxonomic labels.
Established taxonomic labels for some species can omit taxonomic ranks because there is currently no scientific need to define a grouping at that taxonomic level.

In particular, we found there were \num{1448} genera which were missing a subfamily label because their genus had not been grouped into a subfamily by the entomological community.
Note that these genera might at some point in the future be assigned a subfamily, if a grouping of genera within the same family becomes apparent.

This situation of mixed rankings creates a complexity for hierarchical modelling, which for simplicity typically assumes a rigid structure of level across the labelling tree for each sample.
We standardized this by adding a placeholder subfamily name where there was a hole, equal to ``unassigned <Family\_name>''.
For example, for the genus Alpinosciara, the taxonomic label was originally:

{
\small
\quad\texttt{Arthropoda > Insecta > Diptera > Sciaridae > [none] > Alpinosciara}
}

\noindent and after filling the missing subfamily label, it became:

{
\small
\quad\texttt{Arthropoda > Insecta > Diptera > Sciaridae > unassigned Sciaridae > Alpinosciara}
}

\noindent This addition ensures that the mapping from genus to subfamily is injective, and labels which are missing because they are not taxonomically defined are not confused with labels which are missing because they have not been identified.
Furthermore, this ensures that each subsequent rank in the taxonomic labels provides a partitioning of each of the labels in the rank that proceeds it.

\subsection{Taxonomic label cleaning}
\label{s:label-cleaning}

The taxonomic labels were originally entered into the BOLD database by expert entomologists using a drop-down menu for existing species, and typed-in manually for novel species.
Manual data entry can sometimes go awry.
We were able to detect and resolve some typographical errors in the manual annotations, as described below.

\mypara{Genus and species name comparison.}
Since species names take the form ``<Genus\_name> <species\_specifier>'', the genus is recorded twice in samples which possess species labels.
This redundancy provides an opportunity to provide a level of quality assurance on the genus-level annotations
A few samples (82 samples across 13 species) had a species label but no genus label; for these we used the first word of the species label as the genus label.
For the rest of the samples with a species label, we compared the first word of the species label with the genus label, and resolved 166 species where these were inconsistent.
These corrections also uncovered cases where the genus name was entered incorrectly more broadly, and we were able to correct genera values which were entered incorrectly even in cases where they were consistent with their species labels or had no species labels.

\mypara{Conflicted annotations for the same barcode.}
We found many DNA barcodes were repeated across the dataset, with multiple images bearing the same barcode.
Overall, there were on average around two repetitions per unique barcode in the dataset.
It is already well-established that the COI mitochondrial DNA barcode is a (sub)species-level identifier, i.e. same barcode implies same species, and different species implies different barcodes \citep{moritz2004dna,sokal1963principles,blaxter2005defining}.
Hence we have a strong prior that samples with the same barcode should be samples of the same species.
This presents another opportunity to provide quality assurance on the data, by comparing the taxonomic annotations across samples which shared a DNA barcode.
Differences can either arise by typographical errors during data entry, or by differences of opinion between annotators.

We investigated cases where completed levels of the taxonomic annotations differed for the same barcode.
This indicated some common trends as values often compared as different due to stylistic differences, where one annotation differed only by casing, white-space, the absence of a 0 padding digit to an identifier code number, or otherwise misspellings.
We resolved some such disagreements automatically, by using the version more common across the dataset.

The majority of placeholder genus and species names follow one of a couple of formats such as ``<Genus\_name> Malaise1234'', e.g. ``\textit{Oxysarcodexia Malaise4749}''.
Comparing different taxonomic annotations of the same barcode only allows us to find typos where a barcode has been annotated more than once.
However, there are of course more barcodes than species and so there may remain some typos which make two samples of the same species with different barcodes compare as different when they should be the same.
To address this, we found labels which deviated from the standardized placeholder name formats and modified them to fit the standardized format.
Examples of these corrections include adding missing zero-padding on digits, fixing typos of the word ``Malaise'', and inconsistent casing.
In this way, we renamed the species of \num{6756} samples and genus of \num{3675} across \num{7673} records.

We resolved the remaining conflicts between differently annotated samples of the same barcode as follows.
We considered each taxonomic rank one at a time.
In cases where there was a conflict between the annotations, we accepted the majority value if at least 90\% of the annotations were the same.
If the most common annotation was less prevalent than this, we curtailed the annotation at the preceding rank.
Curtailed annotations which ended at a filler value (i.e. a subfamily name of the format ``unassigned <Family\_name>'') were curtailed at the last completed rank instead.
In total, we dropped at least one label from \num{3478} records.

Next, we considered barcodes whose multiple annotations differed in their granularity.
In such cases, we inferred the annotations for missing taxonomic ranks from the samples that were labelled to a greater degree of detail.
In total, we inferred at least one label for \num{172895} records.
We believe these inferred labels are unlikely to have an error rate notably higher than that of the rest of the data.
Even so, we provide details about which ranks were inferred in the metadata field \texttt{inferred\_ranks} in case the user wishes to exclude the inferred labels.
This field takes the following values:
\begin{itemize}
\item 0 --- Original label only (nothing inferred).
\item 1 --- Species label was copied. (Sample was originally labelled to genus-level.)
\item 2 --- Genus and (if present) species labels were copied.
\item 3 --- Subfamily, and every rank beneath it, were copied.
\item 4 --- Family, and every rank beneath it, were copied.
\item 5 --- Order, and every rank beneath it, were copied.
\item 6 --- Class, and every rank beneath it, were copied.
\end{itemize}

\mypara{Non-uniquely identifying species names.}
Finally, we noted that some species names were not unique identifiers for a species.
Theses cases arise where an annotator has used \emph{open nomenclature} to indicate a suspected new species, e.g. ``\textit{Pseudosciara sp.}'', ``\textit{Olixon cf. testaceum}'', and ``\textit{Dacnusa nr. faeroeensis}''.
Since this is not a uniquely identifying placeholder name for the species, it is unclear whether two instances with the same label are the same new species or different new species.
For example, there were \num{1247} samples labelled as ``\textit{Pseudosciara sp.}'', and these will represent a range of new species within the Pseudosciara genus, and not repeated observations of the same new species.
Consequently, we removed such species annotations which did not provide a unique identifier for the species.
In total, \num{198} such species values were removed from \num{5101} samples.

\mypara{Conclusion.}
As a result of this cleaning process we can make the following claims about the dataset, with a high degree of confidence:
\begin{itemize}
    \item All records with the same barcode have the same annotations across the taxonomic hierarchy.
    \item If two samples possess a species annotation and their species annotation is the same, they are the same species. (Similarly for genus level annotations, etc.)
    \item If two samples possess a species annotation and their species annotations differ, they are not the same species. (Similarly for genus level annotations, etc.)
\end{itemize}

\begin{table}[tb]
\caption{Example species from each species set.}
\label{tab:species-set-examples}
\centering
\small
\begin{tabular}{lllrrrr}
\toprule
            &       &         & \multicolumn{4}{c}{\textbf{Number of samples}} \\
\cmidrule(lr){4-7}
            &       &         &     &\textbf{Train/}&     &      \\
\textbf{Species set} & \textbf{Genus} & \textbf{Species} & \textbf{All} & \textbf{Keys} & \textbf{Val} & \textbf{Test} \\
\midrule
seen    & Aacanthocnema    & Aacanthocnema dobsoni            &   3 &   3 &   0 &   0 \\
        & Glyptapanteles   & Glyptapanteles meganmiltonae     &  65 &  45 &   2 &  18 \\
        & Megaselia        & Megaselia lucifrons              & 699 & 640 &  34 &  25 \\
        & Pseudomyrmex     & Pseudomyrmex simplex             & 378 & 335 &  18 &  25 \\
        & Rhopalosiphoninus& Rhopalosiphoninus latysiphon     & 148 & 116 &   7 &  25 \\
        & Stenoptilodes    & Stenoptilodes brevipennis        &  16 &  10 &   1 &   5 \\
        & Zyras            & Zyras perdecoratus               &  10 &   6 &   0 &   4 \\
\addlinespace
unseen  & Anastatus        & Anastatus sp. GG28               &  42 &  24 &   6 &  12 \\
        & Aristotelia      & Aristotelia BioLep531            &  87 &  51 &  13 &  23 \\
        & Glyptapanteles   & Glyptapanteles Whitfield155      &  11 &   6 &   1 &   4 \\
        & Megaselia        & Megaselia BOLD:ACN5814           &  24 &  13 &   3 &   8 \\
        & Orthocentrus     & Orthocentrus Malaise5315         &  39 &  23 &   5 &  11 \\
        & Phytomyptera     & Phytomyptera Janzen3550          &  14 &   8 &   1 &   5 \\
        & Zatypota         & Zatypota alborhombartaDHJ03      &   9 &   8 &   1 &   0 \\
\addlinespace
heldout & Basileunculus    & Basileunculus sp. CR3            & 268 &     &     &     \\
        & Cryptophilus     & Cryptophilus sp. SAEVG Morph0281 &  55 &     &     &     \\
        & Glyptapanteles   & Glyptapanteles Malaise2871       &   1 &     &     &     \\
        & Odontofroggatia  & Odontofroggatia corneri-MIC      &  13 &     &     &     \\
        & Palmistichus     & Palmistichus ixtlilxochitliDHJ01 & 416 &     &     &     \\
        & gelBioLep01      & gelBioLep01 BioLep3792           &  16 &     &     &     \\
        & microMalaise01   & microMalaise01 Malaise1237       &  13 &     &     &     \\
\bottomrule
\end{tabular}
\end{table}

\section{Dataset partitioning --- Additional details}
\label{s:partitioning-extra}

\subsection{Species sets}
\label{a:species-sets}
As summarized in \iftoggle{arxiv}{\autoref{sec:partitioning}}{\S4.1 of the main text}, we first partitioned the data based on their species label into four categories as follows:
\begin{itemize}
\item \emph{Unknown}: samples without a species label (note: these may truly belong in any of the other three categories).
\item \emph{Seen}: all samples whose species label is an established scientific name of a species. Species which did not begin with a lower case letter, contain a period, contain numerals, or contain ``malaise'' (case insensitive) were determined to be labelled with a catalogued, scientific name for their species, and were placed in the \emph{seen} set.
\item \emph{Unseen}: Of the remaining samples, we considered the placeholder species which we were most confident were labelled reliably. These were species outside the seen species, but the genus occurred in the seen set. Species which satisfied this property and had at least 8 samples were placed in the \emph{unseen} set.
\item \emph{Heldout}: The remaining species were placed in \emph{heldout}. The majority of these have a placeholder genus name as well as a placeholder species name, but some have a scientific name for their genus name.
\end{itemize}

This partitioning ensures that the task that is posed by the dataset is well aligned with the task that is faced in the real-world when categorizing insect samples.
Example species for each species set are shown in \autoref{tab:species-set-examples}, and the number of categories for each taxonomic rank are shown in \autoref{tab:species-set-summary}.

\begin{table}[tbh]
\caption{Number of (non-empty) categories for each taxa, per species set.}
\label{tab:species-set-summary}
\centering
\small
\begin{tabular}{lrrrrrrr}
\toprule
\textbf{Species set}  & \textbf{Phylum} & \textbf{Class} & \textbf{Order} & \textbf{Family} & \textbf{Subfamily} & \textbf{Genus} & \textbf{Species} \\
\midrule
{unknown}      & \num{    1} & \num{   10} & \num{   52} & \num{  869} & \num{ 1235} & \num{ 4260} & \num{    0} \\
{seen}         & \num{    1} & \num{    9} & \num{   42} & \num{  606} & \num{ 1147} & \num{ 4930} & \num{11846} \\
{unseen}       & \num{    1} & \num{    3} & \num{   11} & \num{   64} & \num{  118} & \num{  244} & \num{  914} \\
{heldout}      & \num{    1} & \num{    4} & \num{   22} & \num{  188} & \num{  381} & \num{ 1566} & \num{ 9862} \\
\midrule
overall        & \num{    1} & \num{   10} & \num{   55} & \num{  934} & \num{ 1542} & \num{ 7605} & \num{22622} \\
\bottomrule
\end{tabular}
\end{table}

\subsection{Splits}
\label{a:partitioning}
To construct partitions appropriate for a closed world training and evaluation scenario, we partitioned the seen data into \texttt{train}, \texttt{val}, and \texttt{test} partitions.
Because many of the DNA barcodes have more than one sample (i.e. multiple images per barcode), we partitioned the data at the barcode level.
The data was highly imbalanced, so to ensure the \texttt{test} partition had high sample efficiency, we flattened the distribution for the \texttt{test} set.
For each species with at least 2 barcodes and at least 8 samples, we selected barcodes to place in the \texttt{test} set.
We tried to place a number of samples in the \texttt{test} set which scaled linearly with the number of samples for the species, starting with a minimum of 4, and capped at a maximum of 25 (reached at 92 samples total).
The target increased at a rate of $\nicefrac{1}{4}$.
We capped the number of barcodes to place in the \texttt{test} set at a number that increased linearly with the number of barcodes for the species, starting at 1 and increasing at a rate of $\nicefrac{1}{3}$.
This flattened the distribution across species in the \texttt{test} set, as shown in Figures \ref{fig:transfer-samples-test}, \ref{fig:transfer-barcodes-test}, and \ref{fig:partition-dist-test}.

\begin{table}[tbh]
\caption{Number of (non-empty) categories for each taxa, per partition.}
\label{tab:split_summary2}
\centering
\small
\begin{tabular}{lrrrrrrr}
\toprule
\textbf{Partition}  & \textbf{Phylum} & \textbf{Class} & \textbf{Order} & \textbf{Family} & \textbf{Subfamily} & \textbf{Genus} & \textbf{Species} \\
\midrule
\texttt{pretrain}       & \num{    1} & \num{   10} & \num{   52} & \num{  869} & \num{ 1235} & \num{ 4260} & \num{    0} \\
\texttt{train}          & \num{    1} & \num{    9} & \num{   42} & \num{  606} & \num{ 1147} & \num{ 4930} & \num{11846} \\
\texttt{val}            & \num{    1} & \num{    5} & \num{   27} & \num{  350} & \num{  598} & \num{ 1704} & \num{ 3378} \\
\texttt{test}           & \num{    1} & \num{    6} & \num{   27} & \num{  352} & \num{  594} & \num{ 1736} & \num{ 3483} \\
\texttt{key\_unseen}    & \num{    1} & \num{    3} & \num{   11} & \num{   64} & \num{  118} & \num{  244} & \num{  914} \\
\texttt{val\_unseen}    & \num{    1} & \num{    3} & \num{   11} & \num{   62} & \num{  116} & \num{  240} & \num{  903} \\
\texttt{test\_unseen}   & \num{    1} & \num{    3} & \num{   11} & \num{   62} & \num{  113} & \num{  234} & \num{  880} \\
\texttt{other\_heldout} & \num{    1} & \num{    4} & \num{   22} & \num{  188} & \num{  381} & \num{ 1566} & \num{ 9862} \\
\midrule
overall                 & \num{    1} & \num{   10} & \num{   55} & \num{  934} & \num{ 1542} & \num{ 7605} & \num{22622} \\
\bottomrule
\end{tabular}
\end{table}

\begin{figure}[tbh]
    \centering
    \begin{subfigure}{0.49\textwidth}%
    \vspace{2mm}%
    \includegraphics[width=\linewidth]{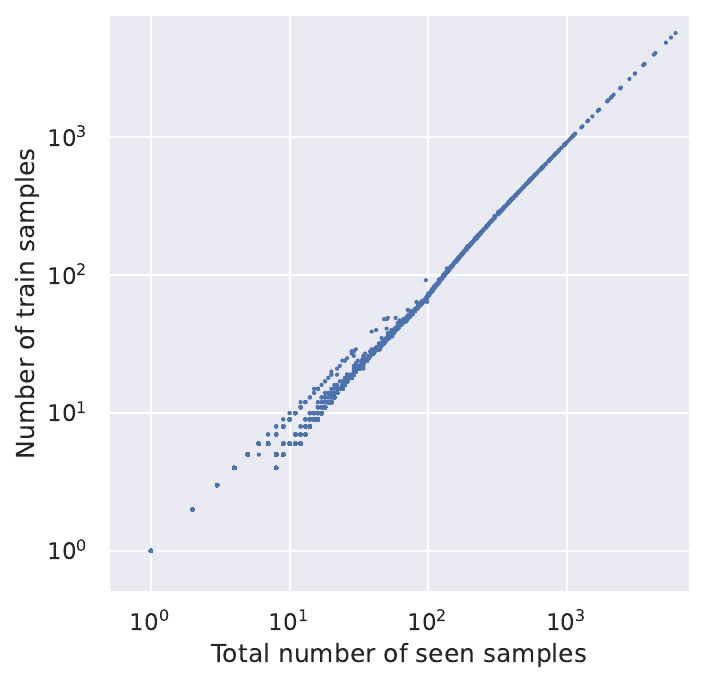}%
    \vspace{-2mm}%
    \caption{\texttt{train} partition.}
    \end{subfigure}
    \hfill
    \begin{subfigure}{0.49\textwidth}%
    \vspace{2mm}%
    \includegraphics[width=\linewidth]{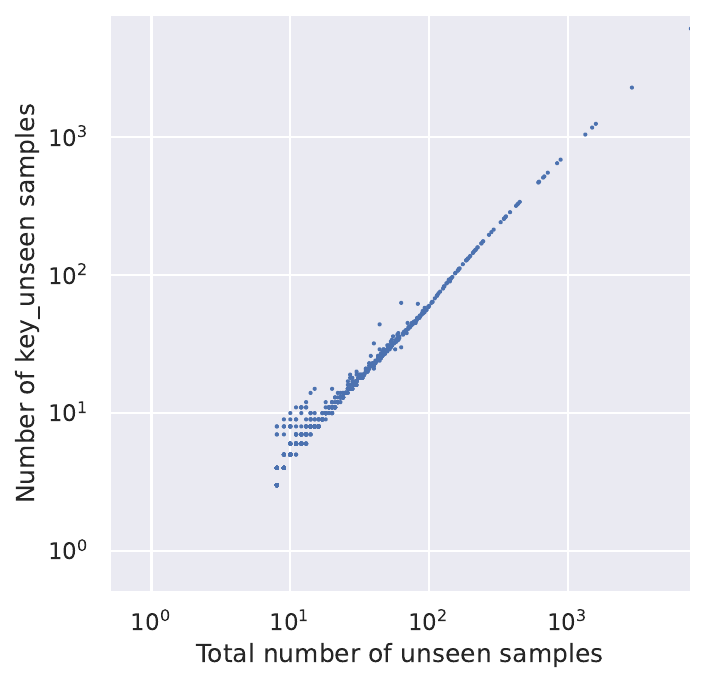}%
    \vspace{-2mm}%
    \caption{\texttt{key\_unseen} partition.}
    \end{subfigure}
    \\
    \begin{subfigure}{0.49\textwidth}%
    \vspace{2mm}%
    \includegraphics[width=\linewidth]{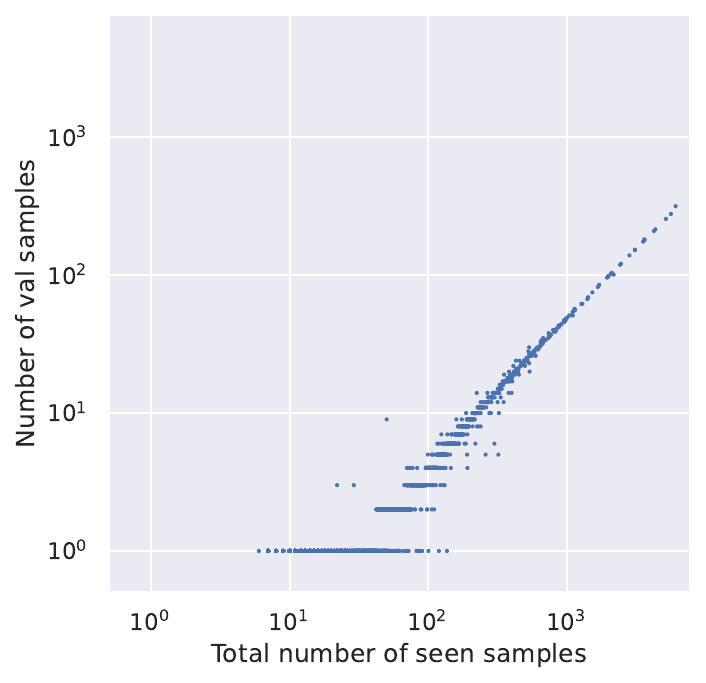}%
    \vspace{-2mm}%
    \caption{\texttt{val} partition.}
    \end{subfigure}
    \hfill
    \begin{subfigure}{0.49\textwidth}%
    \vspace{2mm}%
    \includegraphics[width=\linewidth]{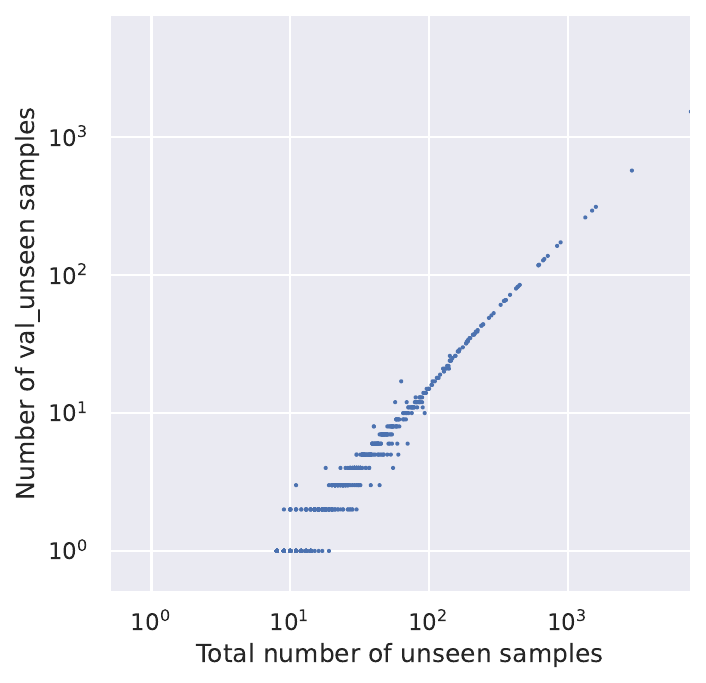}%
    \vspace{-2mm}%
    \caption{\texttt{val\_unseen} partition.}
    \end{subfigure}
    \\
    \begin{subfigure}{0.49\textwidth}%
    \vspace{2mm}%
    \includegraphics[width=\linewidth]{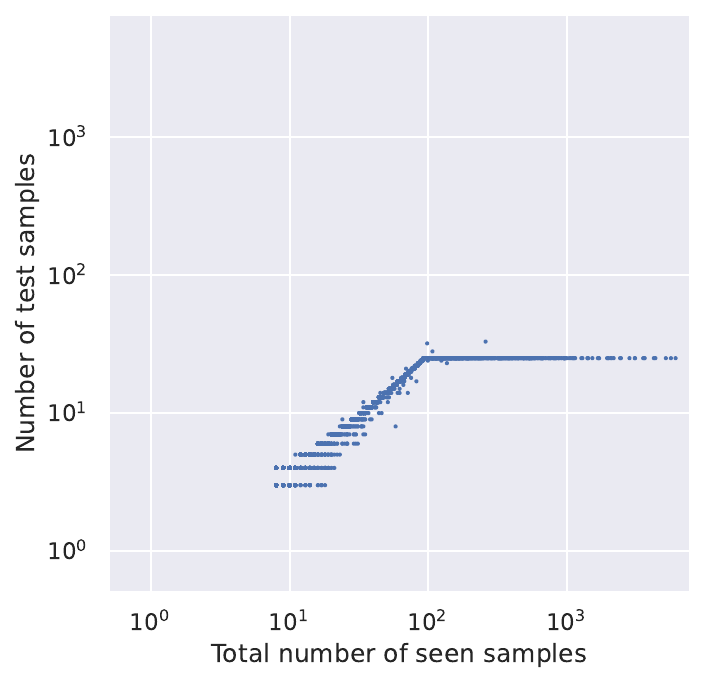}%
    \vspace{-2mm}%
    \caption{\texttt{test} partition.}
    \label{fig:transfer-samples-test}
    \end{subfigure}
    \hfill
    \begin{subfigure}{0.49\textwidth}%
    \vspace{2mm}%
    \includegraphics[width=\linewidth]{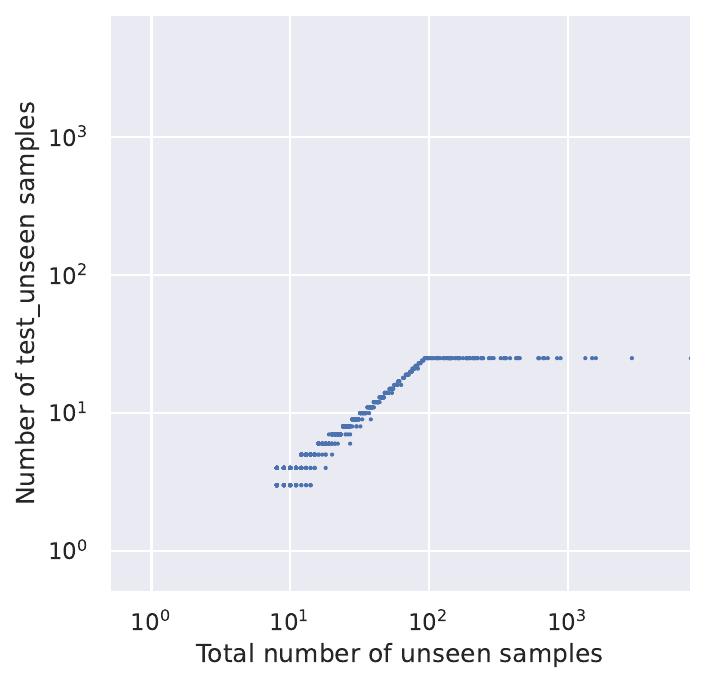}%
    \vspace{-2mm}%
    \caption{\texttt{test\_unseen} partition.}
    \end{subfigure}
    \caption{Number of samples in species set and partition, per species.}
    \label{fig:partition-transfer-samples}
\end{figure}

\begin{figure}[tbh]
    \centering
    \begin{subfigure}{0.49\textwidth}%
    \vspace{2mm}%
    \includegraphics[width=\linewidth]{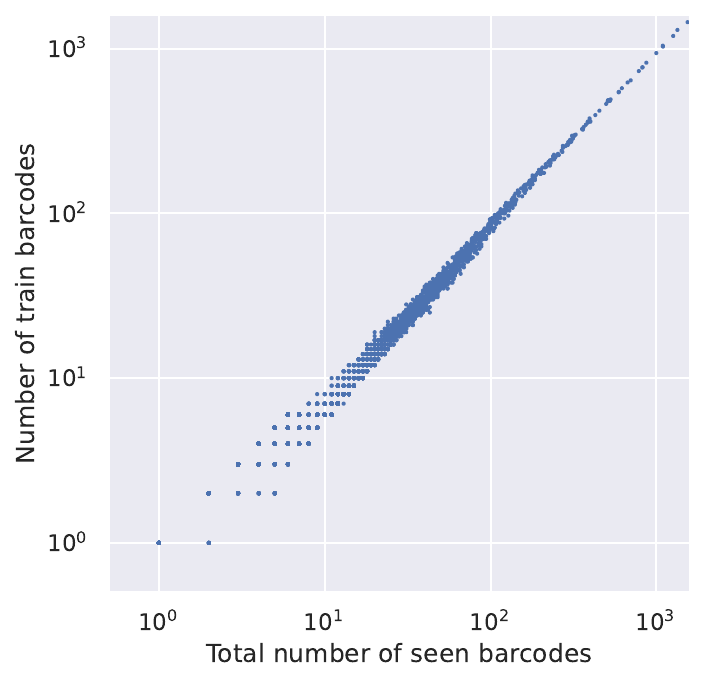}%
    \vspace{-2mm}%
    \caption{\texttt{train} partition.}
    \end{subfigure}
    \hfill
    \begin{subfigure}{0.49\textwidth}%
    \vspace{2mm}%
    \includegraphics[width=\linewidth]{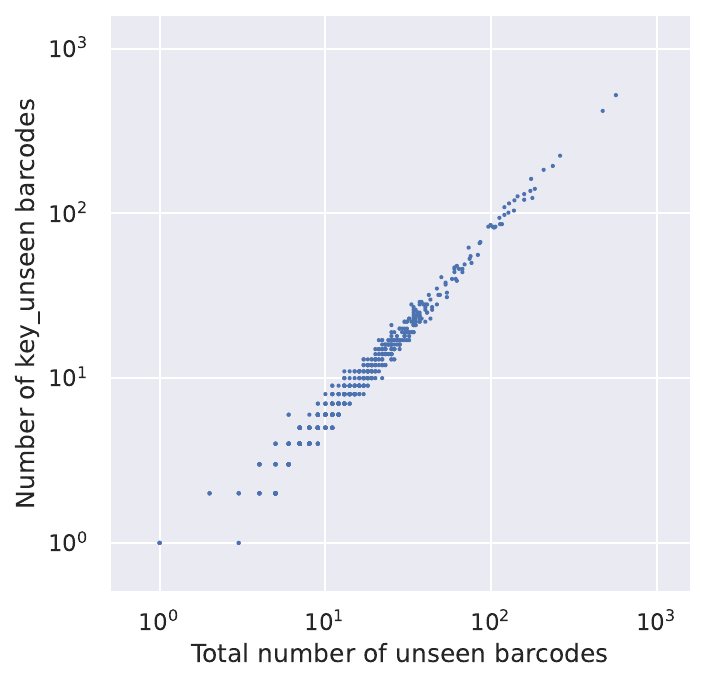}%
    \vspace{-2mm}%
    \caption{\texttt{key\_unseen} partition.}
    \end{subfigure}
    \\
    \begin{subfigure}{0.49\textwidth}%
    \vspace{2mm}%
    \includegraphics[width=\linewidth]{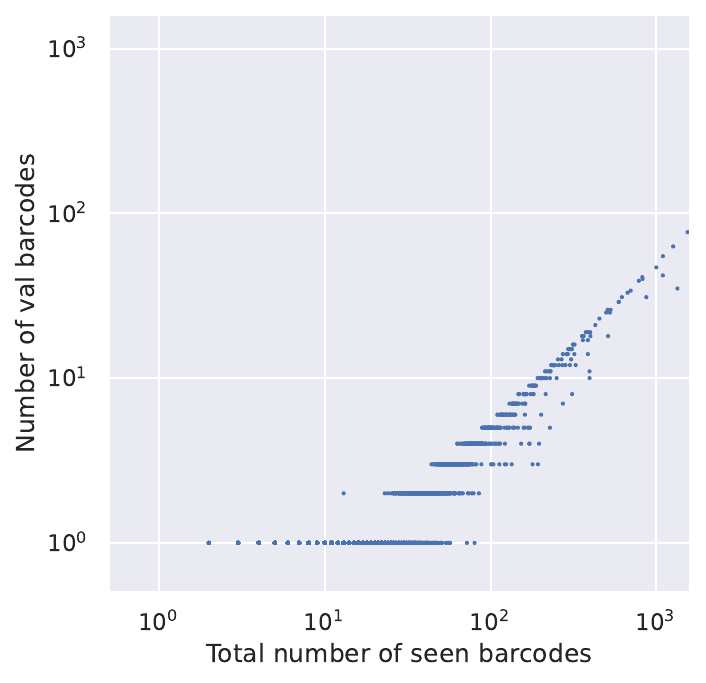}%
    \vspace{-2mm}%
    \caption{\texttt{val} partition.}
    \end{subfigure}
    \hfill
    \begin{subfigure}{0.49\textwidth}%
    \vspace{2mm}%
    \includegraphics[width=\linewidth]{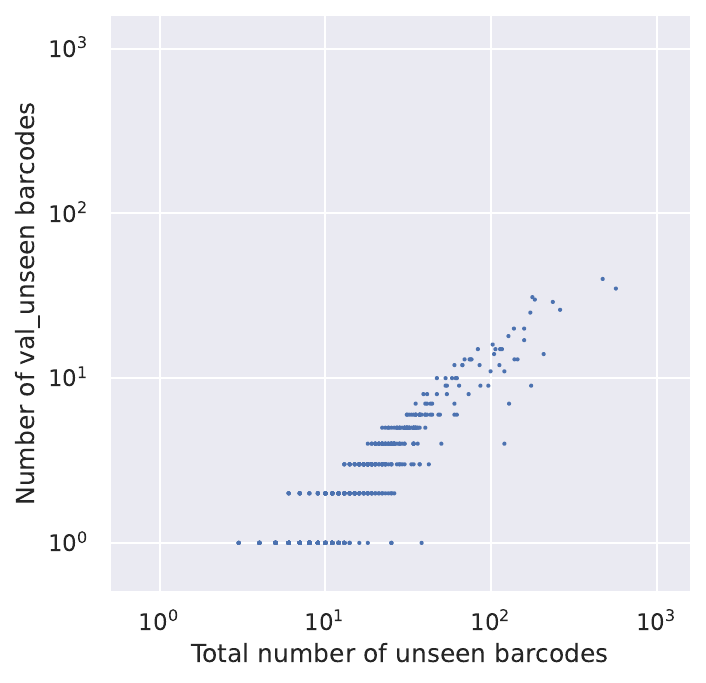}%
    \vspace{-2mm}%
    \caption{\texttt{val\_unseen} partition.}
    \end{subfigure}
    \\
    \begin{subfigure}{0.49\textwidth}%
    \vspace{2mm}%
    \includegraphics[width=\linewidth]{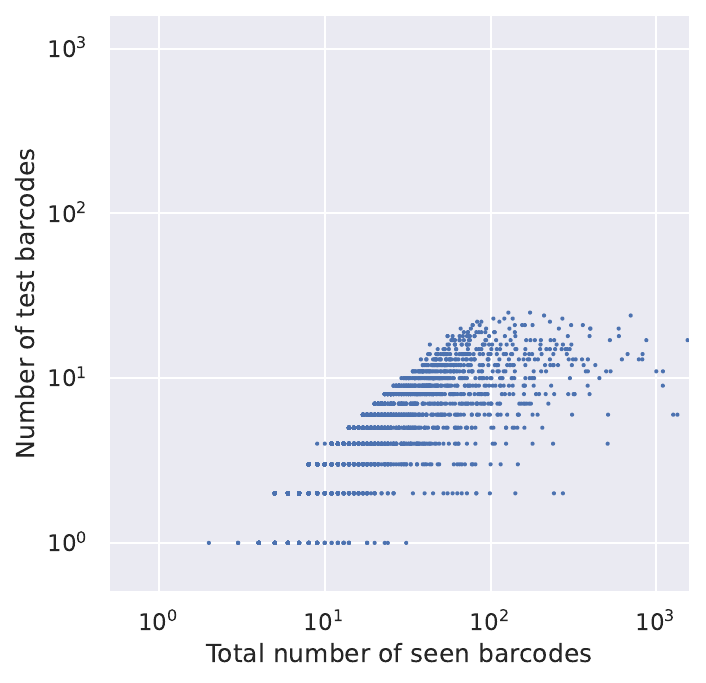}%
    \vspace{-2mm}%
    \caption{\texttt{test} partition.}
    \label{fig:transfer-barcodes-test}
    \end{subfigure}
    \hfill
    \begin{subfigure}{0.49\textwidth}%
    \vspace{2mm}%
    \includegraphics[width=\linewidth]{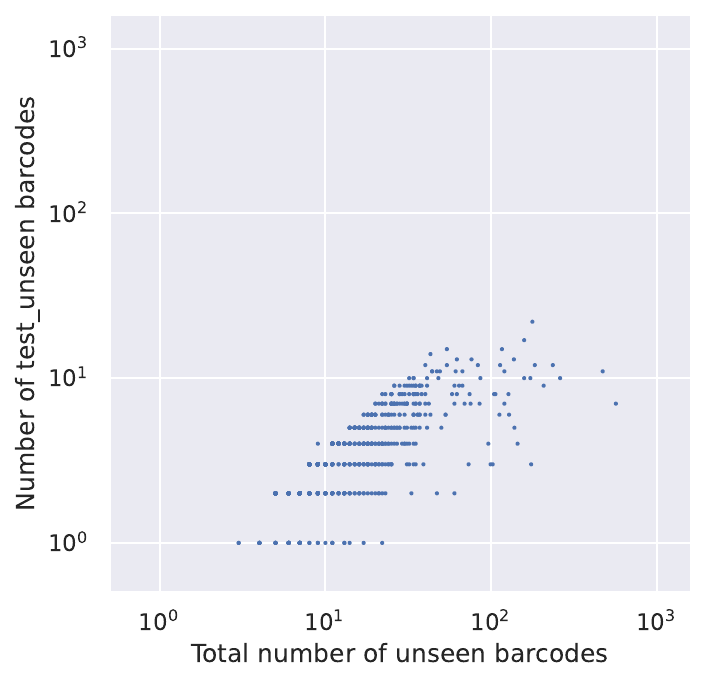}%
    \vspace{-2mm}%
    \caption{\texttt{test\_unseen} partition.}
    \end{subfigure}
    \caption{Number of barcodes in species set and partition, per species.}
    \label{fig:partition-transfer-barcodes}
\end{figure}

\begin{figure}[tbh]
    \centering
    \begin{subfigure}{0.49\textwidth}%
    \vspace{2mm}%
    \includegraphics[width=\linewidth]{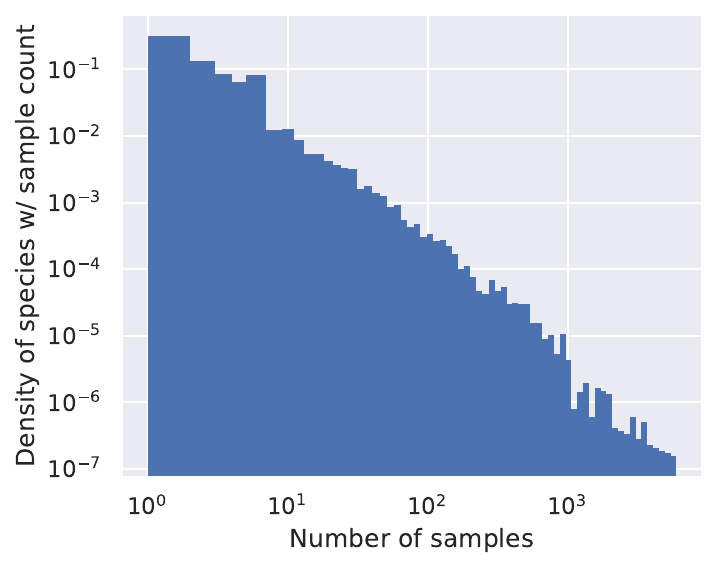}%
    \vspace{-2mm}%
    \caption{\texttt{train} partition.}
    \end{subfigure}
    \hfill
    \begin{subfigure}{0.49\textwidth}%
    \vspace{2mm}%
    \includegraphics[width=\linewidth]{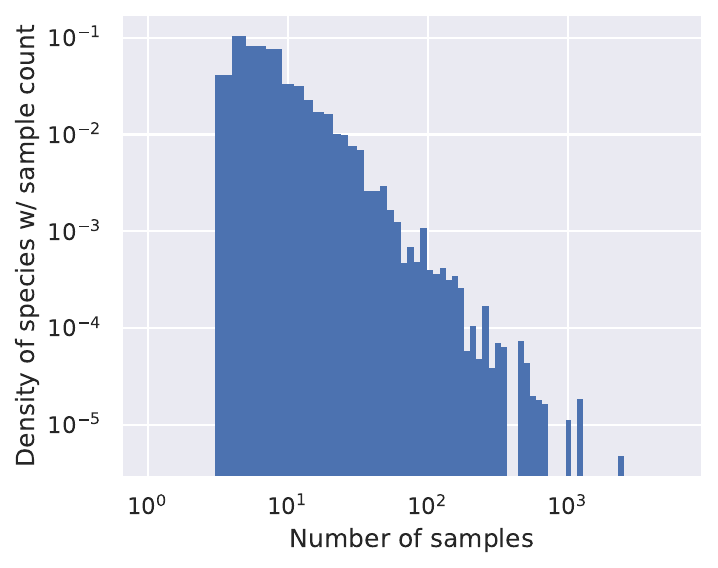}%
    \vspace{-2mm}%
    \caption{\texttt{key\_unseen} partition.}
    \end{subfigure}
    \\
    \begin{subfigure}{0.49\textwidth}%
    \vspace{2mm}%
    \includegraphics[width=\linewidth]{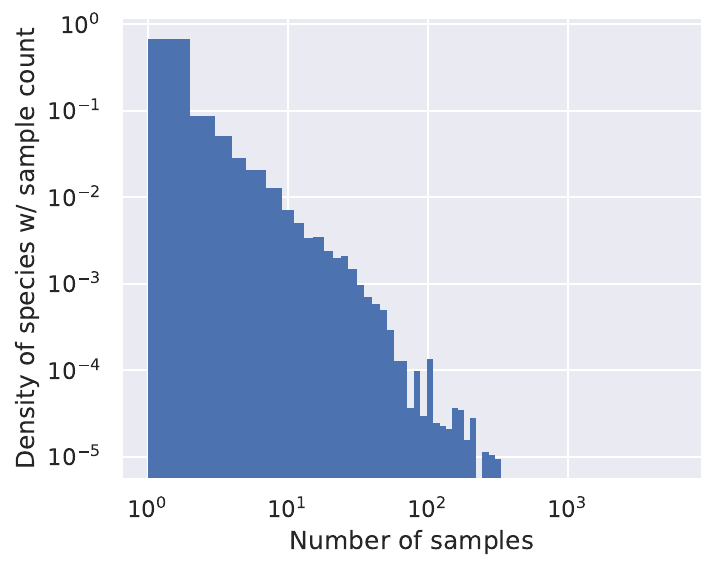}%
    \vspace{-2mm}%
    \caption{\texttt{val} partition.}
    \end{subfigure}
    \hfill
    \begin{subfigure}{0.49\textwidth}%
    \vspace{2mm}%
    \includegraphics[width=\linewidth]{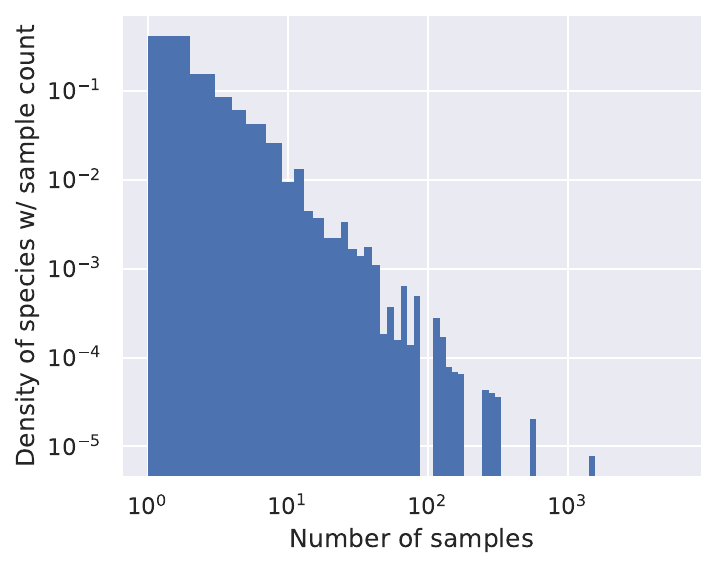}%
    \vspace{-2mm}%
    \caption{\texttt{val\_unseen} partition.}
    \end{subfigure}
    \\
    \begin{subfigure}{0.49\textwidth}%
    \vspace{2mm}%
    \includegraphics[width=\linewidth]{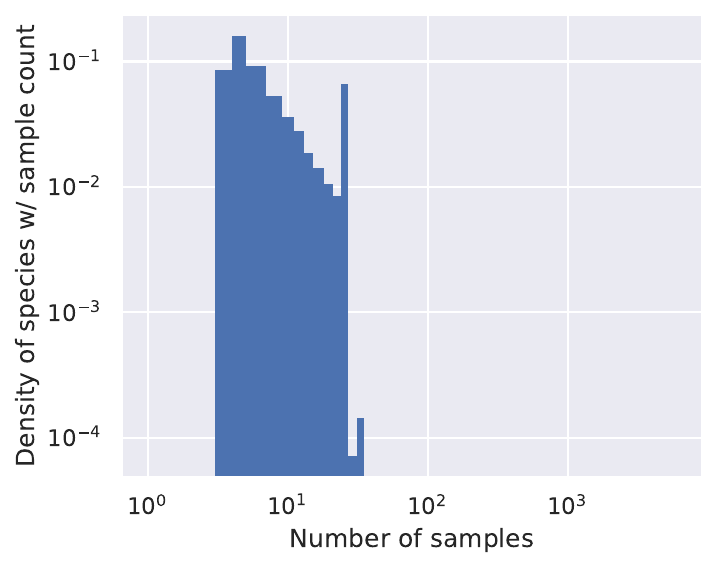}%
    \vspace{-2mm}%
    \caption{\texttt{test} partition.}
    \label{fig:partition-dist-test}
    \end{subfigure}
    \hfill
    \begin{subfigure}{0.49\textwidth}%
    \vspace{2mm}%
    \includegraphics[width=\linewidth]{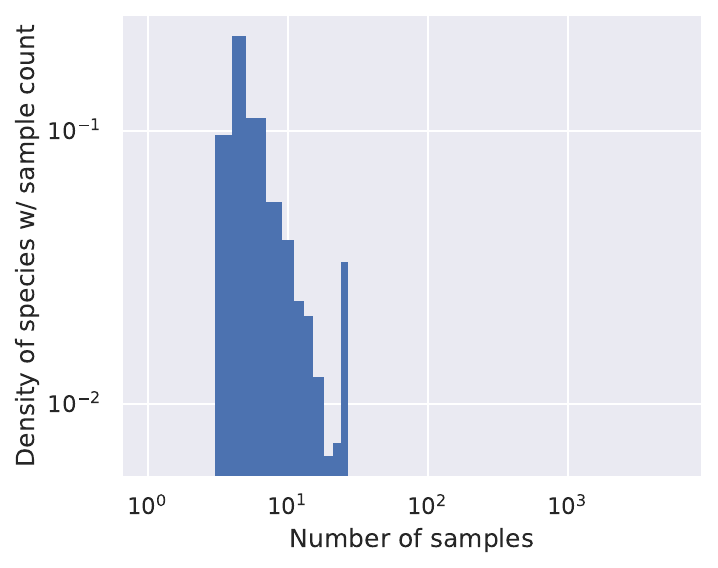}%
    \vspace{-2mm}%
    \caption{\texttt{test\_unseen} partition.}
    \end{subfigure}
    \caption{\textbf{Distribution of species prevalences across the main data partitions.} Note the log-log axes due to the power law distribution of the data.. The majority of species are infrequent, but some species have many samples. The \texttt{train} and \texttt{key\_unseen} partitions have similar distributions to the overall distribution for \emph{seen} and \emph{unseen} species. The \texttt{val} partitions have the same distribution, but shifted left as they they contain a fixed fraction of the samples per species. The \texttt{test} partitions are truncated with a minimum and maximum number of samples per species, which flattens the distribution over species for these partitions.}
    \label{fig:partition-dist}
\end{figure}

\begin{table}[tbh]
\centering
\caption{Number of species in common between each pair of partitions.}
\label{tab:partition_species_cmp}
\small
\begin{tabular}{lrrrrrrrr}
\toprule
  & \rotatebox{90}{\texttt{pretrain}} & \rotatebox{90}{\texttt{train}} & \rotatebox{90}{\texttt{val}} & \rotatebox{90}{\texttt{test}} & \rotatebox{90}{\texttt{key\_unseen}} & \rotatebox{90}{\texttt{val\_unseen}} & \rotatebox{90}{\texttt{test\_unseen}} & \rotatebox{90}{\texttt{other\_heldout}} \\
\midrule
\texttt{pretrain}       & \num{    0} & \num{    0} & \num{    0} & \num{    0} & \num{    0} & \num{    0} & \num{    0} & \num{    0} \\
\texttt{train}          & \num{    0} & \num{11846} & \num{ 3378} & \num{ 3483} & \num{    0} & \num{    0} & \num{    0} & \num{    0} \\
\texttt{val}            & \num{    0} & \num{ 3378} & \num{ 3378} & \num{ 2952} & \num{    0} & \num{    0} & \num{    0} & \num{    0} \\
\texttt{test}           & \num{    0} & \num{ 3483} & \num{ 2952} & \num{ 3483} & \num{    0} & \num{    0} & \num{    0} & \num{    0} \\
\texttt{key\_unseen}    & \num{    0} & \num{    0} & \num{    0} & \num{    0} & \num{  914} & \num{  903} & \num{  880} & \num{    0} \\
\texttt{val\_unseen}    & \num{    0} & \num{    0} & \num{    0} & \num{    0} & \num{  903} & \num{  903} & \num{  878} & \num{    0} \\
\texttt{test\_unseen}   & \num{    0} & \num{    0} & \num{    0} & \num{    0} & \num{  880} & \num{  878} & \num{  880} & \num{    0} \\
\texttt{other\_heldout} & \num{    0} & \num{    0} & \num{    0} & \num{    0} & \num{    0} & \num{    0} & \num{    0} & \num{ 9862} \\
\bottomrule
\end{tabular}
\end{table}

\begin{table}[tbh]
\centering
\caption{Fraction of species (\%) in common between each pair of partitions, relative to the number of species for the row.}
\label{tab:partition_species_pc}
\small
\begin{tabular}{lrrrrrrrr}
\toprule
  & \rotatebox{90}{\texttt{pretrain}} & \rotatebox{90}{\texttt{train}} & \rotatebox{90}{\texttt{val}} & \rotatebox{90}{\texttt{test}} & \rotatebox{90}{\texttt{key\_unseen}} & \rotatebox{90}{\texttt{val\_unseen}} & \rotatebox{90}{\texttt{test\_unseen}} & \rotatebox{90}{\texttt{other\_heldout}} \\
\midrule
\texttt{pretrain}       &   N/A &   N/A &   N/A &   N/A &   N/A &   N/A &   N/A &   N/A \\
\texttt{train}          &   0.0 & 100.0 &  28.5 &  29.4 &   0.0 &   0.0 &   0.0 &   0.0 \\
\texttt{val}            &   0.0 & 100.0 & 100.0 &  87.4 &   0.0 &   0.0 &   0.0 &   0.0 \\
\texttt{test}           &   0.0 & 100.0 &  84.8 & 100.0 &   0.0 &   0.0 &   0.0 &   0.0 \\
\texttt{key\_unseen}    &   0.0 &   0.0 &   0.0 &   0.0 & 100.0 &  98.8 &  96.3 &   0.0 \\
\texttt{val\_unseen}    &   0.0 &   0.0 &   0.0 &   0.0 & 100.0 & 100.0 &  97.2 &   0.0 \\
\texttt{test\_unseen}   &   0.0 &   0.0 &   0.0 &   0.0 & 100.0 &  99.8 & 100.0 &   0.0 \\
\texttt{other\_heldout} &   0.0 &   0.0 &   0.0 &   0.0 &   0.0 &   0.0 &   0.0 & 100.0 \\
\bottomrule
\end{tabular}
\end{table}

\begin{table}[tbh]
\centering
\caption{Number of genera in common between each pair of partitions.}
\label{tab:partition_genus_cmp}
\small
\begin{tabular}{lrrrrrrrr}
\toprule
  & \rotatebox{90}{\texttt{pretrain}} & \rotatebox{90}{\texttt{train}} & \rotatebox{90}{\texttt{val}} & \rotatebox{90}{\texttt{test}} & \rotatebox{90}{\texttt{key\_unseen}} & \rotatebox{90}{\texttt{val\_unseen}} & \rotatebox{90}{\texttt{test\_unseen}} & \rotatebox{90}{\texttt{other\_heldout}} \\
\midrule
\texttt{pretrain}       & \num{ 4260} & \num{ 2372} & \num{ 1190} & \num{ 1206} & \num{  217} & \num{  214} & \num{  209} & \num{  682} \\
\texttt{train}          & \num{ 2372} & \num{ 4930} & \num{ 1704} & \num{ 1736} & \num{  244} & \num{  240} & \num{  234} & \num{  519} \\
\texttt{val}            & \num{ 1190} & \num{ 1704} & \num{ 1704} & \num{ 1517} & \num{  151} & \num{  148} & \num{  145} & \num{  266} \\
\texttt{test}           & \num{ 1206} & \num{ 1736} & \num{ 1517} & \num{ 1736} & \num{  157} & \num{  154} & \num{  151} & \num{  276} \\
\texttt{key\_unseen}    & \num{  217} & \num{  244} & \num{  151} & \num{  157} & \num{  244} & \num{  240} & \num{  234} & \num{  177} \\
\texttt{val\_unseen}    & \num{  214} & \num{  240} & \num{  148} & \num{  154} & \num{  240} & \num{  240} & \num{  232} & \num{  175} \\
\texttt{test\_unseen}   & \num{  209} & \num{  234} & \num{  145} & \num{  151} & \num{  234} & \num{  232} & \num{  234} & \num{  172} \\
\texttt{other\_heldout} & \num{  682} & \num{  519} & \num{  266} & \num{  276} & \num{  177} & \num{  175} & \num{  172} & \num{ 1566} \\
\bottomrule
\end{tabular}
\end{table}

\begin{table}[tbh]
\centering
\caption{Fraction of genera (\%) in common between each pair of partitions, relative to the number of genera for the row.}
\label{tab:partition_genus_pc}
\small
\begin{tabular}{lrrrrrrrr}
\toprule
  & \rotatebox{90}{\texttt{pretrain}} & \rotatebox{90}{\texttt{train}} & \rotatebox{90}{\texttt{val}} & \rotatebox{90}{\texttt{test}} & \rotatebox{90}{\texttt{key\_unseen}} & \rotatebox{90}{\texttt{val\_unseen}} & \rotatebox{90}{\texttt{test\_unseen}} & \rotatebox{90}{\texttt{other\_heldout}} \\
\midrule
\texttt{pretrain}       & 100.0 &  55.7 &  27.9 &  28.3 &   5.1 &   5.0 &   4.9 &  16.0 \\
\texttt{train}          &  48.1 & 100.0 &  34.6 &  35.2 &   4.9 &   4.9 &   4.7 &  10.5 \\
\texttt{val}            &  69.8 & 100.0 & 100.0 &  89.0 &   8.9 &   8.7 &   8.5 &  15.6 \\
\texttt{test}           &  69.5 & 100.0 &  87.4 & 100.0 &   9.0 &   8.9 &   8.7 &  15.9 \\
\texttt{key\_unseen}    &  88.9 & 100.0 &  61.9 &  64.3 & 100.0 &  98.4 &  95.9 &  72.5 \\
\texttt{val\_unseen}    &  89.2 & 100.0 &  61.7 &  64.2 & 100.0 & 100.0 &  96.7 &  72.9 \\
\texttt{test\_unseen}   &  89.3 & 100.0 &  62.0 &  64.5 & 100.0 &  99.1 & 100.0 &  73.5 \\
\texttt{other\_heldout} &  43.6 &  33.2 &  17.0 &  17.6 &  11.3 &  11.2 &  11.0 & 100.0 \\
\bottomrule
\end{tabular}
\end{table}

\begin{figure}[tbh]
    \centering
\includegraphics[width=\linewidth]{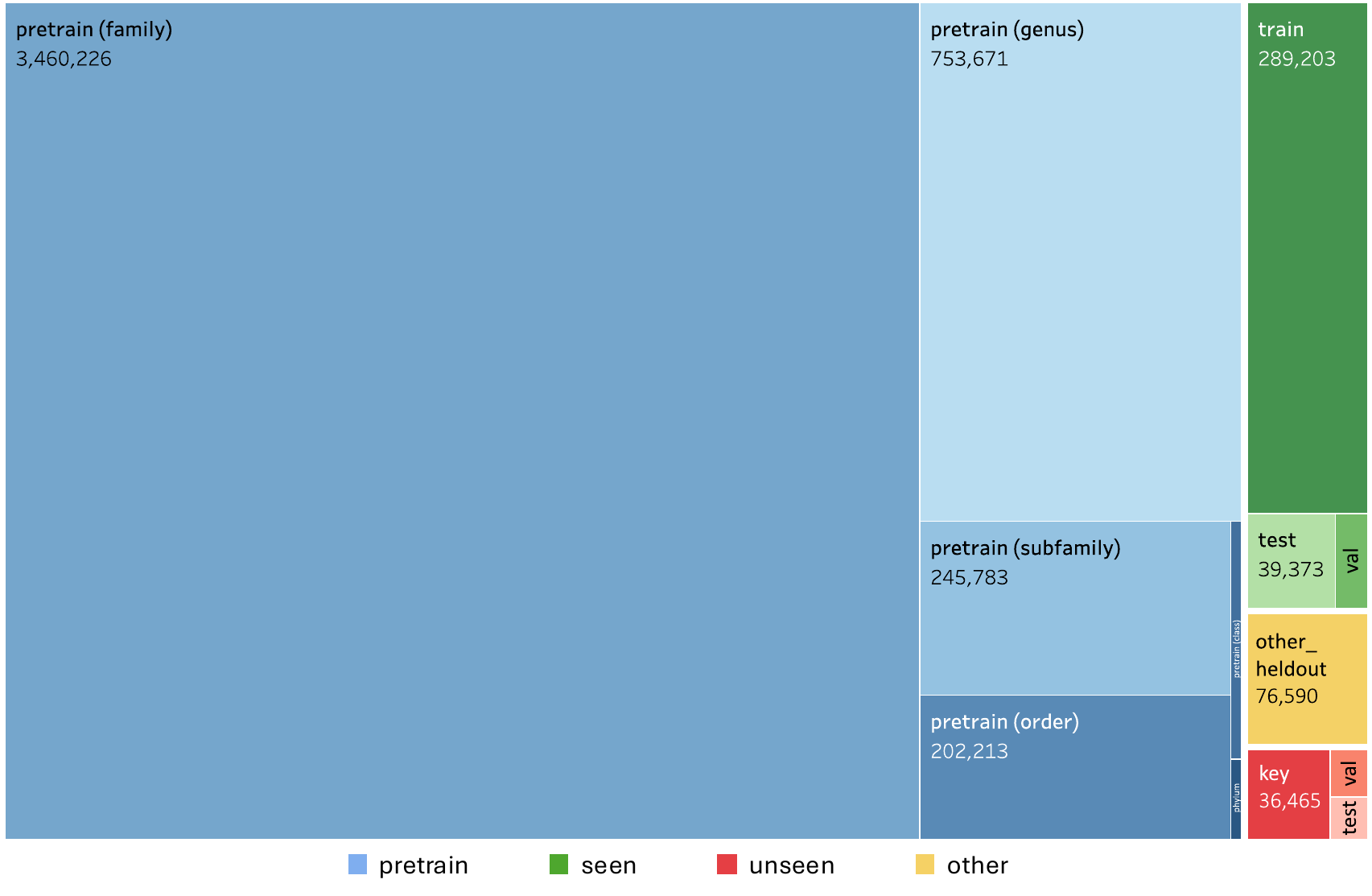}
    \caption{\textbf{Treemap diagram showing number of samples per partition.}
    For the \texttt{pretrain} partition (blues), we provide a further breakdown indicating the most fine-grained taxonomic rank that is labelled for the samples.
    For the remainder of the partitions (all of which are labelled to species level) we show the number of samples in the partition.
    Samples for seen species are shown in shades of green, and unseen in shades of red.
    }
    \label{fig:partition-area}
\end{figure}

To evaluate model performance during model development cycles, we also created a validation partition (\texttt{val}) with the same distribution as the \texttt{test} set. This was partition was created to contain around 5\% of the remaining samples from each of the seen species, by selecting barcodes to place in the \texttt{val} partition.
To mimic the long tail of the distribution, for each species with fewer than 20 samples and at least 6 samples, and for which one of their barcodes had only a single image, we added one single-image barcode to the \texttt{val} partition.
This step added \num{1766} individual samples from the tail; for comparison, our target of 5\% of the samples from the tail would be \num{1955} samples.

The remaining barcodes with samples of seen species are placed in the \texttt{train} partition.
For retrieval paradigms, we use the \texttt{train} partition as keys and the \texttt{val} and \texttt{test} partitions as queries.

For the unseen species, we use the same methodology as for the seen species to create and \texttt{val\_unseen}, with the exception that the proportion of samples placed in the \texttt{val\_unseen} partition was increased to 20\% to ensure it is large enough to be useful.
The remaining samples of unseen species are placed in the \texttt{keys\_unseen} partition.
For retrieval paradigms, we use the \texttt{keys\_unseen} partition as keys and the \texttt{val\_unseen} and \texttt{test\_unseen} partitions as queries.
For open world evaluation, we train on the \texttt{test} partition, without presenting any samples from the unseen species during training, and evaluate on \texttt{test\_unseen}.

The samples of heldout species are placed in the partition \texttt{other\_heldout}.
The utility of these species varies depending on the model paradigm.
In particular, we note that as these species are in neither the seen nor unseen species, they can be used to train a novelty detector without the novelty detector being trained on unseen species.

The samples of unknown species are placed entirely in the pretrain partition, which can be used for self-supervised pretraining, or semi-supervised learning.

To aid comparison between the coverage of the partitions, we show the number of species in common between each pair of partitions (\autoref{tab:partition_species_cmp}), and the percentage of species in common (\autoref{tab:partition_species_pc}).
This is a block-diagonal matrix as species labels do not overlap between species sets.
The \texttt{train} partition has higher diversity than the \texttt{val} and \texttt{test} partitions, which each cover less than 30\% of the seen species.
This is due to the long-tail of the distribution --- of the \num{11846} species, \num{7919} species (two thirds) have 6 or fewer samples, and of these \num{3756} species only have a single sample.
However, these rare species only constituted a small fraction of the \texttt{train} samples---only \num{17572} samples are members of species with 6 or fewer samples, which is 6\% of the \texttt{train} partition.
Due to our selection process for unseen species, in which only species with enough samples to be confident they are accurate are included, a much higher fraction of the unseen species are included in \texttt{val\_unseen} and \texttt{test\_unseen}.

Similarly, we show the number and percentage of genera in common between pairs of partitions (\autoref{tab:partition_genus_cmp} and \autoref{tab:partition_genus_pc}, respectively).
We see that the genera across all seen and unseen species set partitions are contained in the \texttt{train} partition.

In \autoref{fig:partition-area}, we show the number of samples per partition.
The plot illustrates the vast majority of the samples (91\%) are in the pretrain partition, and most samples are only labelled to family level (67\%).

\FloatBarrier

\begin{table}[tb]
\centering
\caption{
\textbf{Distribution of predominant classes and orders across data splits.}
For each taxonomic class present in the dataset, and selected orders which have a prevalence of at least 0.5\% for at least one split, we show the proportion of samples in each split (\%) bearing this taxonomic label.
Values for orders which never occur in a split are left empty.
Background: linear colour scale from 0\% (white) to 75\% (blue).
}
\label{tab:partition-dist-shift}
\resizebox{\textwidth}{!}{%
\small
\newcommand{\cellzero}{}
\newcommand{\cc}[1]{\cellcolor{#1}}
\definecolor{cbg}{cmyk}{0.8,0.4,0,0.012}
\begin{tabular}{llrrrrrrrr}
\toprule
 & & & \multicolumn{3}{c}{\textbf{Seen species}} & \multicolumn{3}{c}{\textbf{Unseen species}} \\
\cmidrule(lr){4-6} \cmidrule(lr){7-9}
\textbf{Class} & \textbf{Order}& \rotatebox{90}{\texttt{pretrain}}& \rotatebox{90}{\texttt{train}}& \rotatebox{90}{\texttt{val}}& \rotatebox{90}{\texttt{test}}& \rotatebox{90}{\texttt{key\_unseen}}& \rotatebox{90}{\texttt{val\_unseen}}& \rotatebox{90}{\texttt{test\_unseen}}& \rotatebox{90}{\texttt{other\_heldout}}\\
\midrule
Arachnida    & Araneae          & \cc{cbg!0}\num{ 0.35} & \cc{cbg!3}\num{ 2.25} & \cc{cbg!3}\num{ 2.15} & \cc{cbg!5}\num{ 4.11} & \cc{cbg!0}\num{ 0.17} & \cc{cbg!0}\num{ 0.16} & \cc{cbg!1}\num{ 0.49} & \cc{cbg!0}\num{ 0.14}\\
             & Mesostigmata     & \cc{cbg!0}\num{ 0.08} & \cc{cbg!0}\num{ 0.13} & \cc{cbg!0}\num{ 0.14} & \cc{cbg!0}\num{ 0.36} & \cc{cbg!1}\num{ 0.49} & \cc{cbg!1}\num{ 0.53} & \cc{cbg!1}\num{ 0.81} & \cc{cbg!0}\num{ 0.16}\\
             & Sarcoptiformes   & \cc{cbg!0}\num{ 0.09} & \cc{cbg!0}\num{ 0.34} & \cc{cbg!0}\num{ 0.33} & \cc{cbg!1}\num{ 0.61} & \cellzero{} & \cellzero{} & \cellzero{} & \cellzero{}\\
             & (Other)          & \cc{cbg!0}\num{ 0.31} & \cc{cbg!0}\num{ 0.13} & \cc{cbg!0}\num{ 0.12} & \cc{cbg!0}\num{ 0.27} & \cellzero{} & \cellzero{} & \cellzero{} & \cc{cbg!0}\num{ 0.01}\\
\addlinespace
Branchiopoda & (Total)          & \cc{cbg!0}\num{ 0.00} & \cc{cbg!0}\num{ 0.01} & \cc{cbg!0}\num{ 0.01} & \cc{cbg!0}\num{ 0.04} & \cellzero{} & \cellzero{} & \cellzero{} & \cellzero{}\\
\addlinespace
Chilopoda    & (Total)          & \cc{cbg!0}\num{ 0.00} & \cc{cbg!0}\num{ 0.00} & \cellzero{} & \cellzero{} & \cellzero{} & \cellzero{} & \cellzero{} & \cc{cbg!0}\num{ 0.01}\\
\addlinespace
Collembola   & Entomobryomorpha & \cc{cbg!1}\num{ 0.57} & \cc{cbg!4}\num{ 2.80} & \cc{cbg!4}\num{ 2.91} & \cc{cbg!1}\num{ 1.03} & \cc{cbg!0}\num{ 0.06} & \cc{cbg!0}\num{ 0.07} & \cc{cbg!0}\num{ 0.14} & \cc{cbg!1}\num{ 0.65}\\
             & (Other)          & \cc{cbg!0}\num{ 0.19} & \cc{cbg!1}\num{ 0.43} & \cc{cbg!1}\num{ 0.45} & \cc{cbg!1}\num{ 0.49} & \cc{cbg!0}\num{ 0.16} & \cc{cbg!0}\num{ 0.17} & \cc{cbg!0}\num{ 0.36} & \cc{cbg!0}\num{ 0.02}\\
\addlinespace
Copepoda     & (Total)          & \cc{cbg!0}\num{ 0.00} & \cc{cbg!0}\num{ 0.00} & \cellzero{} & \cellzero{} & \cellzero{} & \cellzero{} & \cellzero{} & \cellzero{}\\
\addlinespace
Diplopoda    & (Total)          & \cc{cbg!0}\num{ 0.00} & \cc{cbg!0}\num{ 0.00} & \cellzero{} & \cc{cbg!0}\num{ 0.01} & \cellzero{} & \cellzero{} & \cellzero{} & \cellzero{}\\
\addlinespace
Diplura      & (Total)          & \cc{cbg!0}\num{ 0.00} & \cellzero{} & \cellzero{} & \cellzero{} & \cellzero{} & \cellzero{} & \cellzero{} & \cellzero{}\\
\addlinespace
Insecta      & Coleoptera       & \cc{cbg!7}\num{ 5.02} & \cc{cbg!6}\num{ 4.47} & \cc{cbg!6}\num{ 4.20} & \cc{cbg!10}\num{ 7.44} & \cc{cbg!1}\num{ 0.39} & \cc{cbg!1}\num{ 0.43} & \cc{cbg!1}\num{ 0.94} & \cc{cbg!1}\num{ 0.48}\\
             & Diptera          & \cc{cbg!98}\num{73.64} & \cc{cbg!81}\num{60.56} & \cc{cbg!82}\num{61.75} & \cc{cbg!66}\num{49.21} & \cc{cbg!51}\num{38.44} & \cc{cbg!52}\num{38.96} & \cc{cbg!29}\num{21.74} & \cc{cbg!14}\num{10.19}\\
             & Hemiptera        & \cc{cbg!7}\num{ 5.06} & \cc{cbg!10}\num{ 7.75} & \cc{cbg!10}\num{ 7.51} & \cc{cbg!14}\num{10.15} & \cc{cbg!0}\num{ 0.18} & \cc{cbg!0}\num{ 0.12} & \cc{cbg!0}\num{ 0.36} & \cc{cbg!0}\num{ 0.05}\\
             & Hymenoptera      & \cc{cbg!14}\num{10.23} & \cc{cbg!16}\num{11.64} & \cc{cbg!15}\num{11.42} & \cc{cbg!21}\num{16.00} & \cc{cbg!50}\num{37.46} & \cc{cbg!49}\num{36.50} & \cc{cbg!83}\num{62.32} & \cc{cbg!56}\num{41.84}\\
             & Lepidoptera      & \cc{cbg!3}\num{ 2.51} & \cc{cbg!6}\num{ 4.75} & \cc{cbg!6}\num{ 4.26} & \cc{cbg!8}\num{ 5.96} & \cc{cbg!30}\num{22.65} & \cc{cbg!31}\num{23.04} & \cc{cbg!17}\num{12.79} & \cc{cbg!62}\num{46.39}\\
             & Psocodea         & \cc{cbg!1}\num{ 0.84} & \cc{cbg!3}\num{ 2.05} & \cc{cbg!3}\num{ 2.09} & \cc{cbg!2}\num{ 1.22} & \cellzero{} & \cellzero{} & \cellzero{} & \cc{cbg!0}\num{ 0.01}\\
             & Thysanoptera     & \cc{cbg!0}\num{ 0.20} & \cc{cbg!3}\num{ 2.02} & \cc{cbg!3}\num{ 2.04} & \cc{cbg!2}\num{ 1.77} & \cellzero{} & \cellzero{} & \cellzero{} & \cc{cbg!0}\num{ 0.00}\\
             & Trichoptera      & \cc{cbg!0}\num{ 0.17} & \cc{cbg!0}\num{ 0.24} & \cc{cbg!0}\num{ 0.24} & \cc{cbg!1}\num{ 0.58} & \cc{cbg!0}\num{ 0.01} & \cc{cbg!0}\num{ 0.01} & \cc{cbg!0}\num{ 0.05} & \cc{cbg!0}\num{ 0.01}\\
             & (Other)          & \cc{cbg!1}\num{ 0.38} & \cc{cbg!0}\num{ 0.31} & \cc{cbg!0}\num{ 0.30} & \cc{cbg!1}\num{ 0.66} & \cellzero{} & \cellzero{} & \cellzero{} & \cc{cbg!0}\num{ 0.06}\\
\addlinespace
Malacostraca & (Total)          & \cc{cbg!0}\num{ 0.00} & \cc{cbg!0}\num{ 0.09} & \cc{cbg!0}\num{ 0.08} & \cc{cbg!0}\num{ 0.10} & \cellzero{} & \cellzero{} & \cellzero{} & \cellzero{}\\
\addlinespace
Ostracoda    & (Total)          & \cc{cbg!0}\num{ 0.00} & \cc{cbg!0}\num{ 0.00} & \cellzero{} & \cellzero{} & \cellzero{} & \cellzero{} & \cellzero{} & \cellzero{}\\
\bottomrule
\end{tabular}
}%
\end{table}

\begin{figure}[tbh]
    \centering
    \begin{subfigure}{0.495\textwidth}%
    \vspace{2mm}%
    \centerline{\includegraphics[scale=0.7]{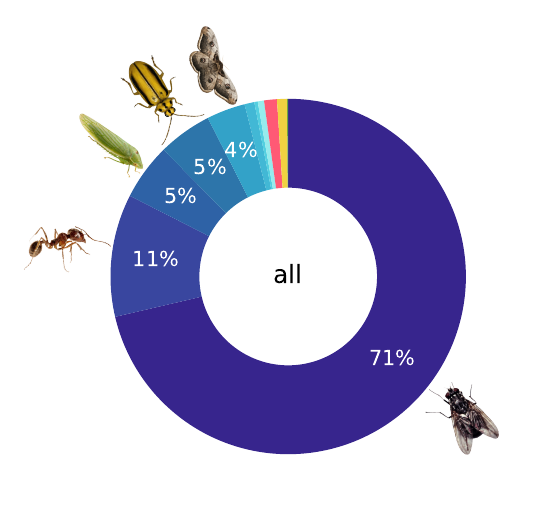}}%
    \vspace{-2mm}%
    \caption{All samples.}
    \label{fig:partition-dist-shift--all}
    \end{subfigure}
    \hfill
    \begin{subfigure}{0.495\textwidth}%
    \vspace{2mm}%
    \centerline{\includegraphics[scale=0.7]{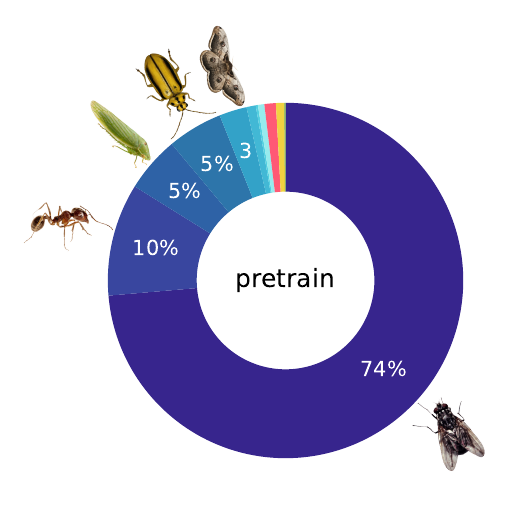}}%
    \vspace{-2mm}%
    \caption{\texttt{pretrain} partition.}
    \label{fig:partition-dist-shift--pretrain}
    \end{subfigure}
    \\
    \begin{subfigure}{0.495\textwidth}%
    \vspace{2mm}%
    \centerline{\includegraphics[scale=0.7]{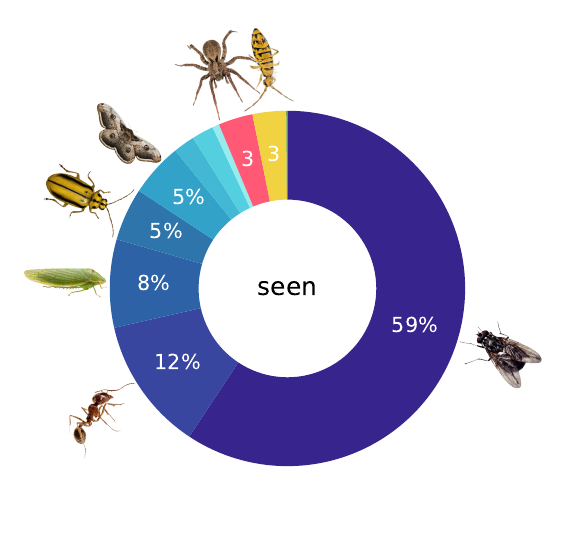}}%
    \vspace{-2mm}%
    \caption{Seen species (\texttt{train}, \texttt{val}, and \texttt{test}).}
    \label{fig:partition-dist-shift--seen}
    \end{subfigure}
    \hfill
    \begin{subfigure}{0.495\textwidth}%
    \vspace{2mm}%
    \centerline{\includegraphics[scale=0.7]{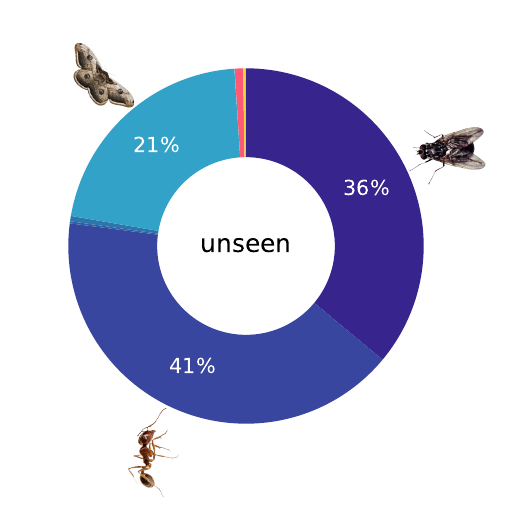}}%
    \vspace{-2mm}%
    \caption{\raggedright Unseen species (\texttt{key\_unseen}, \texttt{val\_unseen}, \texttt{test\_unseen}).}
    \label{fig:partition-dist-shift--unseen}
    \end{subfigure}
    \\
    \begin{subfigure}{0.495\textwidth}%
    \vspace{2mm}%
    \centerline{\includegraphics[scale=0.7]{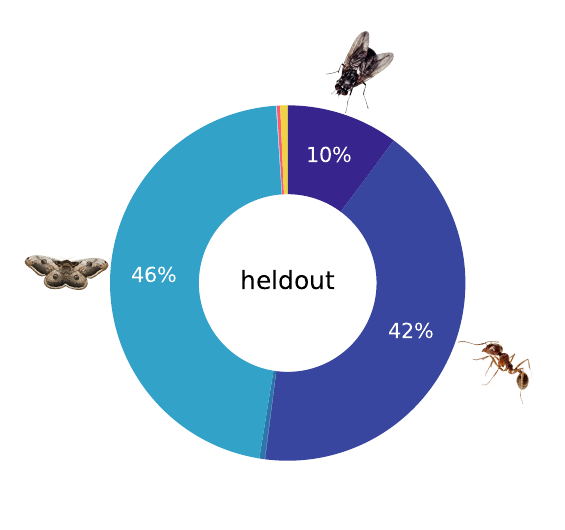}}%
    \vspace{-2mm}%
    \caption{\texttt{other\_heldout} partition.}
    \label{fig:partition-dist-shift--heldout}
    \end{subfigure}
    \caption{\textbf{Distribution of classes and insect orders.} In each panel, the distribution of taxa is shown for one species set of the dataset. Classes are shown in different hues---Arachnida: red, Collembola: yellow, Insecta orders: shades of blue varying by order, other classes: green. Icons are redistributed under CC BY(-NC) or Canva pro license, respectively. See \autoref{tab:partition-dist-shift} for names and more detailed values.}
    \label{fig:partition-dist-shift}
\end{figure}

\begin{figure}[tbh]
    \centering
    \begin{subfigure}{0.495\textwidth}%
    \vspace{2mm}%
    \centerline{\includegraphics[scale=0.7]{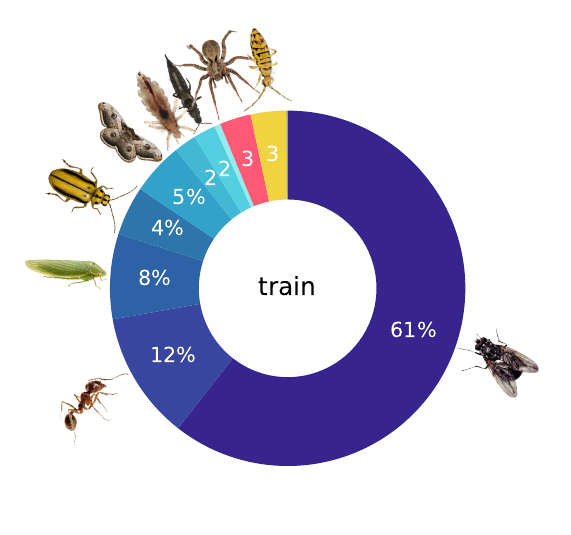}}%
    \vspace{-2mm}%
    \caption{\texttt{train} partition.}
    \label{fig:partition-dist-shift--train}
    \end{subfigure}
    \hfill
    \begin{subfigure}{0.495\textwidth}%
    \vspace{2mm}%
    \centerline{\includegraphics[scale=0.7]{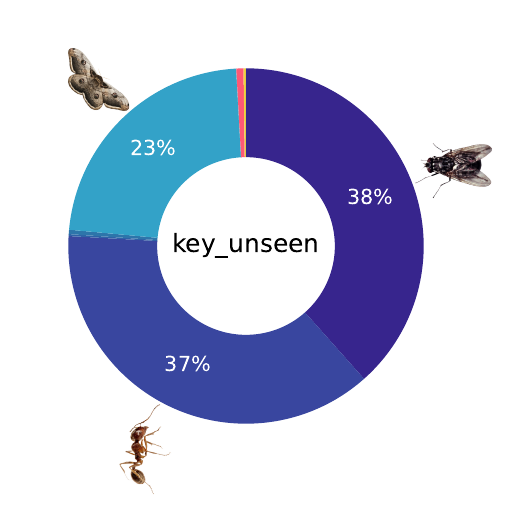}}%
    \vspace{-2mm}%
    \caption{\texttt{key\_unseen} partition.}
    \label{fig:partition-dist-shift--keyunseen}
    \end{subfigure}
    \\
    \begin{subfigure}{0.495\textwidth}%
    \vspace{2mm}%
    \centerline{\includegraphics[scale=0.7]{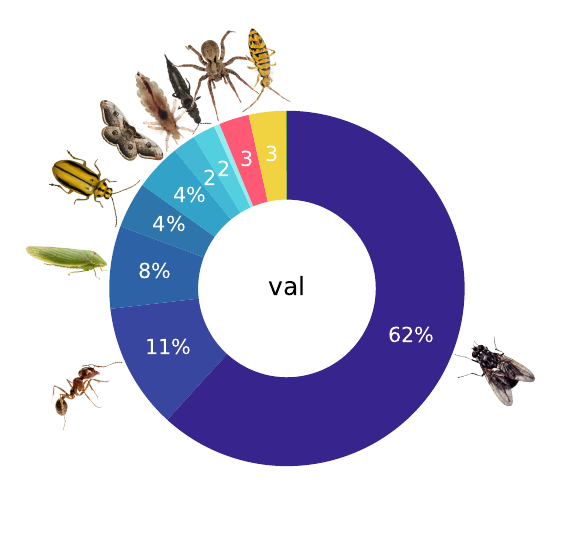}}%
    \vspace{-2mm}%
    \caption{\texttt{val} partition.}
    \label{fig:partition-dist-shift--val}
    \end{subfigure}
    \hfill
    \begin{subfigure}{0.495\textwidth}%
    \vspace{2mm}%
    \centerline{\includegraphics[scale=0.7]{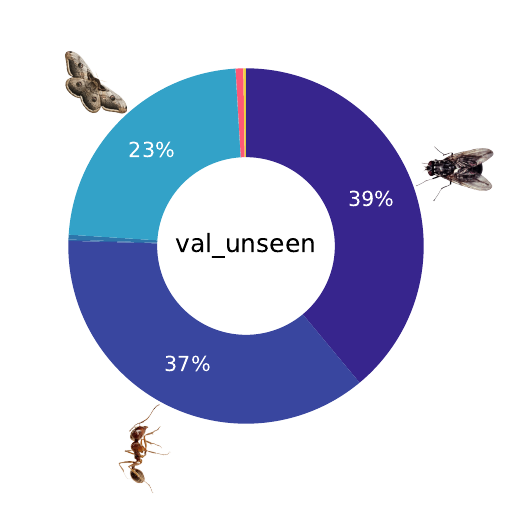}}%
    \vspace{-2mm}%
    \caption{\texttt{val\_unseen} partition.}
    \label{fig:partition-dist-shift--valunseen}
    \end{subfigure}
    \\
    \begin{subfigure}{0.495\textwidth}%
    \vspace{2mm}%
    \centerline{\includegraphics[scale=0.7]{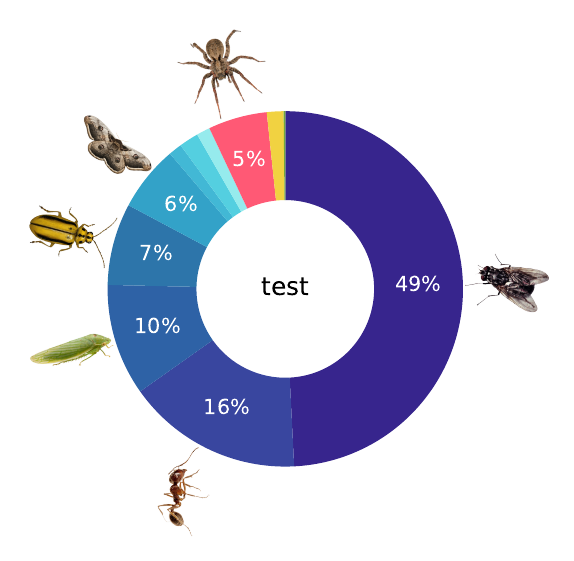}}%
    \vspace{-2mm}%
    \caption{\texttt{test} partition.}
    \label{fig:partition-dist-shift--test}
    \label{fig:dist-test}
    \end{subfigure}
    \hfill
    \begin{subfigure}{0.495\textwidth}%
    \vspace{2mm}%
    \centerline{\includegraphics[scale=0.7]{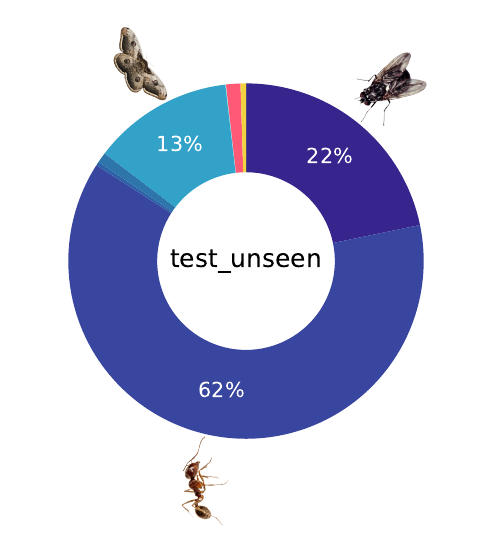}}%
    \vspace{-2mm}%
    \caption{\texttt{test\_unseen} partition.}
    \label{fig:partition-dist-shift--testunseen}
    \end{subfigure}
    \caption{\textbf{Distribution of classes and insect orders.} Each panel shows the distribution for one partition. Classes are shown in different hues---Arachnida: red, Collembola: yellow, Insecta orders: shades of blue varying by order, other classes: green. Icons are redistributed under CC BY(-NC) or Canva pro license, respectively. See \autoref{tab:partition-dist-shift} for names and more detailed values.}
    \label{fig:partition-dist-shift-split}
\end{figure}

\subsection{Distributional shift}
\label{a:distribution-shift}
As described above, we partitioned our data into sets to use for open- and closed-world tasks.
The division of our data was directed by the labels, with scientific names places in the ``seen'' species set and placeholder names in the ``unseen'' species set.
This partitioning method means our open-world dataset should be, by construction, well-aligned with the open-world task seen in practice for novel data collection.
Novel arthropod species are continually being discovered and identified as species that are new to science, and if we assume there is a uniform efficiency for naming across taxa the distribution is of placeholder names is likely to match the distribution of new species discovery.
However, this distribution does not necessarily match that of taxa prevalence, due to several factors such as non-uniform speciation rates across arthropods.

We investigated the difference in the distribution at order and class level for the dataset partitions, tabulated in \autoref{tab:partition-dist-shift} and illustrated in Figures \ref{fig:partition-dist-shift} and \ref{fig:partition-dist-shift-split}.
We observe that the Diptera (fly) class of Insecta dominates the overall and \texttt{pretrain} dataset (\autoref{fig:partition-dist-shift--pretrain}), but ``seen'' partitions (\autoref{fig:partition-dist-shift--seen}) have a flatter distribution with more prevalence of two non-Insecta orders---Arachnida (spiders, etc.) and Collembola (springtails)---and more instances of non-Diptera Insecta classes.
The distribution is even flatter for the \texttt{test} partition (\autoref{fig:partition-dist-shift--test}), due to our capped subsampling methodology when creating the partition.

For ``unseen'' partitions (\autoref{fig:partition-dist-shift--unseen}), we find the data is split nearly equally between three dominant Insecta classes---Diptera (flies), Hymenoptera (bees, ants, etc.), and Lepidoptera (moths, etc.).
The \texttt{test\_unseen} partition (\autoref{fig:partition-dist-shift--testunseen}) contains even more Hymenoptera (around 62\%).
The \texttt{other\_heldout} partition (\autoref{fig:partition-dist-shift--heldout}) has even less Diptera, and is instead dominated by Lepidoptera and Hymenoptera.

Users of the BIOSCAN-5M dataset should thus be sure to consider the effect of this distributional shift on their results if they wish to make direct comparisons between the \texttt{test} performance and the \texttt{test\_unseen} performance---results for these partitions are not intended to be directly comparable to each other.

}{}

\end{document}